\documentclass[twoside]{article}

\usepackage[accepted]{aistats2022}
\usepackage{authblk}

\usepackage[utf8]{inputenc} % allow utf-8 input
\usepackage{hyperref}       % hyperlinks
\usepackage{url}            % simple URL typesetting
\usepackage{booktabs}       % professional-quality tables
\usepackage{amsfonts}       % blackboard math symbols
\usepackage{nicefrac}       % compact symbols for 1/2, etc.
\usepackage{microtype}      % microtypography
\usepackage{xcolor}         % colors
\usepackage[round]{natbib}         % citet and citep

\usepackage{amsmath, amsthm}
\usepackage{bm, bbm} % boldfaced math symbols
\usepackage{comment} % for the comment environment
\usepackage{float} % better placement of figures
\usepackage{graphicx}
\usepackage{subcaption} % for subfigures
\usepackage{gp_robustness}

\usepackage{algorithm}
\usepackage[noend]{algpseudocode}
\usepackage{algorithmicx}
\usepackage{selectp}
\algnewcommand{\LineComment}[1]{  \(\triangleright\) #1}
\definecolor{SkyBlue}{RGB}{14, 118, 188}
\definecolor{BrightRed}{RGB}{223,82, 78}

\hypersetup{pdfborder = {0 0 0.5 [3 3]}, colorlinks = true, linkcolor = BrightRed, citecolor = SkyBlue}

\DeclareMathOperator*{\argmax}{arg\,max} % needs to be in the preamble

\begin{document}
\title{Measuring the robustness of Gaussian processes to kernel choice}
\onecolumn
\author[1,3]{William T. Stephenson}
\author[2,3]{Soumya Ghosh}
\author[1,3]{Tin D. Nguyen}
\author[2,3]{Mikhail Yurochkin}
\author[1,3]{Sameer K. Deshpande}
\author[1,3]{Tamara Broderick}
\affil[1]{MIT}
\affil[2]{IBM Research}
\affil[3]{MIT-IBM Watson AI Lab}
\maketitle

\begin{abstract}
Gaussian processes (GPs) are used to make medical and scientific decisions, including in cardiac care and monitoring of atmospheric carbon dioxide levels. Notably, the choice of GP kernel is often somewhat arbitrary. In particular, uncountably many kernels typically align with qualitative prior knowledge (e.g.\ function smoothness or stationarity). But in practice, data analysts choose among a handful of convenient standard kernels (e.g.\ squared exponential). In the present work, we ask: Would decisions made with a GP differ under other, qualitatively interchangeable kernels? We show how to answer this question by solving a constrained optimization problem over a finite-dimensional space. We can then use standard optimizers to identify substantive changes in relevant decisions made with a GP. We demonstrate in both synthetic and real-world examples that decisions made with a GP can exhibit non-robustness to kernel choice, even when prior draws are qualitatively interchangeable to a user.
\end{abstract}

\section{INTRODUCTION}
\label{sec:intro}
\begin{figure}
\centering
\includegraphics[width=0.48\textwidth]{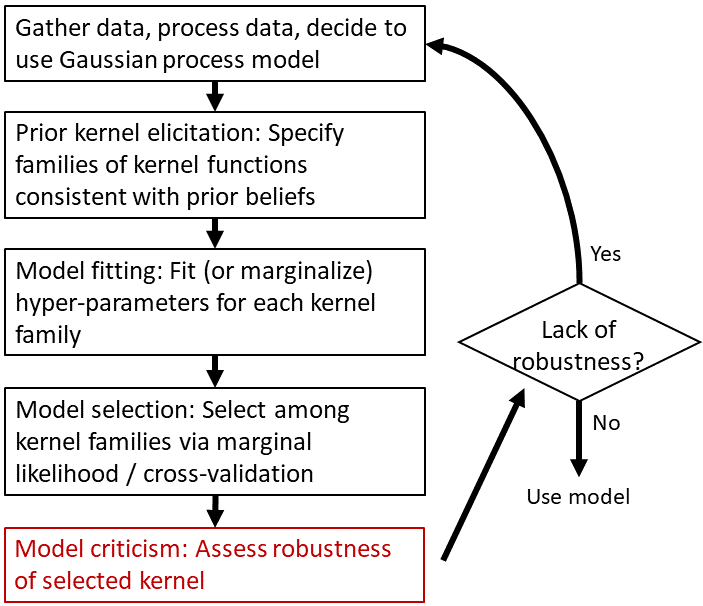}
\caption{\small{Where we sit in the modeling workflow.}}
\label{fig:workflow}
\end{figure}

Gaussian processes (GPs) enable practitioners to estimate flexible functional relationships between predictors and outcomes.
GPs have been used to 
monitor physiological vital signs in hospital patients \citep[e.g.][]{Cheng2020_medgp, Colopy2016_identifying, Futoma2017_learning, Futoma2017_improved}, to estimate the health effects of exposure to airborne pollutants \citep[e.g.][]{FerrariDunson2020_identifying, Lee2017_rigorous, Ren2021_pollution}, and in many other medical and scientific settings.
To use a GP for any application, a practitioner must choose a covariance kernel. The kernel determines the shape, smoothness, and other properties of the latent function of interest \citep[Chap.\ 2]{Duvenaud_thesis}.
Ideally a user would specify a kernel that exactly encodes all of their prior beliefs about the latent function. In practice,
a user often has only vague qualitative prior information and typically selects a kernel from a relatively small set of commonly used families. It seems plausible that other kernels could have been equally compatible with the user's beliefs. 
When a user has no reason to prefer one kernel over another given their prior beliefs, we call the kernels \emph{qualitatively interchangeable}.
We would worry if substantive medical or scientific decisions changed when using a qualitatively interchangeable kernel: that is, if real-life decisions are non-robust to the choice of kernel.
In this paper, we propose a workflow to assess the robustness of the GP posterior under qualitatively interchangeable choices of the kernel. \cref{fig:workflow} situates our work, an example of \emph{model criticism}, in the modeling workflow.  

\textbf{Related work.} Robustness and sensitivity of data analyses to data and model choice have been studied for decades \citep{Andrews1972_robustness, Huber2009_robustStats, Goodfellow2015_adversarial}.
In the context of Bayesian methods, sensitivity to the choice of prior has been studied as well \citep{Berger1994_overview,Gustafson1996_local, Berger2000, Giordano2021_stickbreaking}.
These works assess sensitivity by varying the prior within a small epsilon-ball around the user-specified prior with the intuition that a small ball will mostly contain priors that are acceptable alternatives to the user-specified prior.
In contrast, we note that the class of qualitatively interchangeable kernels is actually the class of alternative priors of interest; we explicitly define and study sensitivity within this class.

Our focus on robustness to kernel choice stands in  contrast to existing work on robustness in GPs, which focus on robustness to data perturbations \citep{KimGhahramani2008_outlier,Hernandez-Lobato2011_multiclass, Jylanki2011_student-t,Ranjan2016_em,Bogunovic2018_optimization,Cardelli2019_aaai}.
Our focus is also distinct from that of works studying convergence rates \citep{vaart2011_gpRates, teckentrup2020_gpRates, WangJing2021_convergence, Wynne2021_convergence} and asymptotic predictive equivalence \citep{Stein1993_kernelEquivalence,Bevilacqua2019_kernelEquivalence,Kirchner2021_kernelEquivalence} for GP regression with misspecified kernels.
These works do not examine how kernel choice affects non-linear functionals like posterior variances, do not study what happens in finite samples, and do not consider kernel choice as an issue of prior specification as we do. 
%Specifically, these existing works do not examine how kernel choice affects non-linear functionals like posterior variance nor do they study what happens in finite samples, as we do. 
%But none of these works examine how kernel choice affects non-linear functionals like posterior variances, nor do they study what happens in finite samples.
%Moreover, these works do not consider kernel choice as an issue of prior specification, and so do not study issues of qualitative interchangeability as we do.
One might hope that automatic kernel discovery procedures \cite[e.g.][]{Benton2019_function,Duvenaud2013_compositional,WilsonAdams2013_sm,Wilson2016_dkl} ostensibly obviate the need for careful kernel specification.
However, choosing a particular kernel via model fitting does not preclude that many other kernels might satisfy a user’s prior beliefs. Indeed, we show that decisions made with kernels selected via modern model fitting can still exhibit non-robustness to kernel choice (\cref{app:co2_ac}).
%One option to avoid issues of kernel robustness is to learn a GP kernel rather than specifying one a priori \citep{Benton2019_function, Duvenaud2013_compositional,WilsonAdams2013_sm,Wilson2016_dkl}. 
%But parameter learning and prediction can often be more efficient when prior information is used, so despite these advances, a great many practitioners still directly choose their kernels. 

\textbf{Model robustness versus model sensitivity.}
We emphasize that the goal of our work is to assess model robustness, which we now distinguish from model sensitivity. \emph{Model sensitivity} measures how much our reported estimates change when we change our model. To assess \emph{model robustness}, though, we also need to know how the model is being used. Often in applied analyses, many models reasonably reflect our prior beliefs and there is an application-specific threshold beyond which changes in reported estimates are deemed important; if our sensitivity is sufficiently high that we can observe an “important” change within the “reasonable” models, we say that our conclusion is \emph{(model) non-robust}. While sensitivity is an objective and measurable quantity, model robustness is inherently qualitative and user-dependent. So the methods we present here are (and should be) qualitative and user-dependent. Although these general ideas are well-established \citep{Berger1994_overview,insua2000_robustBayes}, operationalizing them in the context of GPs is novel and challenging.

\textbf{Our contributions.} We propose and implement the first workflow to discover whether applied decisions based on a GP posterior are robust to the choice of the user-specified prior (i.e.\ the kernel). Our workflow proceeds as follows. 
(A) Keep expanding a class of appropriate kernels around the original kernel until some kernel in this class yields a substantively different decision. 
(B) Assess if this decision-changing kernel is qualitatively interchangeable with the original kernel. If the two kernels \emph{are} interchangeable, we conclude the decision is not robust; a different decision could be reached with the same prior information. 
If the two kernels \emph{are not} interchangeable, we cannot conclude non-robustness. 
We provide a practical implementation of steps (A) and (B).
We demonstrate the practical utility of our workflow by discovering non-robustness in various applied uses of Gaussian processes: (1) predicting whether a hospital patient's heart rate is alarmingly high, (2) predicting future carbon dioxide levels, and (3) classifying MNIST handwritten digits.

%We demonstrate our workflow in various applications: (1) predicting whether an adult patient's resting heart rate exceeds 130 beats per minute (bpm), a potentially dangerous level \citep{Fidler2017_heartRateThreshold}, is sometimes flagged as non-robust by our method and sometimes not; (2) predicted future carbon dioxide levels \citep{RasmussenWilliams2006} can be substantively increased with a qualitatively interchangeable kernel; (3) predictions on MNIST handwritten digits can be altered with minor kernel perturbations.
%Our experiments demonstrate our workflow's ability to assess prior robustness in real analyses using Gaussian processes.
%\input{related-work}

\section{OUR WORKFLOW}
%\subsection{Overview of our approach}
\label{sec:overview}
%\begin{figure}
%\centering
%\includegraphics[width=0.6\linewidth]{figures/figure1b}	
%\caption{\small{The set of kernels qualitatively interchangeable with $k_0$, $K_{QI}$, versus $\Keps$.
%$K_{QI}$ is likely difficult to optimize over and challenging for the user to specify, so we instead optimize over the more regular balls $\Keps$ and check that the result in $K_{QI}$ post-optimization.}}
%\label{fig:searchspace}
%\end{figure}
%
\textbf{Setup and notation.} Consider data $\Data = \{(\bx_{n}, y_{n})\}_{n = 1}^{N}$, with covariates $\bx_n \in \R^{D}$ and outcomes $y_n \in \R.$
We model this data as $y_n \sim \mathcal{N}(f(x_n), \sigma^2)$, where $f: \R^D \to \R$ is an unknown function, and $\sigma > 0$ is a noise parameter. We place a zero-mean Gaussian process (GP) prior on $f$ with kernel $k$.
Equivalently, we place a zero-mean GP prior on $\{y_n\}_{n=1}^N$ with kernel $k'(x_n, x_m) := k(x_n, x_m) + \sigma^2 \delta_{nm}$.
Going forward, we assume that kernels $k'$ are comprised of a base kernel $k$ plus a noise term $\sigma^2 \delta_{nm}$.
Typical base kernels $k$ depend on a vector of hyperparameters $\theta_k$; for convenience, we define $\theta := (\theta_k, \sigma^2)$. Unless stated otherwise, we assume that $\theta$ is estimated using maximum marginal likelihood estimation (MMLE).
%That is, $\theta = \argmax_{\tilde{\theta}} p(Y \mid X, \tilde{\theta})$. 
That is, 
\begin{align}
\hat\theta = (\hat{\theta}_k, \hat\sigma^2) = \argmax_{\theta_k, \sigma^2} p(y_{1:N}, \mid \bx_{1:N}, \theta_k, \sigma^2). \label{eq:MMLE}
\end{align}
Let $k_0$ be the practitioner-chosen base kernel with MMLE hyperparameters $\hat\theta_k$.
Let $\Fstar(k)$ be any scalar functional of the posterior $f \mid \Data$ that is a differentiable function of the base kernel $k$.
Let the level $\bigChange \in \R$ represent a decision threshold in $\Fstar(k)$. That is, we make one decision when $\Fstar(k) \ge \bigChange$ and a different one when $\Fstar(k) < \bigChange$.
For example, let time be a single covariate, and let outcome be the resting heart rate of a hospital patient. $\Fstar(k)$ could be the 95th percentile of the GP posterior at a given time; an alarm might trigger if $\Fstar(k)$ is greater than $\bigChange=$130 bpm but not otherwise \citep{Fidler2017_heartRateThreshold}.
While many applied examples use $\Fstar$ as a function of the posterior at a single test point $\bx^*$, we stress that $\Fstar$ can be any differentiable function of the posterior,
e.g.\ the smooth maximum of means at a set of test points (\cref{sec:co2}).

We want to assess whether our decision would change if we used a different, but qualitatively interchangeable, kernel. Without loss of generality, we assume that $\Fstar(k_0) < \bigChange$. Then we can define non-robustness.
\begin{defn}
	For original kernel $k_0$, we say that our decision $\Fstar(k_0) < \bigChange$ is non-robust to the choice of kernel if there exists a kernel $k_1$ that is qualitatively interchangeable with $k_0$ and $\Fstar(k_1) \ge \bigChange$.
	\label{def:nonRobust}
\end{defn}
%
%Note that in \cref{def:nonRobust} we only consider robustness to the specification of $k_0$, and not the choice of $\hat\sigma$.
%While one could study robustness to $\hat\sigma$, the choice to include a noise parameter and estimate it via MMLE is fairly standard, and so does not represent a particularly arbitrary decision.
%On the other hand, there are many choices available for $k_0$, and so we ask how robust decisions are to this choice.
We emphasize that non-robustness as defined in \cref{def:nonRobust} is dependent on a number of user-dependent quantities -- $k_0$, $\Fstar$, $L$, and the user's qualitative prior beliefs -- as well as the particular observed data $\{(\bx_{n}, y_{n})\}_{n = 1}^{N}$ and $\hat\sigma$.
Also note that here we assess robustness to the specification of the GP governing the function $f$. In the present work, we do not assess sensitivity to the choice of i.i.d.\ normal noise around $f$ or the choice to use a GP prior at all. But allowing more components of the model to vary can only increase sensitivity. So if we find an analysis is non-robust using our current methods, the analysis would remain non-robust if we allowed more model variation.

\textbf{Workflow overview.}
Our workflow is summarized in \cref{alg:workflow}. We start by defining a set $\Keps$ of kernels that are ``$\eps$-near'' $k_0$.
We then optimize:
\begin{align}
	&k_1(\eps) := \argmax_{k \in \Keps} \Fstar(k) \nonumber \\
	&\eps^* =\ \text{smallest}\ \eps\ \text{s.t.}\ \Fstar(k_1(\eps)) \geq L.
	\label{mainOptimizationProblem}
\end{align}
To find $\eps^*$, we slowly increase $\eps$ until $k_1(\eps)$ changes our decision.
We then check whether the decision-changing kernel, $\kperturb$, is qualitatively interchangeable with $k_0$. 
%\cref{fig:searchspace} provides a cartoon illustration of this procedure. 
It remains to precisely define a set of ``$\eps$-near'' kernels and show that we can efficiently solve \cref{mainOptimizationProblem} (\cref{sec:nearbyKernels}), and to provide ways to assess qualitative interchangeability (\cref{sec:QI}).

Note that although \cref{alg:workflow} can detect non-robustness, it cannot certify robustness; it is possible, even though it may be unlikely, that there exists a qualitatively interchangeable kernel that the methodology has not detected but that still changes the decision. 
The inability to decisively declare an analysis robust is generally true of robustness analyses, and the present workflow is no exception. 
This observation is similar in spirit to classical hypothesis tests: a user can reject -- but not accept -- a null hypothesis.

\begin{algorithm}
    \caption{Workflow for assessing robustness of GP inferences to kernel choice}
    \label{alg:workflow}
    \begin{algorithmic}[1] % The number tells where the line numbering should start
    		\State Choose initial kernel $k_0$ using prior information.
        	\State Choose posterior quantity of interest $\Fstar$. \LineComment{E.g.\ Posterior mean at test point $\xtest$}
        	\State Define decision threshold $\bigChange$ \LineComment{E.g.\ 130 bpm is an alarming resting heart rate}
        	\State Define ``$\eps$-near'' kernels $\Keps$, for $\eps > 0$ \LineComment{\cref{sec:nearbyKernels}}\label{line:Kchoice}
        	\State Solve \cref{mainOptimizationProblem} to get $k_1(\eps^*)$\LineComment{\cref{sec:nearbyKernels}}
        	\State Assess if $k_0$ and $\kperturb$ are qualitatively interchangeable.\LineComment{\cref{sec:QI}}
        	\If{$k_0$ and $\kperturb$ qualitatively interchangeable}
        		\Return ``$\Fstar$ is non-robust to the choice of kernel.''
        	\Else 
        		\, \Return ``Did not find that $\Fstar$ is non-robust to the choice of kernel.''
        	\EndIf
    \end{algorithmic}
\end{algorithm} 

\subsection{Nearby kernels and efficient optimization}
\label{sec:nearbyKernels}
We give two practical examples of how to choose $\Keps$ in the present work and detail how to solve \cref{mainOptimizationProblem} in each case. 
%In both cases, we assume we have prior information that our kernel $k$ is smooth. 
First, we consider the case where we assume $k \in \Keps$ should be stationary. Second, we allow non-stationary kernels $k \in \Keps$.

\textbf{Stationary kernels.}
Stationary kernels $k$ satisfy $k(\bx_n, \bx_m) := k(\tau)$, where $\tau := \bx_n - \bx_m$.
By Bochner's theorem, every stationary kernel can be represented by a positive measure \cite[Thm.\ 4.1]{RasmussenWilliams2006}.
In the kernel discovery literature, it is common to make the additional assumption that this measure has a density $S(\omega) =  \int e^{ -2\pi i \tau^T \omega} k(\tau) d\tau$ \citep{WilsonAdams2013_sm,Benton2019_function,Wilson2016_dkl}.
These authors show that the class of stationary kernels with a spectral density is a rich, flexible class of kernels.
So, we optimize over spectral densities $S(\omega)$ -- which are positive integrable functions on the reals -- to recover stationary kernels.
To make this optimization problem finite dimensional, we optimize the spectral density over a finite grid of frequencies $\omega$.
All of our examples here use 1-dimensional covariates, so we use the trapezoidal rule to recover $k$.
For our constraint set $\Keps$, we use an $\eps$ ball in the $\ell_\infty$ norm around the spectral density of $k_0$ for some $\eps > 0$.
We find this simple constraint set to be sufficient for the examples in this paper; however, if users have specific prior beliefs about how the spectral density of $k_1$ should be constrained, this $\Keps$ can be modified.
We summarize this constraint set and the resulting optimization objective in \cref{constraintSet:stationary}; see \cref{app:spectral_density} for more details, including selection of $\omega_1, \dots, \omega_G$. %

\begin{algorithm}[t]
    \caption{Objective and $\Keps$ for stationary kernels}
    \begin{algorithmic}[1]
    \label{constraintSet:stationary}
    \Statex \textbf{Objective}
    \State \textbf{Input}: Frequencies $\omega_1, \dots, \omega_G$, and density values $S(\omega_1), \dots, S(\omega_G)$.

       \State Approximate the integral $k(\tau) = \int e^{2\pi i \tau^T \omega} S(\omega) d\omega$ at $\tau$ needed to evaluate $\Fstar$ (e.g.\ trapezoidal rule).
       \State \Return{ $\Fstar(k)$.}
       \end{algorithmic}       
       \begin{algorithmic}[1]
       \Statex
       \Statex \textbf{Constraint on $S(\omega_1), \dots, S(\omega_G)$ defining $\Keps$}
       \State \textbf{Input}: Frequencies $\omega_1, \dots, \omega_G$, density values $S(\omega_1), \dots, S(\omega_G)$, constraint set size $\eps > 0$.
           \State Compute $S_0(\omega_1), \dots, S_0(\omega_G)$ (spectral density of $k_0$) via trapezoidal rule or exact formula.
    \If{Spectral density $S$ of $k$ satisfies:
       \begin{align*} 
       		\max \big( 0, (1-\eps) S_0(\omega_g) \big) 
       		 &\leq 
       		 	S(\omega_g)
       		 \leq 
       		 	(1+\eps) S_0(\omega_g), \\
       		 &\quad g = 1, \dots, G.
       \end{align*}
       }
       \textbf{return} ``In constraint set'' \\
       \textbf{Else} \textbf{return} ``Not in constraint set''
     \EndIf
    \end{algorithmic}
\end{algorithm} 

\textbf{Non-stationary kernels.}
In many modeling problems, stationarity may be a choice of convenience rather than prior conviction, or one may believe non-stationarity is probable. In either case, we wish to allow non-stationary kernels in the neighborhood $\Keps$. 
A convenient technique for constructing non-stationary kernels relies on input warping~\cite[Sec 4.2.3]{RasmussenWilliams2006}. Given a kernel $k_0$ and a non-linear mapping $g$, we define a perturbed kernel $k(\bx, \bx') = k_0(g(\bx), g(\bx'))$. 
This construction guarantees that the perturbed kernel $k$ is a kernel function as long as $k_0$ is a valid kernel. We let the function $g$ have parameters $w$ and set
$
g(\bx; w) := \bx + h(\bx; w), 
$
where $h: \R^D \rightarrow \R^D$ is a small neural network with weights $w$. 
By controlling the magnitude of $h$, we can control the deviations from $k_0$.

We could optimize the weights $w$ under the constraint $\|h(\bx;w)\|_2^2 \leq \eps$. 
However, it is unclear how to enforce this constraint.
Instead, we select a grid of $M$ points $\tilde{\bx}_1, \ldots, \tilde{\bx}_M \in \R^D$ and add a regularizer to our objective,
$
\frac{1}{\eps} \frac{1}{M}\sum_{m=1}^M ||h(\tilde{\bx}_m, w)||_2^2,
$
%
%\begin{equation}
%\hat{w} = \argmin_{w} \; \lvert F^{\star}(k(w)) - \Delta \rvert^{2} + \frac{1}{\eps} \frac{1}{m}\sum_i ||h(x_i, w)||_2 ,  
%\label{nonStationaryOpt}
%\end{equation}
%
where $\eps$ controls the regularization strength.
We find using a grid of points to be a computationally cheap, mathematically simple, and empirically successful approximation to regularizing the entire function.
We summarize our objective as a function of the network weights $w$  in \cref{constraintSet:nonStationary}.
Note that we have also changed our objective to include a generic loss $\ell$; some care needs to be taken to ensure that the optimal $k_1(\eps)$ is finite.
See \cref{sec:co2,sec:MNIST} for specific implementations of $\ell$.
Given the $\hat w$ minimizing the objective in \cref{constraintSet:nonStationary}, we set $k_1(\eps)(\bx, \bx') = k_0(g(\bx ; \hat w), g(\bx'; \hat w))$.
\begin{algorithm}
    \caption{Objective for non-stationary kernels}
    \begin{algorithmic}[1]
    \label{constraintSet:nonStationary}
    \State \textbf{Input:} Grid points $\tilde{\bx}_1, \dots, \tilde{\bx}_M$, regularizer strength $\eps > 0$, neural network weights $w \in \R^D$.
    \State Define neural network $h(x; w)$ with weights $w$.
    \State Define $k(\bx, \bx') := k_0(\bx + h(\bx; w), \bx' + h(\bx'; w))$.
    %\State \Return{$\lvert F^{\star}(k_1) - \Delta \rvert^{2} + \frac{1}{\eps} \frac{1}{M}\sum_{m=1}^M ||h(x_m; w)||_2^2$ }
    \State \Return{$\ell(k; \Fstar, \bigChange) + \frac{1}{\eps} \frac{1}{M}\sum_{m=1}^M ||h(\tilde{\bx}_m; w)||_2^2$ }
    \end{algorithmic}
   \label{alg:non-stationary}
\end{algorithm} 
\subsection{Assessing qualitative interchangeability}
\label{sec:QI}
We introduce two assessments, similar to prior predictive checks \citep{Gabry2019_bayesianViz}, to assess qualitative interchangeability between two kernels $k_0$ and $k_1$.

\textbf{Visual comparison of prior draws.}
When the covariates $\bx$ are low-dimensional, we can plot a small collection of functions drawn from each of the two distributions $\GP(0, k_{0})$ and $\GP(0, k_{1}).$ 
To ensure that visual differences between the priors are due to actual differences in the kernels and not randomness in the draws, we use \emph{noise-matched} prior draws. To define noise-matched draws, recall that one can draw from an $N$-dimensional Gaussian distribution $\mathcal{N}(0, \Sigma)$ by computing the Cholesky decomposition $LL^\top = \Sigma$;
if we draw $z \sim \mathcal{N}(0, I_N)$, we then have $Lz \sim \mathcal{N}(0, \Sigma)$.
We say that draws from two multivariate Gaussians are noise-matched if they use the same $z$.

If the user believes the two plots express the same qualitative information, the kernels are qualitatively interchangeable under this test.
A potential drawback to this method is that
when covariates are high-dimensional, it may be difficult to effectively visualize prior draws. 
We address this concern next.
%(2) The initial motivation for our paper was that some users may have difficulties expressing their prior beliefs; it might not be surprising, then, if some users may have difficulty in visually assessing prior beliefs. We address these concerns next.%To address these concerns, we provide a second test next.

\begin{comment}
\textbf{Gram matrix comparison.} 
No matter the dimension of the covariates, we can numerically compare the $GP(0,k_0)$ and $GP(0,k_1)$ priors on $\ftrainvec$.
In particular, these priors are multivariate normal distributions with covariance matrices equal to the Gram matrices $k_{0}(X,X)$ and $k_{1}(X,X)$, whose $(i,j)$ entries are, respectively, $k_{0}(\bx_{i}, \bx_{j})$ and $k_{1}(\bx_{i}, \bx_{j}).$
We will assess the difference between $k_0$ and $k_1$ by some distance $d(k_0(X,X), k_1(X,X))$.
Here, we use the 2-Wasserstein distance between the implied multivariate normals.
In \cref{app:histograms}, we note that other choices of $d$ are possible; however, we show that none of the results from any of our experiments are sensitive to this choice.
\end{comment}
\textbf{Comparison through Wasserstein distances.}
Our second test for qualitative interchangeability computes a distance between the priors $\GP(0,k_0)$ and $\GP(0, k_1)$ and uses hyperparameter uncertainty in $k_0'$ to help users understand whether this distance is large.
%No matter the dimension of the covariates, we can numerically compare the $GP(0,k_0')$ and $GP(0,k_1')$ priors, where we recall that $k_0'(\bx_i, \bx_j)+ \hat\sigma^2 \delta_{ij}$, and likewise for $k_1'$.
Although directly computing distances between Gaussian processes is difficult, we can compare the $\GP(0,k_0')$ and $\GP(0,k_1')$ priors on the set $\{\bx_1, \dots, \bx_N\}$.
On this set, these priors are just multivariate normal distributions with covariance matrices equal to the Gram matrices $k_{0}(X,X) + \hat\sigma^2 I_N$ and $k_{1}(X,X) + \hat\sigma^2 I_N$, where $X \in \R^{N \times D}$ is the matrix of covariates.
%The 2-Wasserstein distance between multivariate normal distributions is efficient to compute, well defined as $N \rightarrow \infty$, and converges to the Wasserstein metric between $\GP(0, k_{0})$ and $\GP(0, k_{1})$~\citep[Theorem 8]{Mallasto17}. 
Going forward, we denote by $d(k_0', k_1')$ the 2-Wasserstein distance between these multivariate normals.
We use the 2-Wasserstein as a default choice because it directly corresponds to quantities easily interpretable by users: a 2-Wasserstein distance of $\alpha$ means that coordinate-wise standard deviations differ by at most $\alpha$  \citep[Thm.\ 3.4][]{huggins_2020_validated}.
%That is, our assessment asks whether the coordinate-wise differences in standard deviations of $k_0'$ and $k_1'$ on the training data are substantially different than what we might expect under sampling variability in the MMLE $\hat\theta$.
While we feel the 2-Wasserstein distance provides a good default choice of $d$, users can substitute other choices if there are application-specific reasons why another distance is more meaningful.
In \cref{app:histograms}, we discuss this possibility and also show that our results are typically not sensitive to the choice of $d$ throughout our experiments.

Even though users may be able to understand the meaning of $d(k_0', k_1')$, users may still have difficulty understanding whether $d(k_0', k_1')$ is large or not.
In particular, users may not have an understanding of the scale of $d$.
To help users understand the scale of $d$, we use a particular form of uncertainty about $k_{0}'.$
Since we learn the hyperparameters $\hat{\theta}$ of $k_{0}'$ from finite data, there remains frequentist sampling uncertainty about $\hat{\theta}$, which we denote by the distribution $q(\hat\theta).$
We make $R$ i.i.d.\ draws $\{\theta^{(r)}\}_{r=1}^R$ from $q(\hat\theta)$ (or an approximation of $q$).
For each $r$, we compute $d(k_0', k^{'(r)})$, where $k^{'(r)}$ has the same functional form as $k_0'$ but with hyperparameters $\theta^{(r)}$ instead of $\hat\theta$.
To the extent that bootstrap resamples capture frequentist sampling variability, users should be as open to using most $k^{'(r)}$ as they are to using $k_0’$.
Thus if $d(k_{0}', k_{1}')$ is small relative to the $d(k_{0}', k^{'(r)})$'s, we say that $k_0$ and $k_1$ are qualitatively interchangeable.
Note that we cannot necessarily reject qualitative interchangeability if $d(k_0', k_1')$ is large relative to the $d(k_0', k^{'(r)})$.
For example, if we observe more and more data $(\bx, y)$ with the $\bx$'s contained in a compact region of $\R^D$, then we expect our uncertainty about the hyperparameters to go to zero.
Thus, for fixed $k_1'$, $d(k_0', k_1')$ will eventually always be large relative to the $d(k_0', k^{'(r)})$.

In our experiments, we make the following choices. 
Unless otherwise stated, we approximately sample from $q(\hat\theta)$ by drawing bootstrap samples from $\{(\bx_n, y_n)\}_{n=1}^N$ and re-solving \cref{eq:MMLE} with the bootstrapped data. %And we compare Gram matrices using a relative Frobenius norm: namely, $\| k_1(X,X) - k_0(X,X) \|_{F} / \|k_0(X,X)\|_F$, where $\|\cdot \|_F$ is the Frobenius norm. 
We construct a histogram of the 2-Wasserstein distance between $k^{'(r)}$ and $k_0'$ across $r$, with a marker indicating the position of the 2-Wasserstein between $k_1'$ and $k_0'$. If the marker lies to the left of or within the histogram, we conclude that $k_1$ and $k_0$ are qualitatively interchangeable.
\subsection{Workflow illustration on synthetic data}
\label{sec:syntheticExample}

\begin{figure*}[t]
%\begin{subfigure}[b]{0.245\textwidth}
\includegraphics[width = 0.245\textwidth]{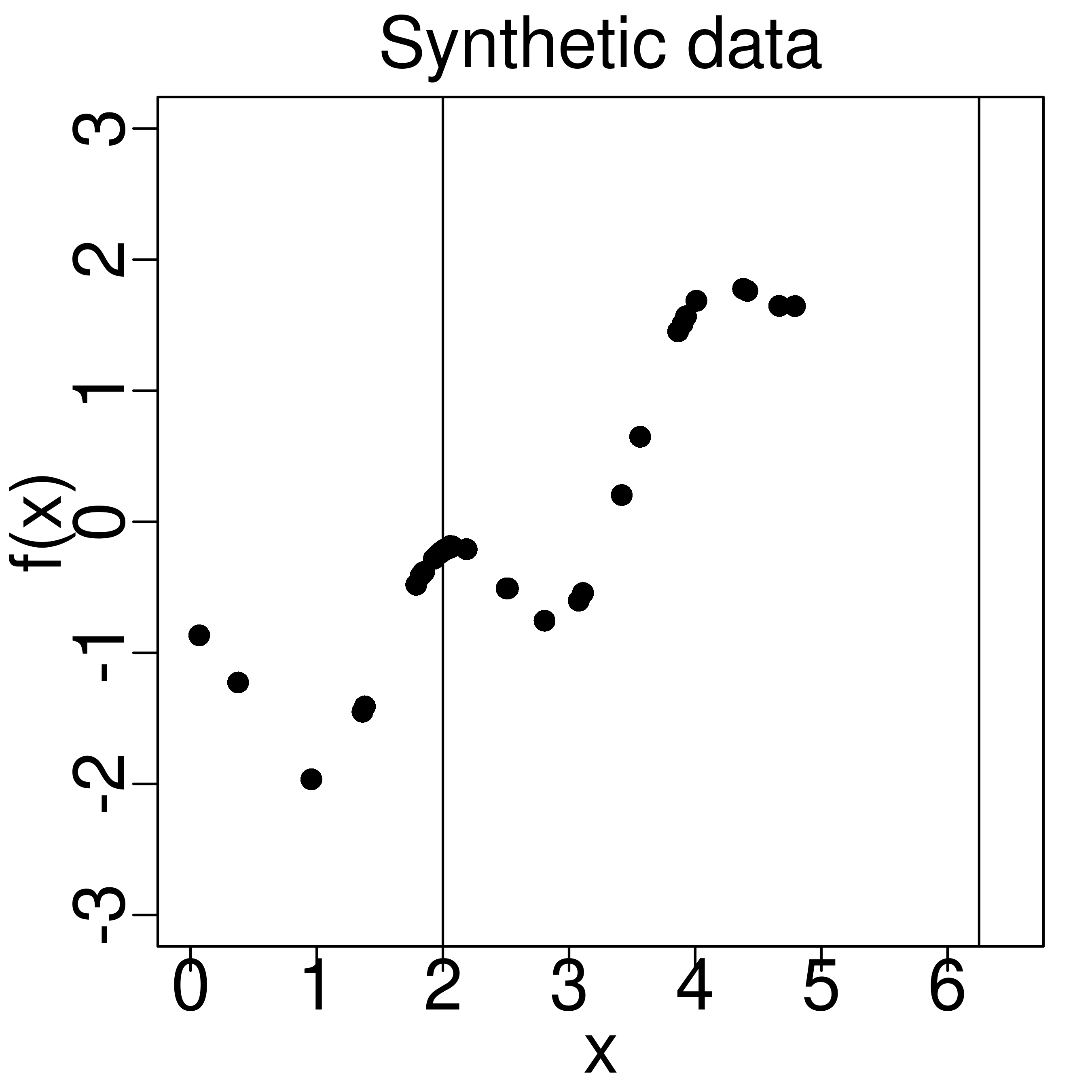}
%\caption{}
%\label{fig:synthetic_data}
%\end{subfigure}
%\begin{subfigure}[b]{0.245\textwidth}
\includegraphics[width = 0.245\textwidth]{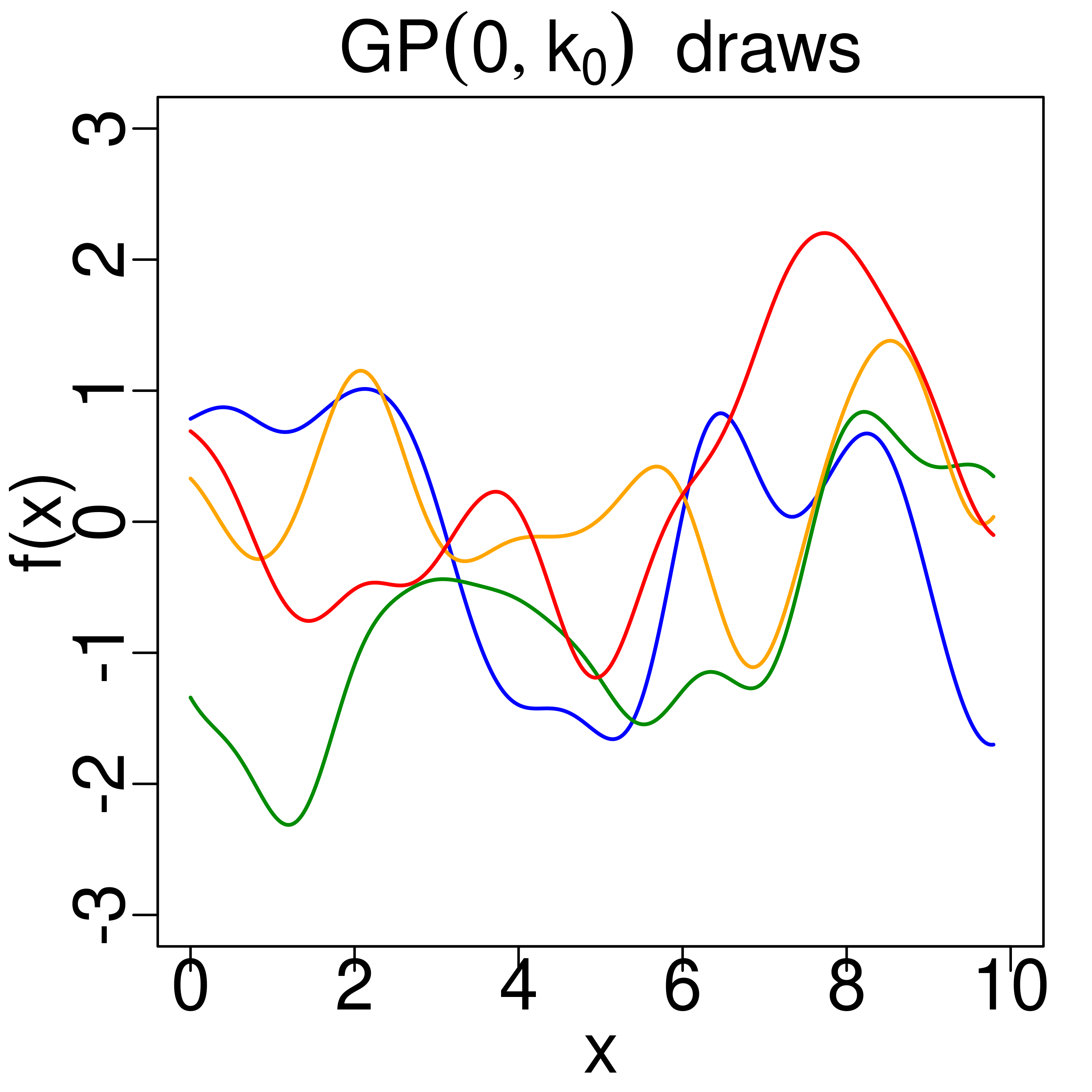}
%\caption{}
%\label{fig:synthetic_prior0}
%\end{subfigure}
%\begin{subfigure}[b]{0.3\textwidth}
%\includegraphics[width = \textwidth]{figures/synthetic_noise=0.010/syntheticExample_changeVsEpsilon}
%\caption{}
%\label{fig:synthetic_changeVsEpsilon}
%\end{subfigure}
%\begin{subfigure}[b]{0.245\textwidth}
%\includegraphics[width = \textwidth]{figures/synthetic_noise=0.010/syntheticExample_prior1_extrapolation.png}
\includegraphics[width = 0.245\textwidth]{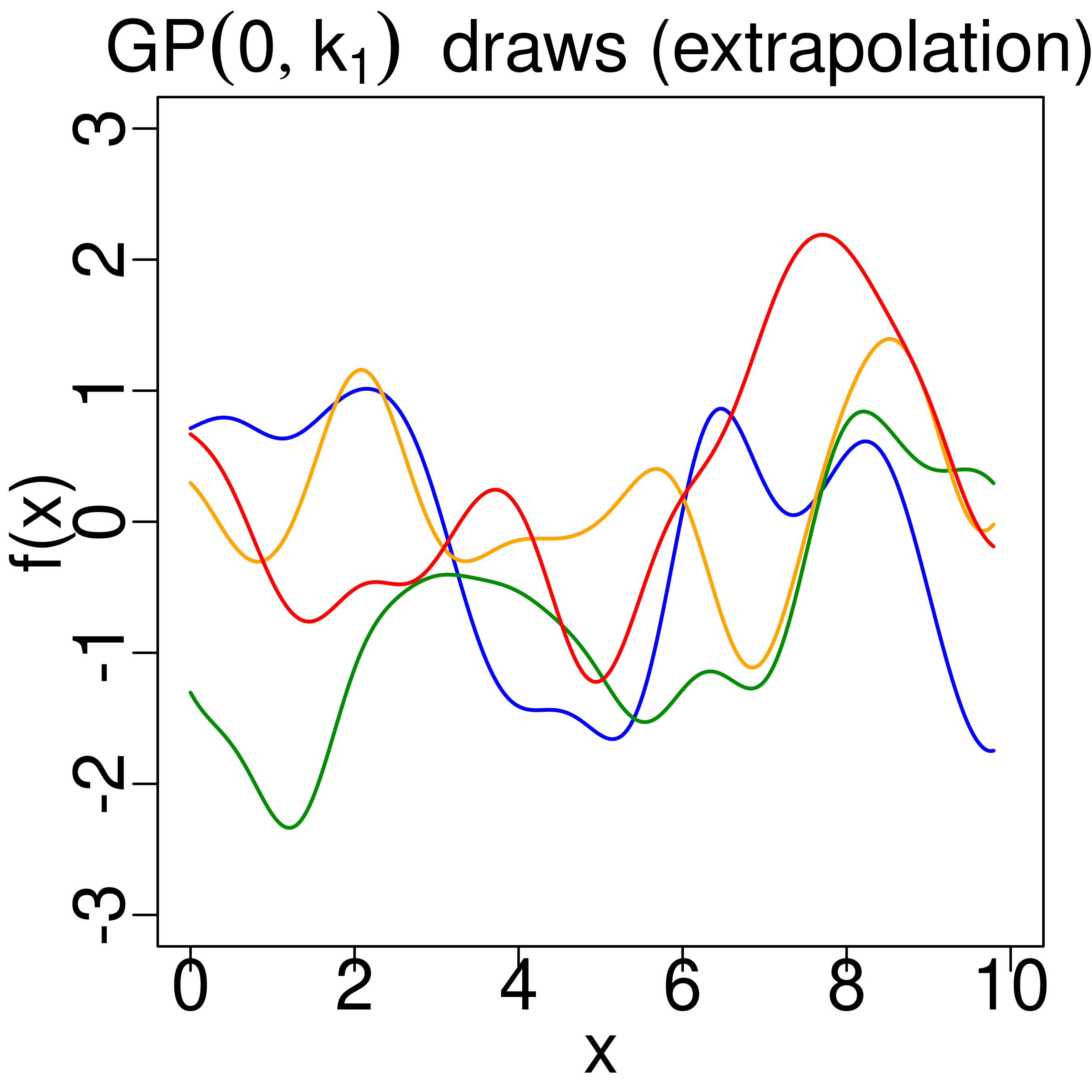}
%\caption{}
%\label{fig:synthetic_prior1_extrapolation}
%\end{subfigure}
%\begin{subfigure}[b]{0.245\textwidth}
%\caption{}
%\label{fig:synthetic_prior1_interpolation}
%\end{subfigure}
\includegraphics[width = 0.245\textwidth]{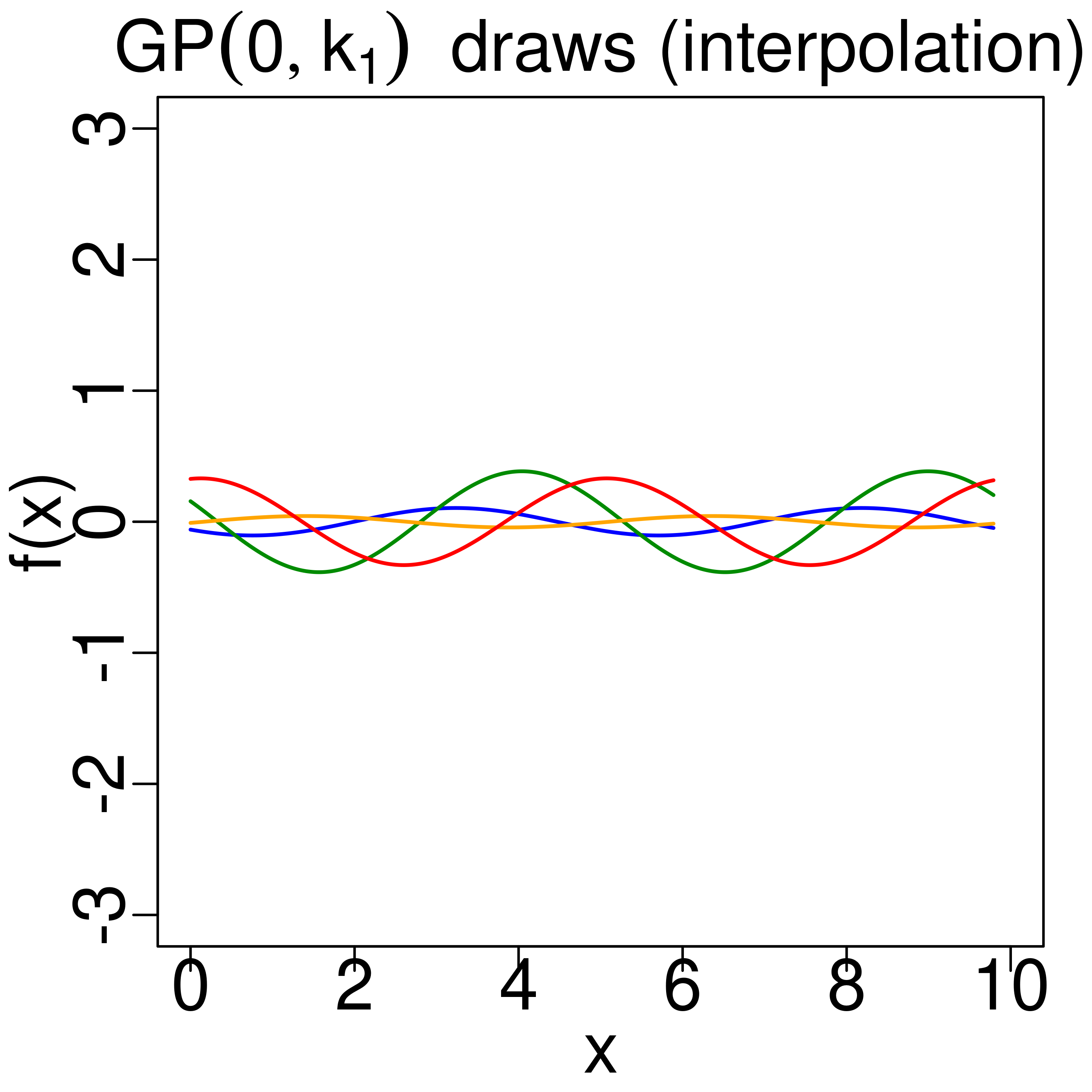}
\caption{\small{\emph{(Far left):} Synthetic data. Vertical lines denote our interpolation point ($\xtest = 2.0$) and extrapolation point ($\xtest = 6.25$). (\emph{Center left:}) Draws from the original prior $\GP(0, k_{0})$. (\emph{Center right}): Draws from $\GP(0,\kextrap)$. (\emph{Far right}): Draws from $\GP(0,\kinterp)$. 
The prior draws are noise-matched to the draws from $k_0$ (\cref{sec:QI}). %The gray area in the right three plots is the 99.7\% uncertainty interval for the original prior ($\GP(0,k_{0})$). 
}}
\label{fig:syntheticData_all}
\end{figure*}

\textbf{Data and decision.}
Before turning to real data, we illustrate our workflow with a synthetic-data example. 
We consider $N=24$ data points with a single covariate; see the leftmost panel of \cref{fig:syntheticData_all}.
We assume we have qualitative prior beliefs that (i) $f$ is smooth and (ii) our beliefs about $f$ are invariant to translation along the single covariate (stationarity). In this case a plausible kernel choice is a squared exponential kernel: $k_0(x_1, x_2) = \exp\big( -0.5 (x_1 - x_2)^2 / \ell\big)$.
%Specifically, we use the kernel from \citet{Colopy2016_identifying} (see \cref{heartRateKernel} of \cref{app:heartRate}), which we experiment with further in \cref{sec:heartRate}.
We estimate $k_0$'s hyperparameter $\ell$ and the noise parameter $\sigma$ via maximum marginal likelihood estimation (MMLE).
Four draws from the resulting prior are shown in the second panel of \cref{fig:syntheticData_all}.

For the purposes of this illustration, we will look at two separate decisions. One is at $\xtest = 2.0,$ which is within a dense region of training data (interpolation). And one is at $\xtest = 6.25,$ which is outside the range of the training data (extrapolation).
Our functional of interest at either point will be the change in posterior mean, $\Fstar(k) := \mu(\xtest, k) - \mu(\xtest,k_0)$, where $\mu(\xtest,k)$ is the posterior mean at test point $x$ under kernel $k$.
%$$
%\Fstar(k) := \frac{\mu(\xtest, k_0) - \mu(\xtest,k)}{\sigma(\xtest, k_{0})},
%$$
%where $\mu(\xtest, k)$ and $\sigma(\xtest,k)$ are the posterior mean and standard deviation, respectively, of $f(\xtest)$ corresponding to kernel $k.$
We suppose that we would make a different decision if the posterior mean changed by a small amount: $\Fstar \geq \bigChange = 0.01$.
Intuitively, we expect only minor changes to the prior to be needed to bring about such a posterior change in our extrapolation example.
In our interpolation example, we expect a substantial change to the prior to be needed to change the posterior even a small amount, as we have intentionally chosen our interpolation point to sit in a region of dense training data.
We now show that the output of our method matches this intuition in both cases.

\begin{figure*}[h]
\centering
	\includegraphics[width = 0.3\textwidth]{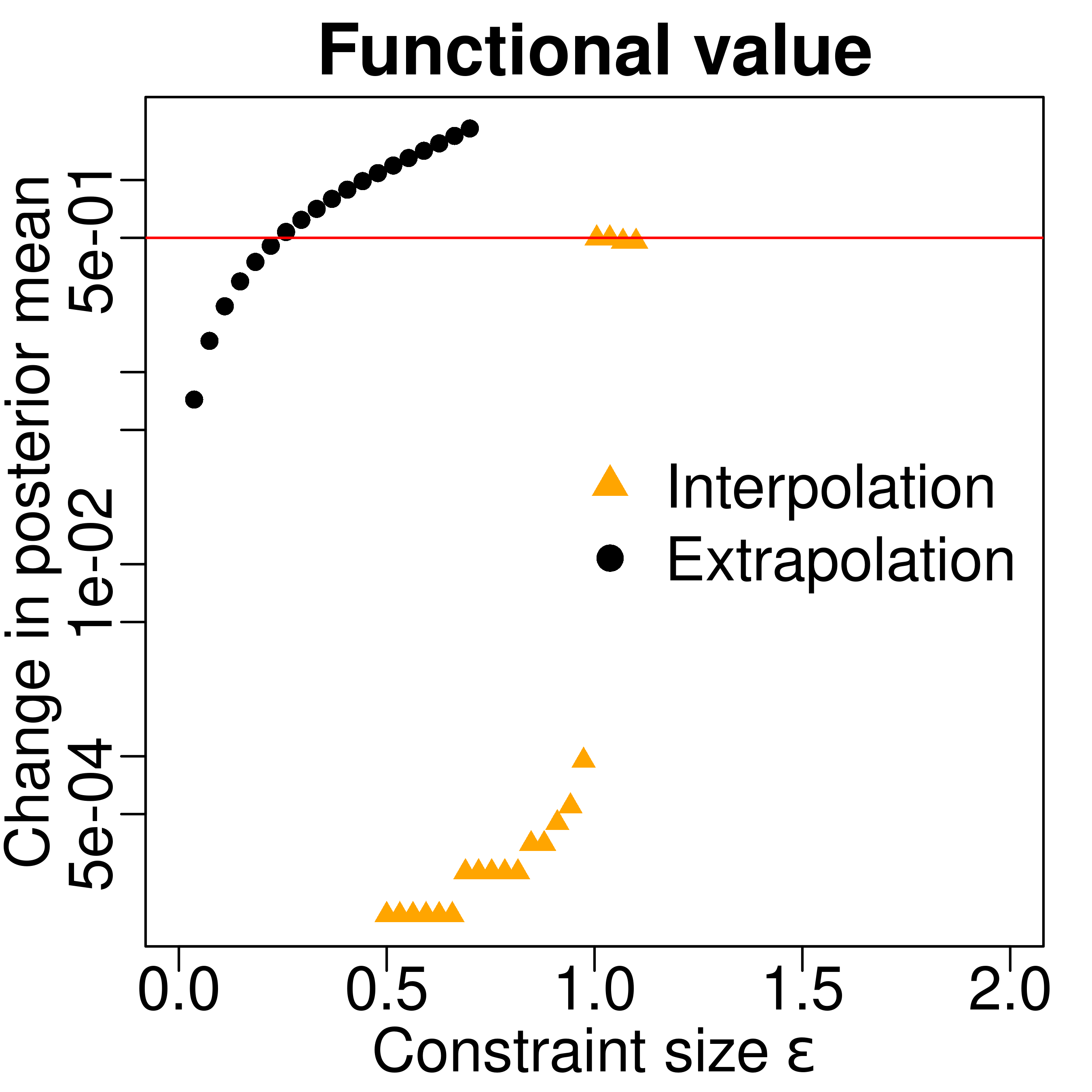}
\includegraphics[width = 0.31\textwidth]{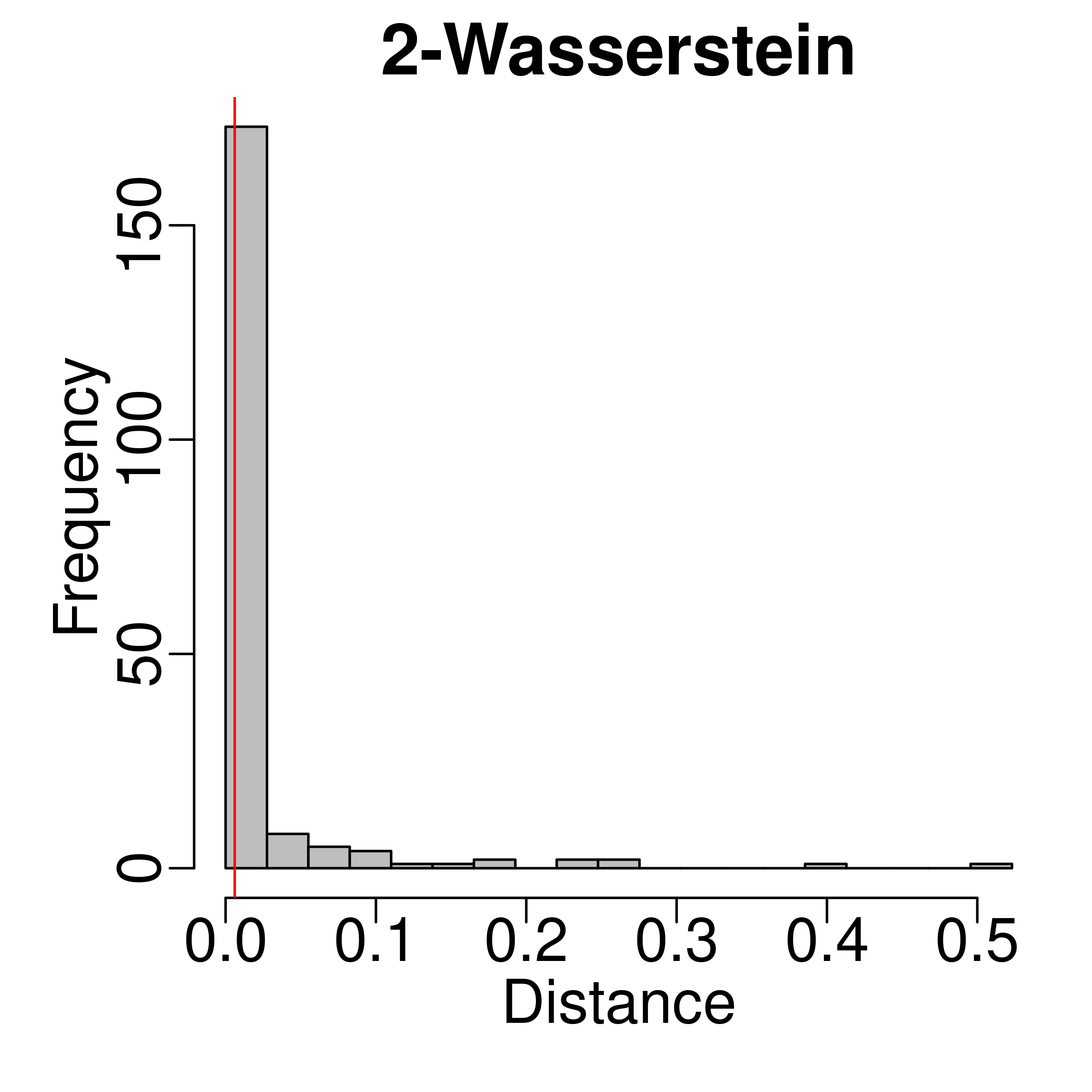}
\includegraphics[width = 0.31\textwidth]{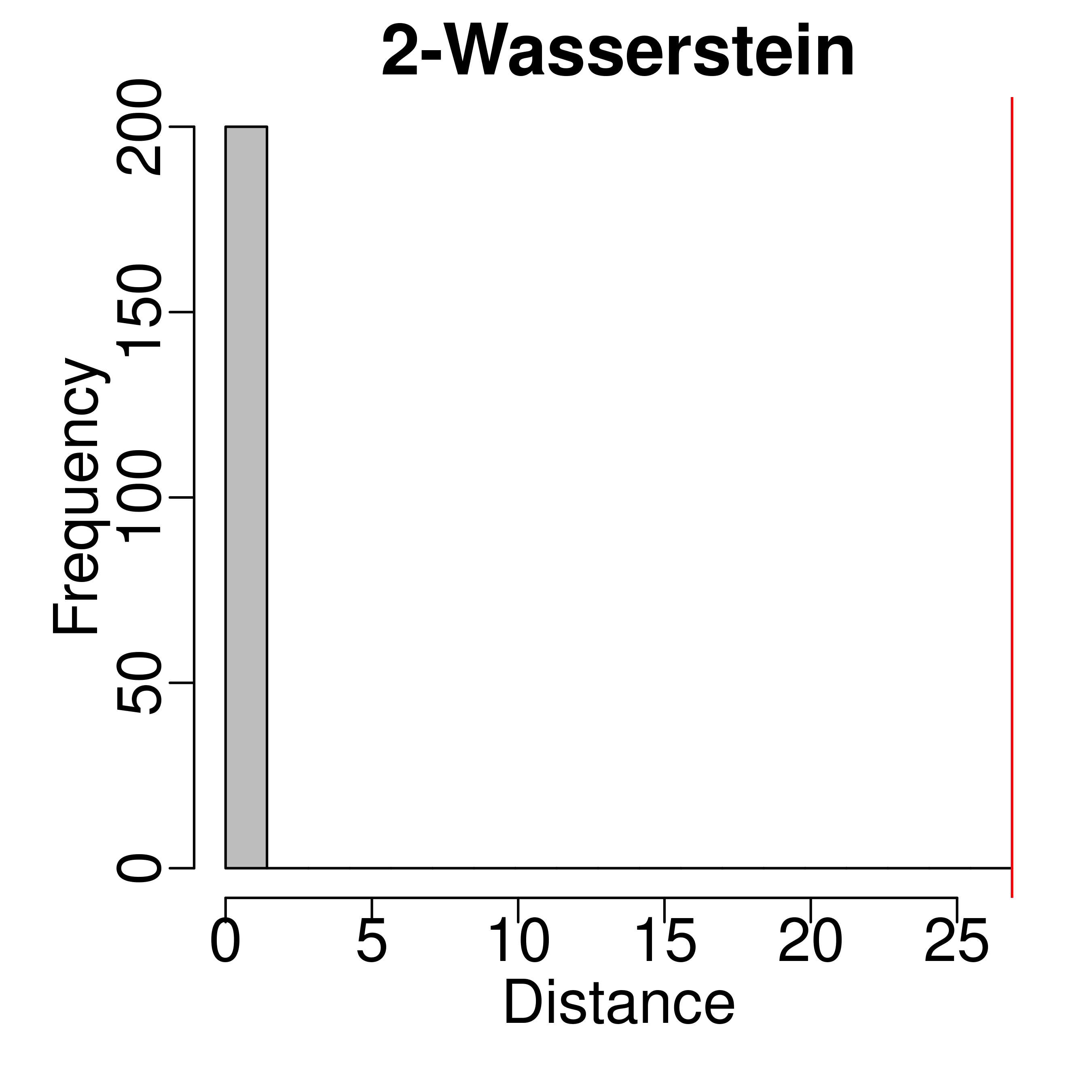}
	\caption{\small{(\emph{Left}): Maximal value of the function $\Fstar$ as a function of constraint set $\varepsilon.$ Comparison of the 2-Wasserstein distance between $k_{0}(X,X)$ and $k_{1}(X,X)$ to the posterior variation due to hyperparameter uncertainty for extrapolation (\emph{middle}) and interpolation (\emph{right}). The red line corresponds to our decision-changing kernel $k_1$.}}
	\label{fig:syntheticHistograms}
\end{figure*}

\textbf{Nearby kernels.}
Since we assume stationarity, we choose \cref{constraintSet:stationary} at line~\ref{line:Kchoice} of \cref{alg:workflow}. 
\cref{fig:syntheticHistograms} (first panel) shows what happens as we increase $\eps$ to solve \cref{mainOptimizationProblem}. The black dots (extrapolation) quickly cross the decision threshold line, so we have solved \cref{mainOptimizationProblem}. The orange triangles (interpolation) show that a larger $\eps$ is required to breach our decision threshold.

\textbf{Qualitative interchangeability: Visual comparison of prior draws.} We now demonstrate our first test for qualitative interchangeability.
For our extrapolation example ($\xtest = 6.25$), let $\kextrap$ be the solution to \cref{mainOptimizationProblem}. 
The third panel of \cref{fig:syntheticData_all} shows prior draws with $\kextrap$; the draws are noise-matched with the second panel. 
$\kextrap$ is so similar to $k_0$ that the two sets of prior draws (second and third panels) are visually indistinguishable.
We say that $\kextrap$ is qualitatively interchangeable with $k_0$.
%Both are smooth and stationary by constraint; the length scale (cf.\ the number of ``wiggles'') and amplitudes seem unchanged. 

Let $\kinterp$ be the solution to \cref{mainOptimizationProblem} for our interpolation example ($\xtest = 2.0$).
%Since even the largest constraint size value considered ($\eps = 10$) is not enough to induce the required change, we define $\kinterp$ for our interpolation example ($\xtest= 2.00$) to be $k_1(10)$ from \cref{mainOptimizationProblem}.
The fourth panel of \cref{fig:syntheticData_all} shows prior draws with $\kinterp$; the draws are noise-matched with the second panel. 
Again, by design, both sets of draws (second and fourth panels) are stationary and smooth. 
However, the magnitudes of peaks and troughs with $\kinterp$ are much smaller than those with $k_0$.
Thus, we say that $\kinterp$ is not qualitatively interchangeable with $k_0$ under this test.

\textbf{Qualitative interchangeability: 2-Wasserstein comparison.} 
Histograms of the 2-Wasserstein distance between $k_{0}$ and $k^{(r)}$ appear in \cref{fig:syntheticHistograms}.
$\kextrap$ sits within the histogram of alternative kernels generated via hyperparameter uncertainty (center), whereas $\kinterp$ sits outside this uncertainty region (right). 
As in our prior visualization comparison, we say that $\kinterp$ is not qualitatively interchangeable with $k_0$ under our 2-Wasserstein comparison, whereas $\kextrap$ is.

Finally, following our workflow, we conclude that our extrapolation example is non-robust to the choice of kernel in the sense of \cref{def:nonRobust}. On the other hand, in our interpolation example, we do not find non-robustness. 

\section{STATIONARY PERTURBATIONS TO A MODEL OF HEART RATES}
%!TEX root = neurips_2021.tex
\label{sec:heartRate}
We now provide an example of using our workflow to assess the sensitivity of GP predictions of hospital patient deterioration. \citet{Colopy2016_identifying} use a GP to model individual patients' heart rates and predict potentially troubling behavior at a future time $\xtest$. We check whether this prediction is robust to kernel choice.

%an application of Gaussian processes to predicting deteriorating patients in an intensive care unit (ICU).
%\cite{Colopy2016_identifying} use a GP to model individual patients' heart rates over time. Given a patient's heart rate observed up to time $t$, the GP posterior distribution at a later time $t'$ can be used to predict whether the patient's heart rate will display troubling behavior at time $t'$.
%We use our framework to understand whether such a prediction is sensitive to the choice of kernel.
%
\begin{figure*}

%	\begin{tabular}{ccc}
%		\includegraphics[scale=0.3]{figures/heartRate_sensitive_data.png} &
%		\includegraphics[scale=0.3]{figures/heartRate_sensitive_prior0.png} & 
%		\includegraphics[scale=0.3]{figures/heartRate_sensitive_epsilonPlot.png} \\ 
%		\includegraphics[scale=0.3]{figures/heartRate_sensitive_prior1.png} &
%		\includegraphics[scale=0.3]{figures/heartRate_sensitive_posterior.png} &
%		\includegraphics[scale=0.3]{figures/heartRate_sensitive_histogram.png}
%	\end{tabular}
\includegraphics[width = 0.325\textwidth]{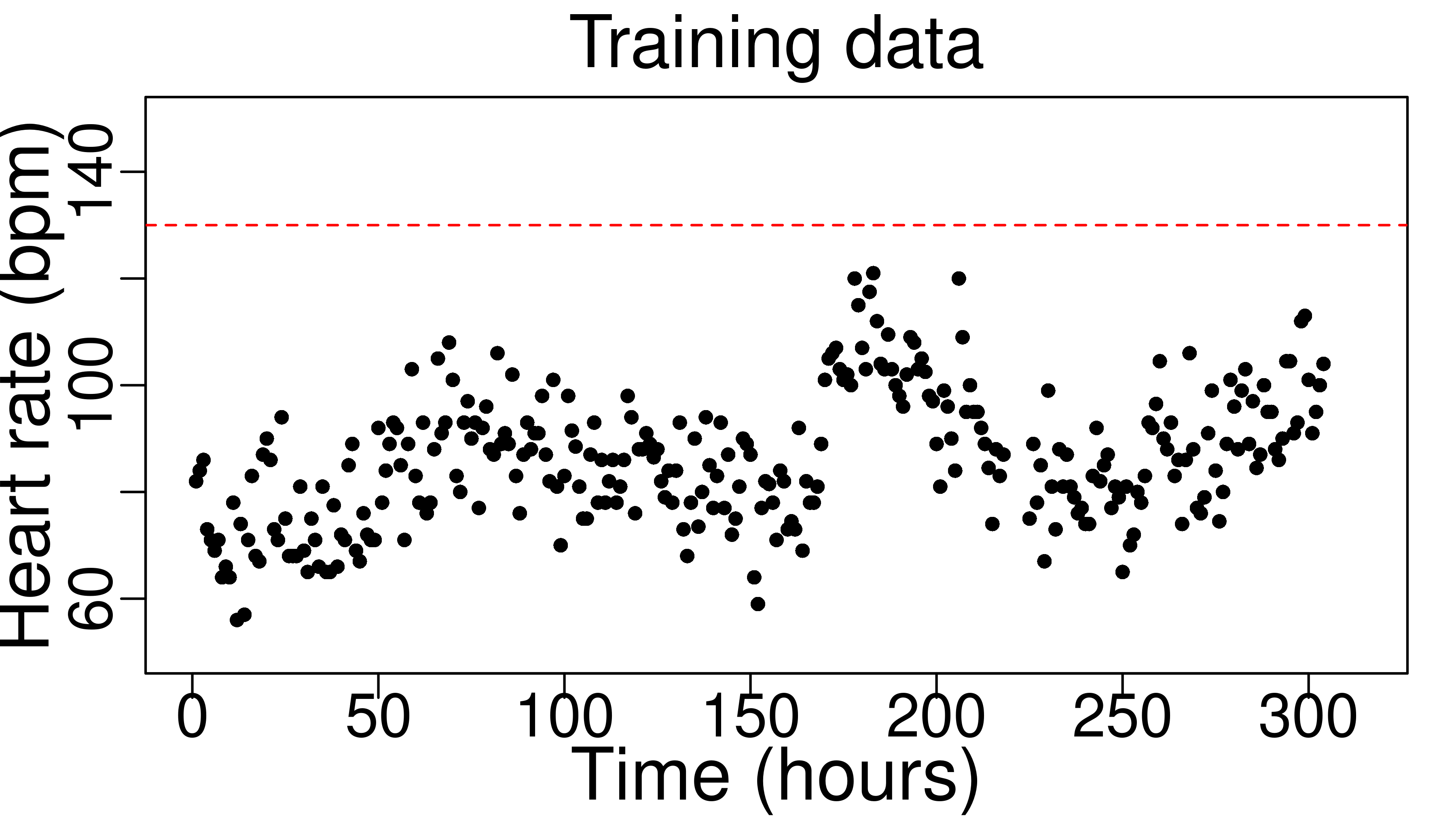} 
\includegraphics[width = 0.325\textwidth]{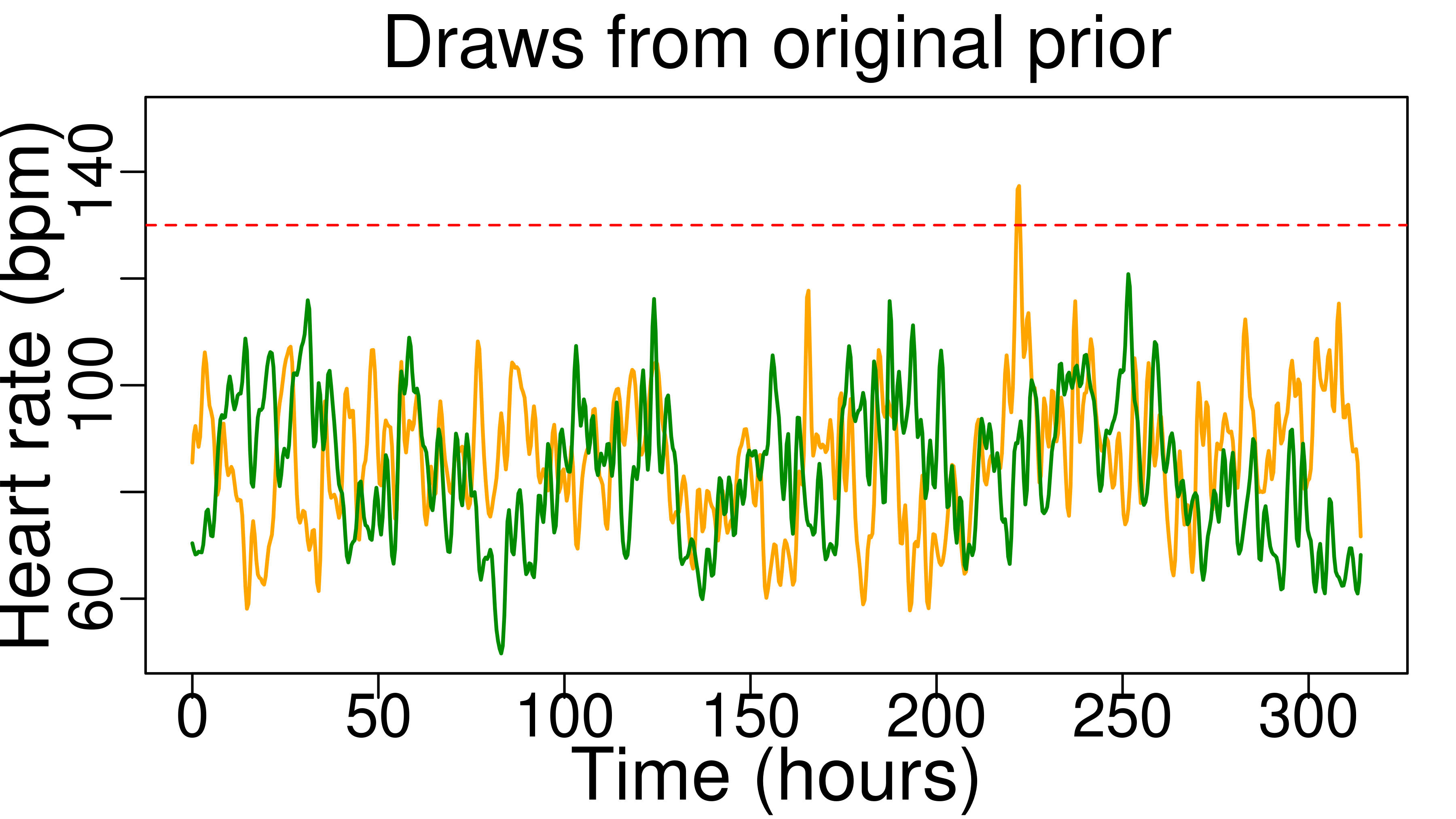} 
\includegraphics[width = 0.325\textwidth]{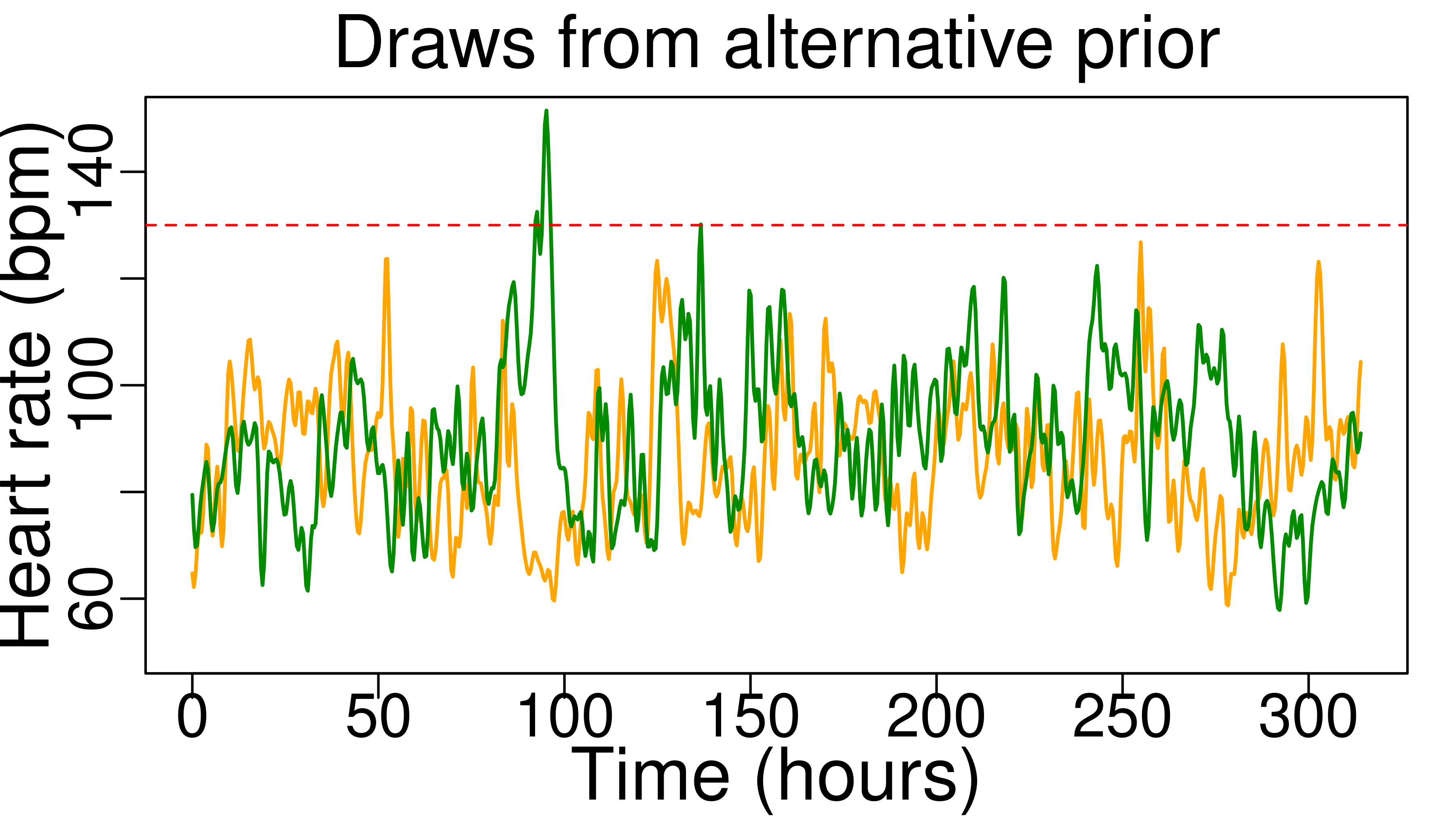} 

\includegraphics[width = 0.325\textwidth]{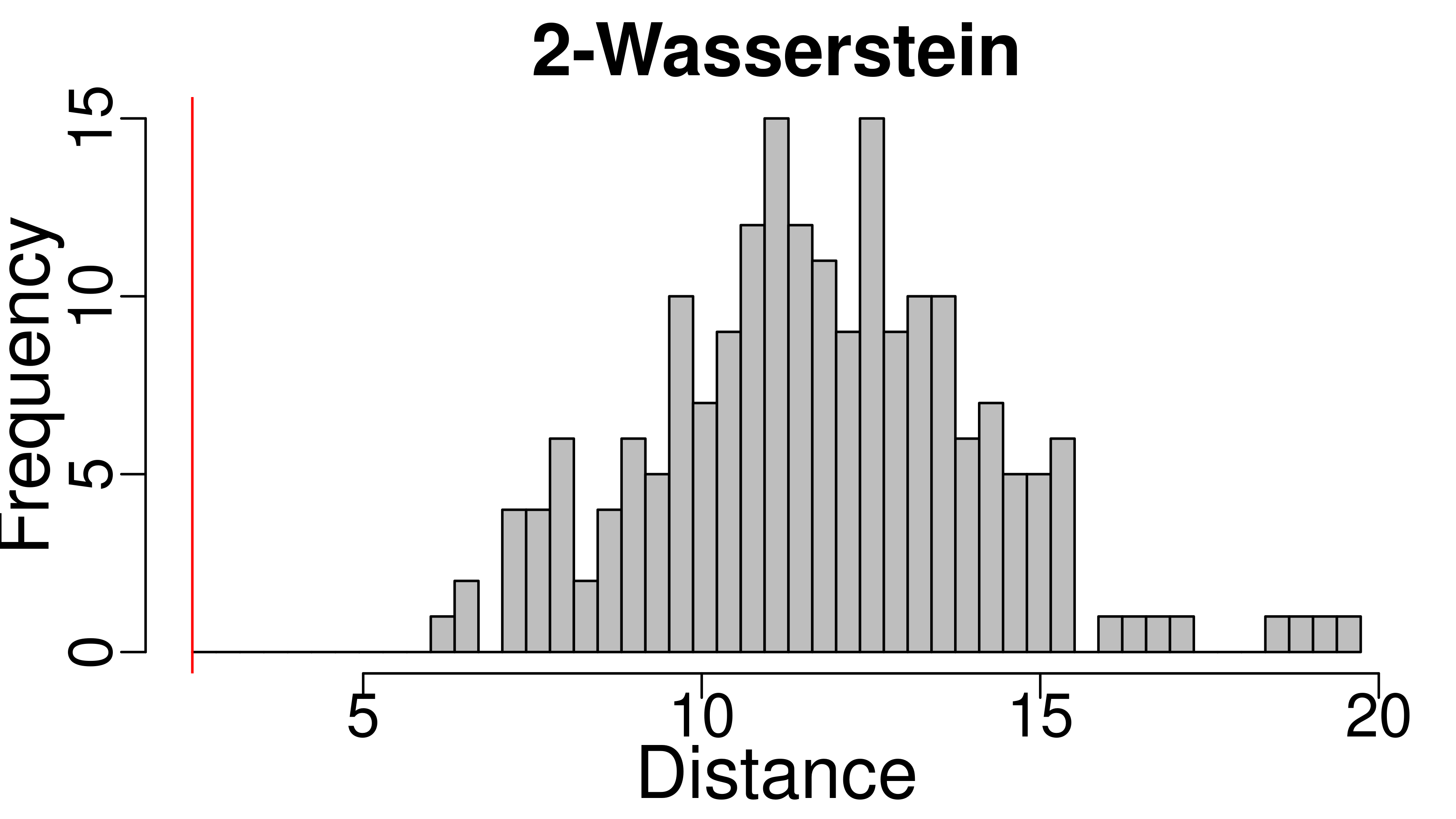}
\includegraphics[width = 0.325\textwidth]{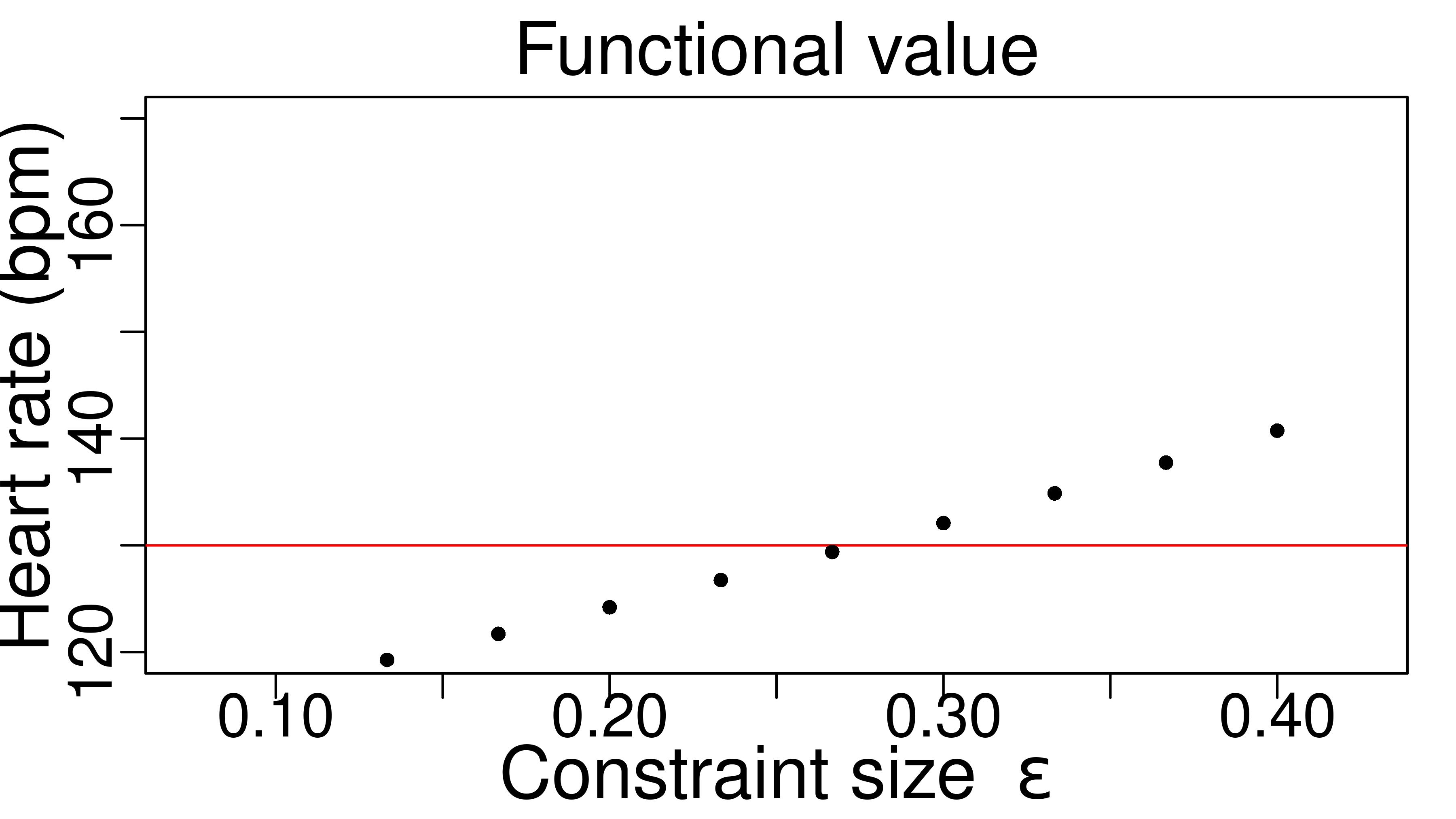} 
\includegraphics[width = 0.325\textwidth]{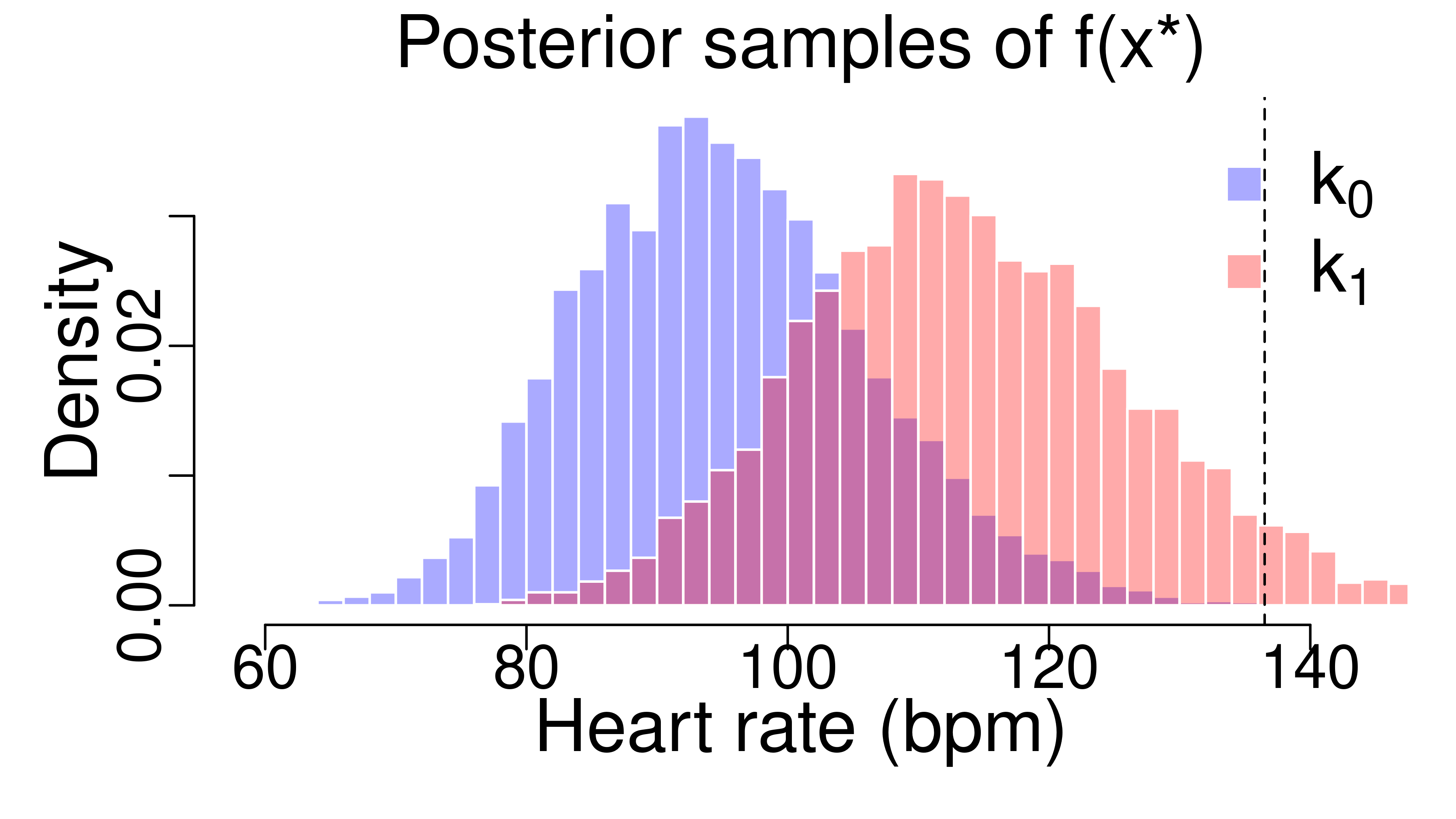}
\caption{\small{Sensitivity of heart rate analysis in \cref{sec:heartRate}. 
(\emph{Top row}): (\emph{left}) Observed data.
(\emph{middle and right}) Noise-matched draws from original prior $\GP(0,k_{0})$ (middle) and alternative prior $\GP(0, k_1)$ (right).
(\emph{Bottom row}): (\emph{left}) Comparison of the difference between $k_{0}$ and $k_{1}$ (red line) to bootstrapped hyperparameter uncertainty (histogram). 
(\emph{middle}) Once we expand the constraint set to $\eps = 0.24$ the predicted 95\% quantile of heart rate at $\xtest$ exceeds 130 bpm (red line). (\emph{right}) Comparison of posterior distributions $f(\xtest) \mid \Data$ computed using $k_{0}$ (blue) and $\kperturb$ (red)}.
}
	\label{fig:heartRate}
\end{figure*}

\textbf{Data, model, and decision.} \cite{Colopy2016_identifying} observe an outcome, heart-rate data measured in beats per minute (bpm), as a function of one covariate, time. The authors choose their GP model to have mean equal to zero and a kernel equal to the sum of a squared exponential and Matérn(5/2) kernel; see \cref{app:heartRate}.
We fit the overall kernel's hyperparameters via MMLE and refer to the resulting kernel as $k_0$.
Some standard hospital alarm systems activate at 130 bpm \citep{Fidler2017_heartRateThreshold}, a threshold describing a worryingly-high resting heart rate. So we consider the task of predicting whether the 95th percentile of the GP posterior is above $\bigChange = 130$ bpm.	
Most predictions in \cite{Colopy2016_identifying} take place 1.5 hours in the future, so we set $\Fstar$ to be the 95th quantile at 1.5 hours hours after the last observed data point.
The data from \cite{Colopy2016_identifying} is confidential, so we use heart-rate data from the 2019 Computing in Cardiology Challenge \citep{compCardiology1,compCardiology2}.

\textbf{Prior beliefs.} \cite{Colopy2016_identifying} note that $k_0$ encodes the belief that ``longer
trends (on the order of hours) are governed by the
smooth RBF kernel, while minutely variations in [heart-rate] are
governed by a twice-differentiable Matérn(5/2) kernel.'' 
Although \cite{Colopy2016_identifying} are not explicit about assuming stationarity, we presume it is a reasonable prior belief here; while we expect that a patient's heart rate may change while in the hospital, our prior beliefs about the timing of any changes may be roughly uniform. 
We thus choose $\Keps$ according to the stationary specification in \cref{sec:nearbyKernels}.

\textbf{Robustness.} \cref{fig:heartRate} depicts our workflow (\cref{alg:workflow}) in action. 
We solve \cref{mainOptimizationProblem} to obtain $\kperturb$ such that $\Fstar(\kperturb) \geq L$.
We then compare noise-matched samples from the priors using $k_0$ and $\kperturb$. 
The noise-matched samples do not clearly represent different pieces of prior information; both prior plots display functions that are fairly rough with similar length scales.
Finally, we see that the 2-Wasserstein distance between $k_{0}$ and $\kperturb$ is substantially smaller than the 2-Wasserstein distance between $k_{0}$ and kernels from the sampling uncertainty in the MMLE hyperparameters.	
Our tests suggest $k_0$ and $\kperturb$ are qualitatively interchangeable; we conclude that the prediction that $\Fstar$ will breach the alarm threshold is non-robust in the sense of \cref{def:nonRobust}.
While this outcome may be surprising, it is not entirely unintuitive.
The patient's heart rate is trending up toward the end of the observed data.
In \cref{app:heartRate}, we show an example where the observed data is trending downward at the end of the observed data.
In the latter case, the resulting kernel $\kperturb$ fails both of our tests of qualitative interchangeability and so we cannot conclude non-robustness.

\section{NON-STATIONARY PERTURBATIONS TO A MODEL OF \co2 LEVELS}
\label{sec:co2}
\begin{figure*}
\centering
%	\begin{tabular}{lll}
		\includegraphics[width=0.485\textwidth]{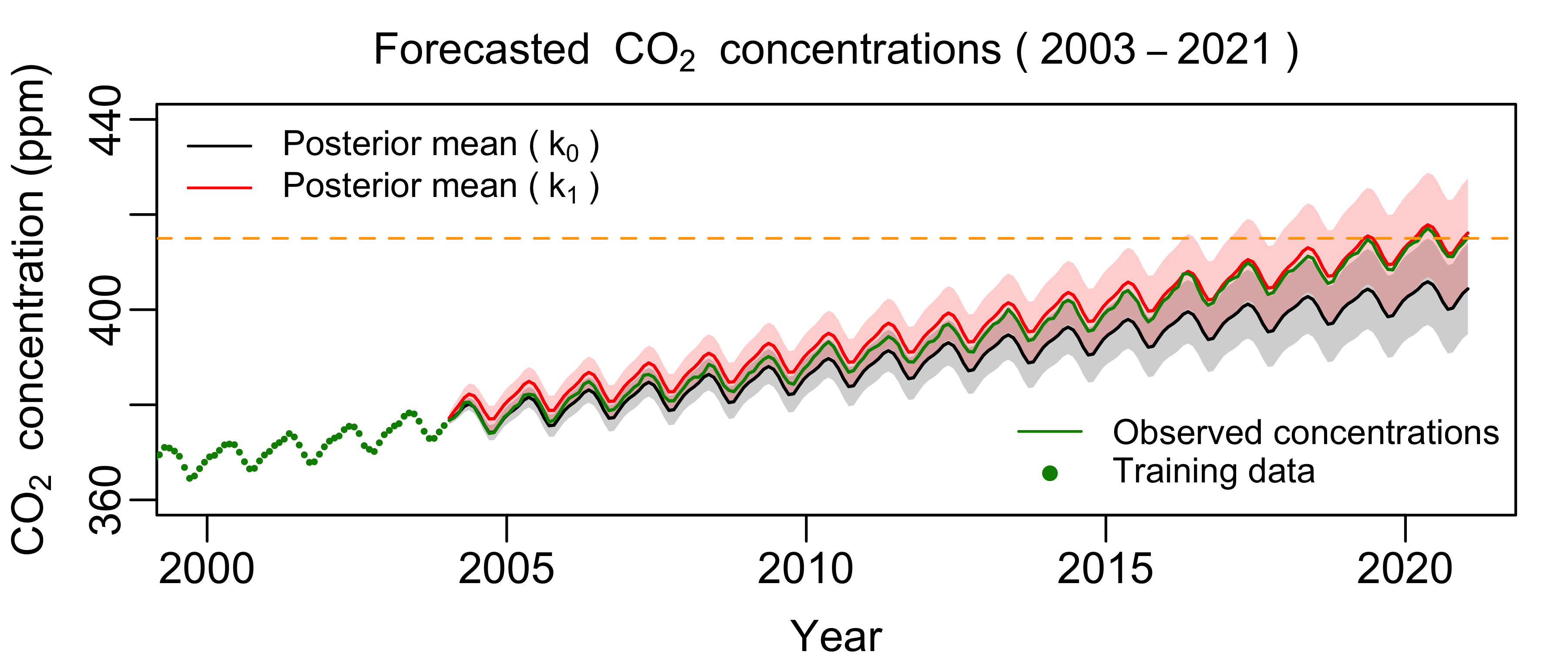}
		\includegraphics[width = 0.485\textwidth]{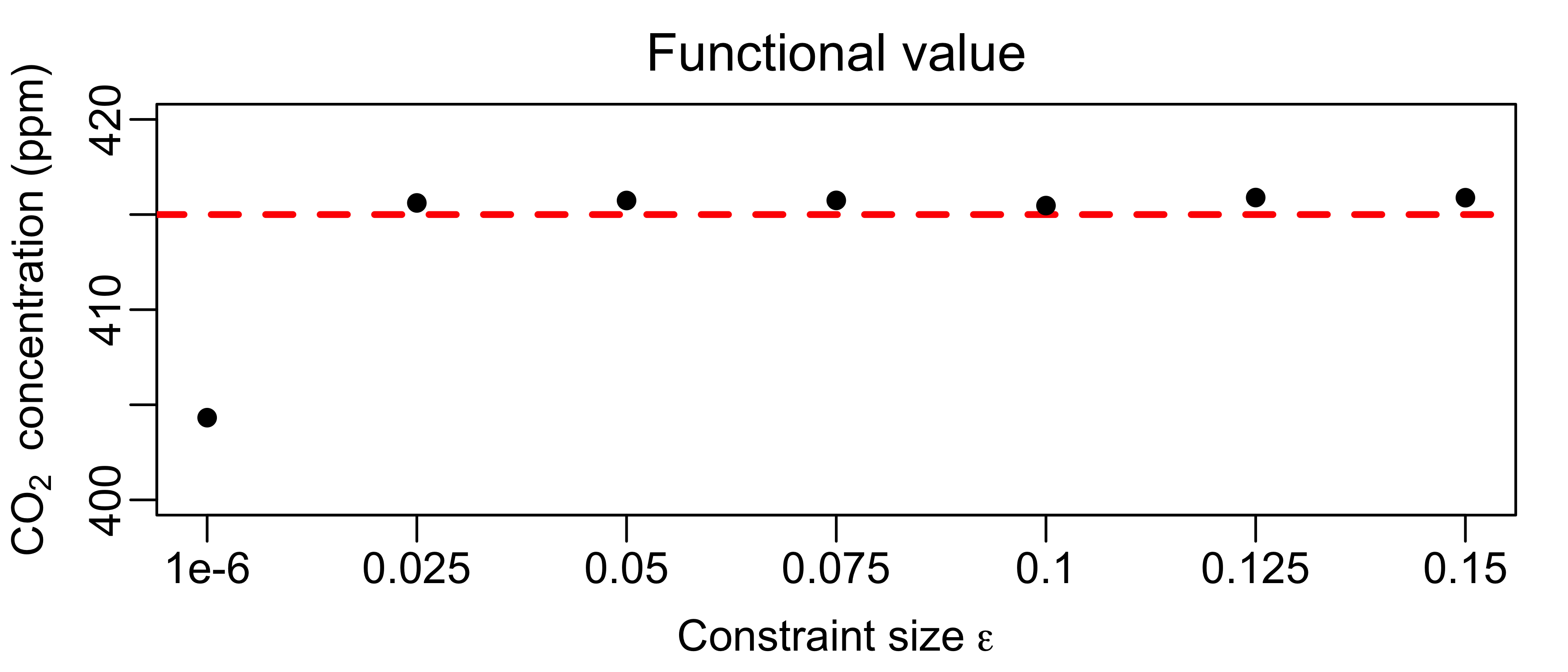}

		\includegraphics[width=0.485\textwidth]{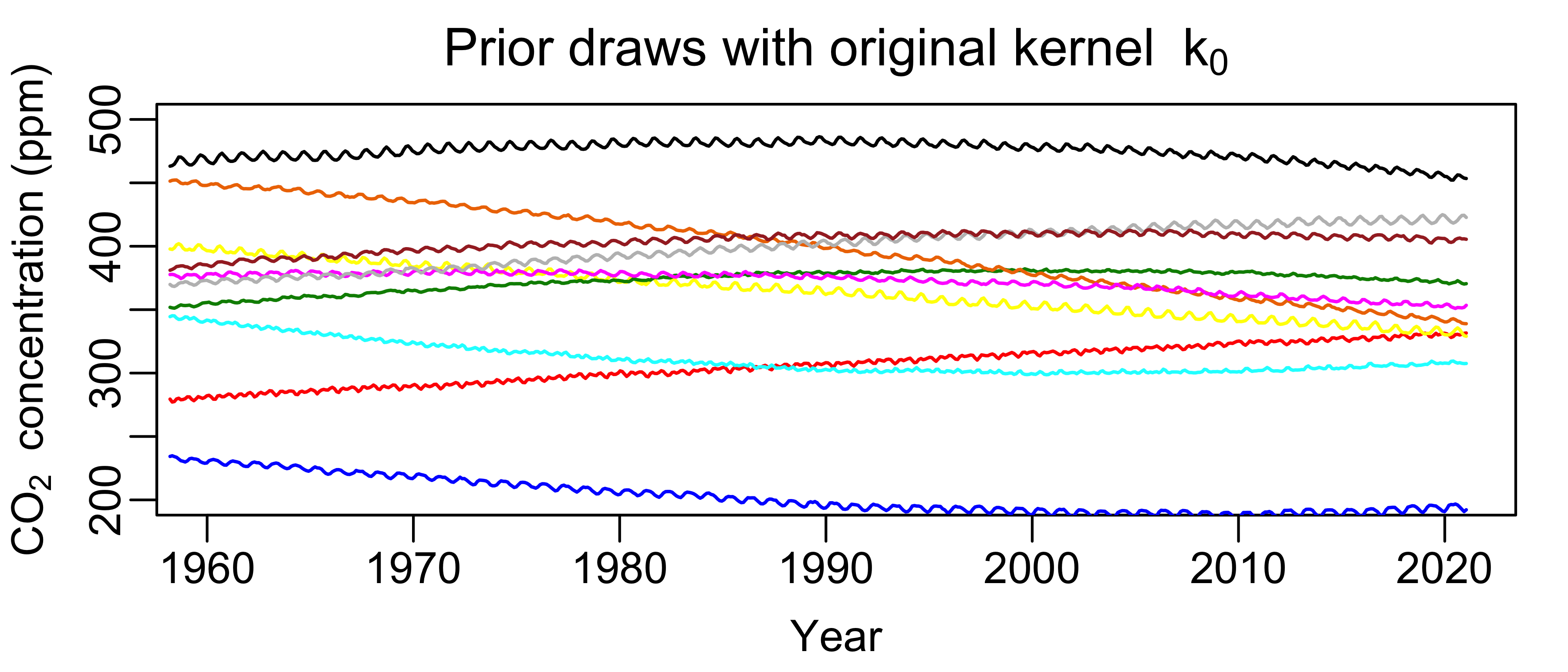}
		\includegraphics[width=0.485\textwidth]{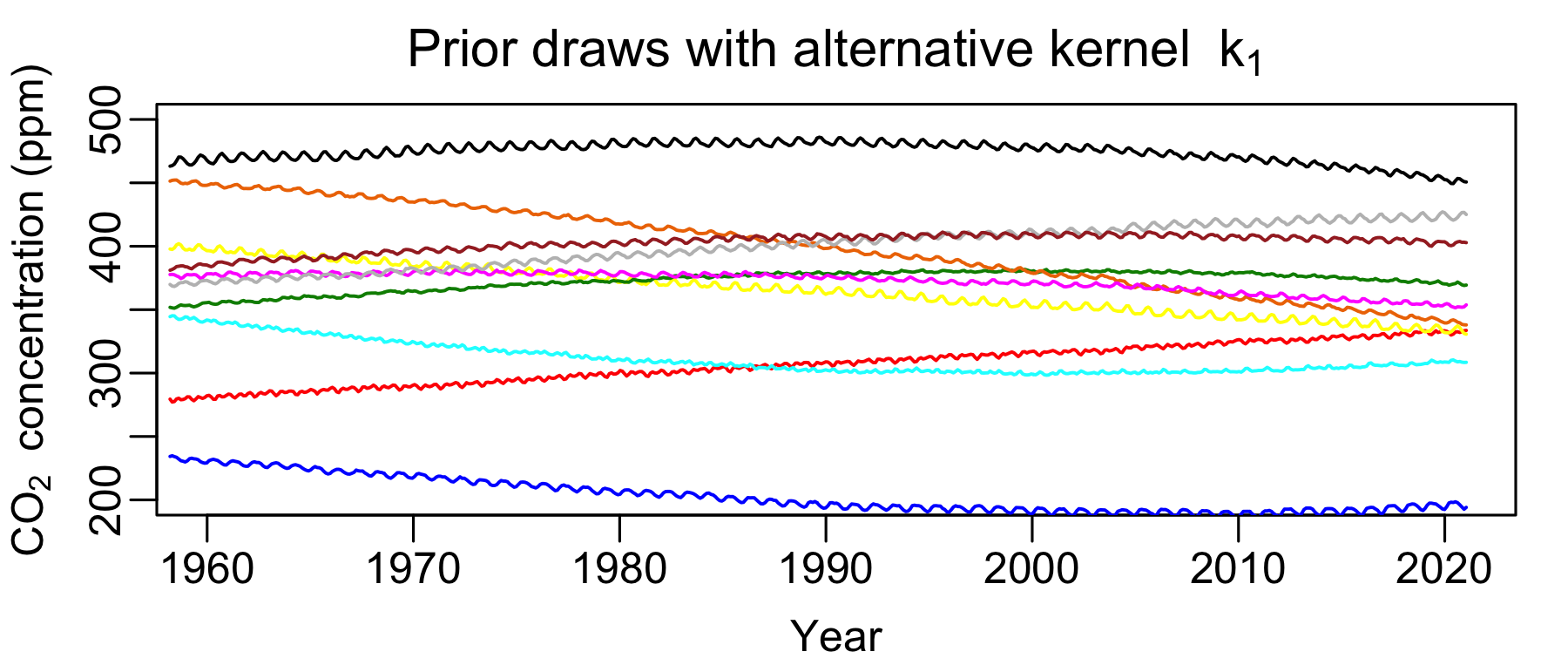}
%	\end{tabular}
\caption{\small{Sensitivity of the Mauna Loa analysis in \cref{sec:co2}. (\emph{Top-left}): Predictions made with the original kernel $k_{0}$ (black) and a qualitatively interchangeable kernel $k_{1}$ (red). (\emph{Top-right}): $\Fstar$, the mean \co2 level in June 2020, as a function of $\eps$. (\emph{Bottom}): Noise-matched draws from a $\GP(0,k_{0})$ (\emph{left}) and $\GP(0,\kperturb)$ (\emph{right}) prior. See \cref{app:co2} for a closer inspection of each prior draw.}}
	%\caption{\wts{Get these panels to be labeled w/ a, b, c, d like in other figs} Sensitivity analysis of the Mauna Loa  \cref{sec:co2}. (\emph{Top}): Predictions under the original (black) and the perturbed (red) kernels from 2003 to present. (\emph{Bottom-left and Bottom-center}): Noise-matched prior draws from $k_0$ and $k_1$, respectively. See \wts{appendix} for a closer inspection of each prior draw. (\emph{Bottom-right}): $\Fstar$ -- the mean \co2 level in June 2020 -- as a function of constraint set size and prior draws from the original and perturbed kernels.}
	\label{fig:MaunaLoa}
\end{figure*}

In a now-classic analysis of carbon dioxide (\co2) levels at Mauna Loa, \cite{RasmussenWilliams2006} predicted future \co2 levels based on data up to 2003. 
With data up to 2021, we can now see that the \cite{RasmussenWilliams2006} analysis substantially underestimates present-day \co2 levels; compare the gray region (99.7\% quantile of the original predictions) to the green (true levels) in \cref{fig:MaunaLoa}.
In this section, we show that this prediction of modern \co2 levels is non-robust to kernel choice. In \cref{app:co2_ac}, we repeat the analysis with a kernel whose structure is learned using the automatic statistician~\citep{Duvenaud2013_compositional} and discover similar lack of robustness. 

%As it turns out, their model significantly underestimates current \co2 levels; compare the gray region (99.7\% quantile of the original predictions) to the the green (true levels) in \cref{fig:MaunaLoa}.

\textbf{Data, model, and decision.}
%We start by recreating the analysis of \cite{RasmussenWilliams2006}.
At the present day, monthly data for \co2 emissions is available from the year 1958 through 2021.
But \cite{RasmussenWilliams2006} use training data up to 2003.
\cite{RasmussenWilliams2006} use a kernel that is a sum of four basic kernels, where each term plays a specific role; e.g.\ a periodic term models the periodic seasonal trend in \co2 levels.
See \cref{app:co2} for a full description of the kernel. We take this kernel with hyperparameters fit via MMLE as our $k_0$.
%\cite{RasmussenWilliams2006} ran their original analysis in 2006 and only used data up to 2003. 
Actual \co2 levels breached 415 ppm \emph{for the first time in human history}~\citep{smith19} in 2019.
Under $k_0$, this level lies more than three standard deviations away from the predicted means in all of 2019.
To see whether a qualitatively interchangeable $k_1$ would better predict modern \co2 levels, we let $\Fstar$ be the smooth-max of all posterior means in 2019. 
We will say the posterior has substantively changed if $\Fstar \geq \bigChange = 415$ ppm.

\textbf{Prior beliefs.} While $k_0$ is a stationary kernel we might also have non-stationary prior information such as known historical or expected future developments in climate policy or technology. 
So we choose $\Keps$ according to the non-stationary input-warping specification in \cref{sec:nearbyKernels}.
For our regularizer grid, we use 600 evenly spaced points $\tilde{\bx}_1, \dots, \tilde{\bx}_{600}$ between 1958 and 2021 to control $h$ throughout our time period of interest.
Input warping the entire kernel as $k = k_0(g(\bx), g(\bx'))$ would violate an important piece of prior information that we have about \co2 levels: \co2 has a regular seasonality, with minimal levels in the winter and maximal levels in the summer.
The original $k_0$ accounts for this feature of the data with a periodic term; \cref{fig:MaunaLoa} (top) shows that this periodicity lines up very well with the training data.
To produce an alternative kernel that accounts for this piece of prior knowledge, we leave the periodic portion of the kernel unwarped. 
To parameterize $g$, we use a fully connected network with two hidden layers, $50$ units, and ReLU nonlinearities.
To ensure the optimal $k_1$ is finite, we take the loss in \cref{constraintSet:nonStationary} to be $\ell(k; \Fstar, \bigChange) = (\Fstar(k) - \bigChange)^2$ to guarantee our objective is bounded below.

%Our original $k_0$ is the original kernel from \cite{RasmussenWilliams2006}, as described in \cref{app:co2}, with hyperparameters fit via MMLE.
%We will consider prediction at the point $\xtest$ corresponding to June of 2020 and let $\Fstar$ be the posterior mean at $\xtest$.
%We will say the posterior has substantively changed if $\Fstar = \bigChange$, where $\bigChange$ is the actual observed \co2 level in June of 2020.
%$k_0$ is a stationary kernel;
%however, it seems plausible to consider alternative kernels that are non-stationary.
%One might imagine including non-stationary prior information, such as known historical or expected future developments in climate policy or technology.

% sg commenting out subfigure, which is liberal with its white space use.
\begin{comment}
\begin{figure}

\begin{subfigure}[b]{0.95\textwidth}
\centering
\includegraphics[width = \textwidth]{figures/mauna_loa_posterior}
\caption{}
\label{fig:MaunaLoa_curves}
\end{subfigure}

\begin{subfigure}[b]{0.32\textwidth}
\centering
\includegraphics[width = \textwidth]{figures/mauna_loa_prior_original}
\caption{}
\label{fig:MaunaLoa_prior_original}
\end{subfigure}
\begin{subfigure}[b]{0.32\textwidth}
\centering
\includegraphics[width = \textwidth]{figures/mauna_loa_prior_perturbed}
\caption{}
\label{fig:MaunaLoa_prior_perturb}
\end{subfigure}
\begin{subfigure}[b]{0.32\textwidth}
\centering
\includegraphics[width = \textwidth]{figures/maunaloa-constraintset}
\caption{}
\label{fig:MaunaLoa_prior_epsilon}
\end{subfigure}
\caption{Sensitivity analysis of the Mauna Loa  \cref{sec:co2}. \skd{edit this and re-make the constraint set plot}(\emph{Top}): Predictions under the original and the perturbed kernels from 2003 to present. (\emph{Bottom-left}): $\Fstar$ -- the mean \co2 level in June 2020 -- as a function of constraint set size and prior draws from the original and perturbed kernels. \sg{TODO: Elaborate caption}.}
\label{fig:MaunaLoa}
\end{figure}
\end{comment}

\textbf{Robustness.} 
We now use our workflow to ask whether qualitatively interchangeable kernels might have better predicted the record-breaking \co2 levels in 2019; see \cref{fig:MaunaLoa} for our results.
%In \cref{fig:MaunaLoa}, we show the results of our framework. 
We lower the regularization strength (i.e.\ increase $\eps$ in \cref{constraintSet:nonStationary}) until $\Fstar \ge \bigChange$.
In the bottom of \cref{fig:MaunaLoa}, we plot noise-matched prior draws from $k_0$ alongside draws from the resulting $\kperturb$.
Differences between draws from $k_0$ and $\kperturb$ are almost visually indistinguishable on this scale.
A closer inspection in \cref{app:co2} confirms that the two priors capture the same yearly periodic trend. 
These same zoomed-in plots show that the priors are not completely indistinguishable; however, in our opinion, the draws display the same prior beliefs. 
The 2-Wasserstein comparison in \cref{fig:mauna_loa_histograms} of \cref{app:co2} further confirms that the perturbed kernel sits well within the histogram of alternate kernels stemming from hyperparameter uncertainty.
We conclude that future predictions of \co2 levels using the original $k_0$ are non-robust to the choice of kernel in the sense of \cref{def:nonRobust}.

\section{NON-STATIONARY PERTURBATIONS IN CLASSIFYING MNIST DIGITS}
\label{sec:MNIST}

%\begin{figure}
%\includegraphics[width=0.22\textwidth]{figures/mnist_epsilon_new}
%\includegraphics[width=0.22\textwidth]{figures/mnist_wasserstein}
%\caption{\small{Sensitivity of MNIST analysis in \cref{sec:MNIST}. \emph{(Left)}: $\Fstar$ as a function of regularizer strength. \emph{(Right)}: Histogram of the 2-Wasserstein distances between the $1000$ (one for each test image) input-warped kernel Gram matrices (in red) plotted with those arising from kernel hyperparameter uncertainty (in black).}}
%\label{fig:MNIST1}
%\end{figure}

\begin{figure}
\centering
\includegraphics[width=1.0\textwidth]{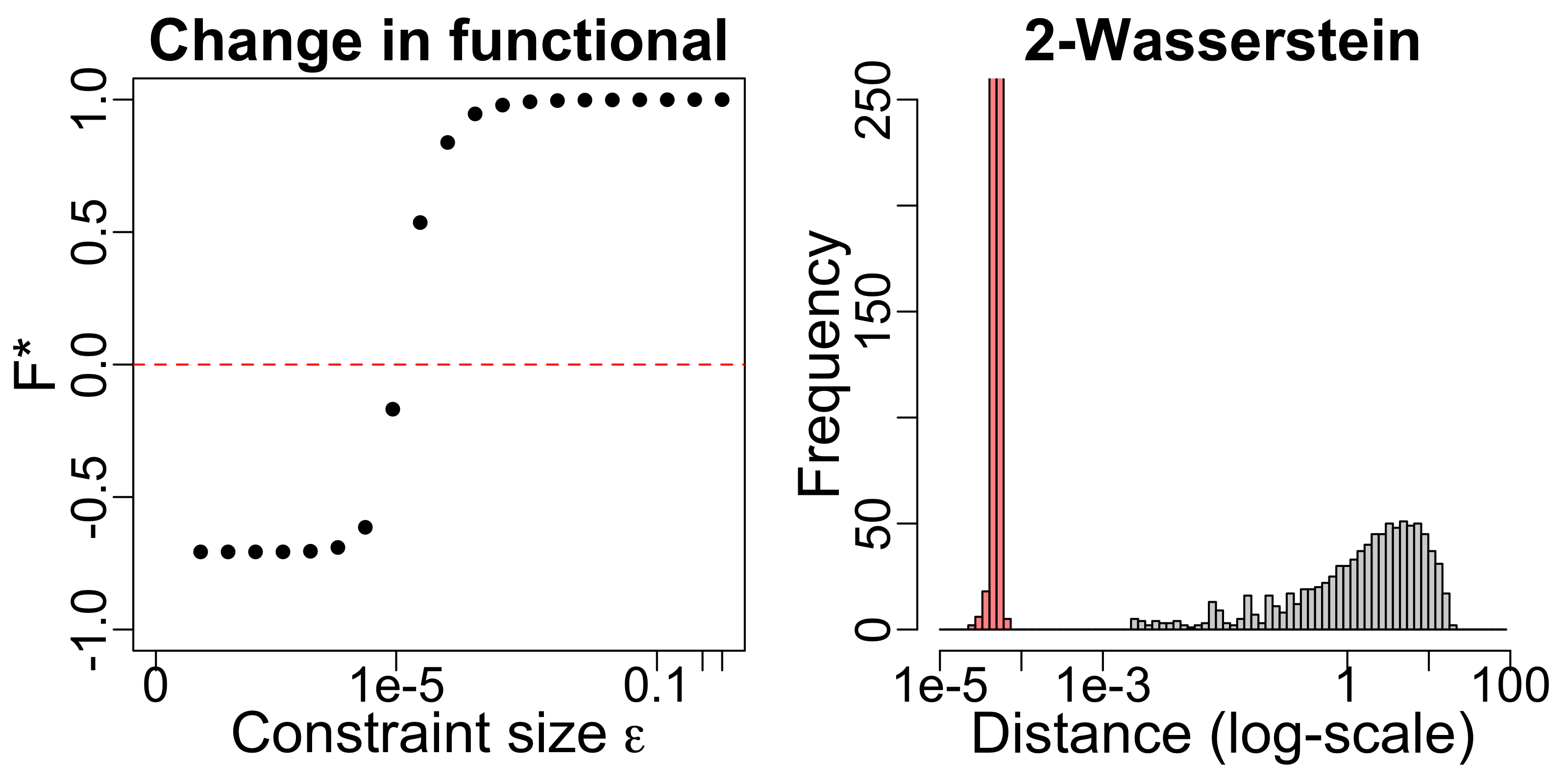}
\caption{\small{Sensitivity of MNIST analysis in \cref{sec:MNIST}. \emph{(Left)}: $\Fstar$ as a function of regularizer strength. \emph{(Right)}: Histogram of the 2-Wasserstein distances between the $1000$ (one for each test image) input-warped kernel Gram matrices (in red) plotted with those arising from kernel hyperparameter uncertainty around $k_0$ (in black).}}
\label{fig:MNIST1}
\end{figure}

So far we have restricted our attention to low-dimensional covariates. To evaluate our approach in a high-dimensional setting, we reproduce the MNIST image classification experiment of \cite{Lee2018_dnn_gp}. 
It is rare to have specific beliefs about high-dimensional functions, so in this case we do not consider $k_0$ as arising from prior beliefs. Rather we imagine $k_0$ is used purely for convenience and predictive quality -- but that a malicious actor is interested in changing the kernel to achieve different test predictions without detection.

\textbf{Data, model, and decision.} Similar to \cite{Lee2018_dnn_gp}, we use 1000 randomly sampled MNIST images for a training set, and a separate 1000 images for a test set. Given a test image $\xtest$, \cite{Lee2018_dnn_gp} predict the class label $c \in \{1,\dots,C\}$ by using a $C$-output GP with compositional structure, considered as the infinite-width limit of a sequence of Bayesian neural networks \citep{Lee2018_dnn_gp, AMatthews2018}.
Let $f_c(\xtest)$ be the $c$th output. The authors classify any image $\xtest$ by picking the class $c$ that has the posterior mean of $f_c(\xtest)$, i.e.\ $\mu_c(\xtest)$, closest to 0.9; see \cref{app:MNIST} for details. 
We use the kernel and hyperparameters from \citet{Lee2018_dnn_gp} for $k_0$.
We imagine that the malicious actor wants to change the label of a single test image $\xtest$ from its current label $c_0$ to a different label $c_1$. For concreteness, we set $c_{1} := |c_{0}-1|$. We consider 1000 separate iterations of this exercise, once for each of the 1000 test images. 
For a particular $\xtest$, we set our posterior quantity of interest to be $\Fstar = |\mu_{c_{0}}(\xtest) - 0.9| - |\mu_{c_{1}}(\xtest) - 0.9|$. Since $\Fstar \geq 0$ implies that we have changed the prediction for $\xtest$, we set our decision threshold $L = 0$.

\textbf{Malicious actor.}
Instead of considering a range of priors that match prior beliefs, we here consider a range of priors that will allow a malicious actor to avoid detection.
Since we have no prior belief of stationarity, we use the non-stationary construction from \cref{sec:nearbyKernels}. 
We find that optimizing $\Fstar$ directly leads to kernels where for $c \neq c_1$, $\mu_{c}(\xtest)$ takes on values at least an order of magnitude higher than for the original kernel. This change could be easily detected by an automated system. For the purposes of the malicious actor, we therefore consider these kernels to not be qualitatively interchangeable.
We instead optimize a surrogate loss that maximizes the log probability of $c_{1}$ being correct and all other classes being incorrect; see \cref{app:MNIST}.
We find that optimizing this surrogate loss leads to more benign-looking $\mu_c(\xtest)$ and achieves $\Fstar \geq L$.

\textbf{Robustness.}
\cref{fig:MNIST1} shows the results of our workflow applied to this problem.
We find a sufficiently large setting of $\eps$ that allows us, across all 1000 test-image problems, to change every decision. In particular, for $\eps = 10^{-4},$ we are able to find perturbed kernels that change the predicted class label in every case.
It is not clear how to visualize our priors in this application. So, of the approaches in \cref{sec:QI}, we use only the hyperparameter uncertainty  visualization to assess qualitative interchangeability.
\cite{Lee2018_dnn_gp} optimize the hyperparameters of $k_0$ over a grid.
Instead of bootstrapping this procedure, we note that the size of the grid defines a natural variability in the hyperparameters $\hat\theta$.
To be conservative, we sample $\theta^{(r)}$ from an area around the hyperparameters selected by \cite{Lee2018_dnn_gp} that is over 10 times smaller than the full grid. 
(Using the full original grid would find more extreme non-robustness.) %14
The Gram matrices corresponding to our perturbed kernels are much closer to $k_{0}(X,X)$ than are the Gram matrices corresponding to each $\theta^{(r)}$. 
We conclude that classification of handwritten digits using $k_0$ is non-robust to the choice of kernel in the sense of \cref{def:nonRobust}.

%
%
%\begin{comment}
%\cref{fig:MNIST} presents our results.
%Using $\eps = 10^{-4},$ we are able to find perturbed kernels that change the predicted class label for each of the 1000 test images %considered (\cref{fig:MNIST}).
%It is not clear how to visualize our priors in this context, so we only assess qualitative interchangeability using our hyperparameter uncertainty approach from \cref{sec:QI}.
%\cite{Lee2018_dnn_gp} optimize the hyperparameters of $k_0$ over a small grid, by maximizing validation set accuracy. To be conservative, we choose $\theta^{(r)}$ to lay on a grid that is $14$ times smaller than the part of the grid over which they found high validation performance.
%The Gram matrices corresponding to our perturbed kernels are much closer to $k_{0}(X,X)$ than are the Gram matrices corresponding to each $\theta^{(r)}$. 
%We conclude that classifications of handwritten digits with the compositional kernel proposed by~\cite{Lee2018_dnn_gp} are non-robust to the choice of kernel in the sense of \cref{def:nonRobust}.
%\end{comment}

\section{DISCUSSION}

\label{sec:discussion}
In this paper, we proposed and implemented a workflow for measuring the sensitivity of GP inferences to the choice of the kernel function.
We used our workflow to discover substantial non-robustness in a variety of practical examples, but also showed that many analyses are not flagged as non-robust by our method.
There are many exciting directions for expanding on the present work -- both within our existing workflow and beyond.
We discuss these directions below.

\textbf{Improving our workflow.} 
Our workflow is made up of many modular parts, and in some parts choices were made for mathematical convenience (e.g.\ the particular constraint in our stationary objective or the particular regularizer in our non-stationary objective in \cref{constraintSet:nonStationary}).
Perhaps different choices could be made to allow easier detection of non-robustness -- or to allow for certification of robustness (whereas our workflow can only fail to find non-robustness)

\textbf{What should we do about non-robustness?} 
Our framework flags non-robustness but does not show how to make an analysis more robust. The instances of non-robustness we have found suggest it might be worthwhile to develop methods to robustify GP inferences to the choice of kernel. 
One challenge would be how to best balance robustness against the
ability to adapt to the prior assumptions at hand: if a method is completely robust to any
change in prior assumptions, there is no point in specifying a prior at all!

\textbf{Studying model selection and robustness.}
Our example in \cref{sec:co2} shows that the use of sophisticated kernel selection tools does not necessarily mean robustness issues are not present.
However, it could be that non-robustness is typically lessened or removed by the use of such tools.
Or maybe certain classes of kernel selection tools ameliorate non-robustness issues more than others.
It remains to formalize and study these questions.

\subsubsection*{Acknowledgements}
The authors thank Ryan Giordano for useful conversations about this work. This work was supported by the MIT-IBM Watson AI Lab, an NSF CAREER Award, an ARPA-E project with program director David Tew, and an ONR MURI.

%\small % The style file says we can reduce to small font size!
%\bibliographystyle{plain}
\bibliographystyle{apalike}
\bibliography{ms}

\begin{thebibliography}{}

\bibitem[Andrews et~al., 1972]{Andrews1972_robustness}
Andrews, D.~F., Bickel, P.~J., Hampel, F.~R., Huber, P.~J., Rogers, W.~H., and
  Tukey, J.~W. (1972).
\newblock Robust estimates of location: Survey and advances.
\newblock Technical report, Princeton University.

\bibitem[Benton et~al., 2019]{Benton2019_function}
Benton, G.~W., Maddox, W.~J., Salkey, J.~P., Albinati, J., and Wilson, A.~G.
  (2019).
\newblock Function-space distribution over kernels.
\newblock In {\em Advances in Neural Information Processing Systems (NeurIPS)}.

\bibitem[Berger, 2000]{Berger2000}
Berger, J.~O. (2000).
\newblock {\em Bayesian Robustness}, pages 1--32.
\newblock Springer New York, New York, NY.

\bibitem[Berger et~al., 1994]{Berger1994_overview}
Berger, J.~O., Moreno, E., Pericchi, L.~R., Bayarri, M.~J., Bernardo, J.~M.,
  Cano, J.~A., De~la Horra, J., Mart{\'\i}n, J., R{\'\i}os-Ins{\'u}a, D.,
  Betr{\`o}, B., Dasgupta, A., Gustafson, P., Wasserman, L., Kadane, J.~B.,
  Srinivasan, C., Lavine, M., O'Hagan, A., Polasek, W., Robert, C.~P., Goutis,
  C., Ruggeri, F., Salinetti, G., and Sivaganesan, S. (1994).
\newblock An overview of robust {Bayesian} analysis.
\newblock {\em Test}, 3(1):5 -- 124.

\bibitem[Bevilacqua et~al., 2019]{Bevilacqua2019_kernelEquivalence}
Bevilacqua, M., Faouzi, T., Furrer, R., and Porcu, E. (2019).
\newblock Estimation and prediction using generalized {W}endland covariance
  functions under fixed domain asymptotics.
\newblock {\em Annals of Statistics}, 47(2).

\bibitem[Bogunovic et~al., 2018]{Bogunovic2018_optimization}
Bogunovic, I., Scarlett, J., Jegelka, S., and Cevher, V. (2018).
\newblock Adversarially robust optimization with {Gaussian} processes.
\newblock In {\em Advances in Neural Information Processing Systems (NeurIPS)}.

\bibitem[Bradbury et~al., 2018]{jax2018_github}
Bradbury, J., Frostig, R., Hawkins, P., Johnson, M.~J., Leary, C., Maclaurin,
  D., Necula, G., Paszke, A., Vander{P}las, J., Wanderman-{M}ilne, S., and
  Zhang, Q. (2018).
\newblock {JAX}: composable transformations of {P}ython+{N}um{P}y programs.

\bibitem[Cardelli et~al., 2019]{Cardelli2019_aaai}
Cardelli, L., Kwiatkowska, M., Laurenti, L., and Patane, A. (2019).
\newblock Robustness guarantees for {Bayesian} inference with {Gaussian}
  processes.
\newblock In {\em AAAI Conference on Artificial Intelligence}.

\bibitem[Cheng et~al., 2020]{Cheng2020_medgp}
Cheng, L.-F., Dumitrascu, B., Darnell, G., Chivers, C., Draugelis, M., Li, K.,
  and Engelhardt, B.~E. (2020).
\newblock Sparse multi-output {Gaussian} processes for online medical time
  series prediction.
\newblock {\em BMC Medical Informatics and Decision Making}, 20(1):152 -- 174.

\bibitem[Colopy et~al., 2016]{Colopy2016_identifying}
Colopy, G.~W., Pimentel, M. A.~F., Roberts, S.~J., and Clifton, D.~A. (2016).
\newblock {Bayesian Gaussian} processes for identifying the deteriorating
  patient.
\newblock In {\em 2016 38th Annual International Conference of the {IEEE}
  Engineering in Medicine and Biology Society (EMBC)}.

\bibitem[de~G.~Matthews et~al., 2018]{AMatthews2018}
de~G.~Matthews, A.~G., Hron, J., Rowland, M., Turner, R.~E., and Ghahramani, Z.
  (2018).
\newblock Gaussian process behaviour in wide deep neural networks.
\newblock In {\em International Conference on Learning Representations (ICLR)}.

\bibitem[Duvenaud, 2014]{Duvenaud_thesis}
Duvenaud, D. (2014).
\newblock {\em Automatic model construction with {Gaussian} processes}.
\newblock PhD thesis, University of Cambridge.

\bibitem[Duvenaud et~al., 2013]{Duvenaud2013_compositional}
Duvenaud, D., Lloyd, J.~R., Grosse, R., Tenenbaum, J.~B., and Ghahramani, Z.
  (2013).
\newblock Structure discovery in nonparametric regression with compositional
  kernel search.
\newblock In {\em Proceedings of the 30th International Conference on Machine
  Learning (ICML)}.

\bibitem[Ferrari and Dunson, 2020]{FerrariDunson2020_identifying}
Ferrari, F. and Dunson, D.~B. (2020).
\newblock Identifying main effects and interactions among exposures using
  {Gaussian} processes.
\newblock {\em Annals of Applied Statistics}, 14(4):1743 -- 1758.

\bibitem[Fidler et~al., 2017]{Fidler2017_heartRateThreshold}
Fidler, R.~L., Pelter, M.~M., Drew, B.~J., Palacios, J.~A., Bai, Y., Stannard,
  D., Aldrich, J.~M., and Hu, X. (2017).
\newblock Understanding heart rate alarm adjustment in the intensive care units
  through an analytical approach.
\newblock {\em PLoS One}, 12(11):1 -- 10.

\bibitem[Futoma et~al., 2017a]{Futoma2017_learning}
Futoma, J., Hariharan, S., and Heller, K. (2017a).
\newblock Learning to detect sepsis with a multitask {G}aussian process {RNN}
  classifier.
\newblock In {\em Proceedings of the 34th International Conference on Machine
  Learning (ICML)}, volume~70.

\bibitem[Futoma et~al., 2017b]{Futoma2017_improved}
Futoma, J., Hariharan, S., Heller, K., Sendak, M., Brajer, N., Clement, M.,
  Bedoya, A., and O'Brien, C. (2017b).
\newblock An improved multi-output {Gaussian} process {RNN} with real-time
  validation for early sepsis detection.
\newblock In {\em Proceedings of the 2nd Machine Learning for Healthcare
  Conference (MLHC)}.

\bibitem[Gabry et~al., 2019]{Gabry2019_bayesianViz}
Gabry, J., Simpson, D., Vehtari, A., Betancourt, M., and Gelman, A. (2019).
\newblock Visualization in {B}ayesian workflow.
\newblock {\em Journal of the Royal Statistical Society: Series A (Statistics
  in Society)}, 182(2):389 -- 402.

\bibitem[Giordano et~al., 2021]{Giordano2021_stickbreaking}
Giordano, R., Liu, R., Jordan, M.~I., and Broderick, T. (2021).
\newblock Evaluating sensitivity to the stick breaking prior in {Bayesian}
  nonparametrics.
\newblock \textit{arXiv preprint arXiv:2107.03584v2}.

\bibitem[Goldberger et~al., 2000]{compCardiology2}
Goldberger, A.~L., Amaral, L. A.~N., Glass, L., Hausdorff, J.~M., Ivanov,
  P.~C., Mark, R.~G., Mietus, J.~E., Moody, G.~B., Peng, C.-K., and Stanley, H.
  (2000).
\newblock Physio{B}ank, {P}hysio{T}oolkit, and {P}hysio{N}et: Components of a
  new research resource for complex physiologic signals.
\newblock {\em Circulation}, 101(23):e215 -- e220.

\bibitem[Goodfellow et~al., 2015]{Goodfellow2015_adversarial}
Goodfellow, I.~J., Shlens, J., and Szegedy, C. (2015).
\newblock Explaining and harnessing adversarial examples.
\newblock In {\em International Conference on Learning Representations (ICLR)}.

\bibitem[Gustafson, 1996]{Gustafson1996_local}
Gustafson, P. (1996).
\newblock Local sensitivity of posterior expectations.
\newblock {\em Annals of Statistics}, 24(1):174 -- 195.

\bibitem[Hern\'{a}ndez-Lobato et~al., 2011]{Hernandez-Lobato2011_multiclass}
Hern\'{a}ndez-Lobato, D., Hern\'{a}ndez-Lobato, J.~M., and Dupont, P. (2011).
\newblock Robust multi-class {Gaussian} process classification.
\newblock In {\em Advances in Neural Information Processing Systems (NeurIPS)}.

\bibitem[Huber and Ronchetti, 2009]{Huber2009_robustStats}
Huber, P.~J. and Ronchetti, E. (2009).
\newblock {\em Robust Statistics}.
\newblock John Wiley and Sons, Inc.

\bibitem[Huggins et~al., 2020]{huggins_2020_validated}
Huggins, J.~H., Kasprzak, M., Campbell, T., and Broderick, T. (2020).
\newblock Validated variational inference via practical posterior error bounds.
\newblock In {\em Proceedings of the 19th International Conference on
  Artificial Intelligence and Statistics (AISTATS)}.

\bibitem[Insua and Ruggeri, 2000]{insua2000_robustBayes}
Insua, D.~R. and Ruggeri, F. (2000).
\newblock {\em Robust Bayesian Analysis}.
\newblock Springer.

\bibitem[Jyl{{\"a}}nki et~al., 2011]{Jylanki2011_student-t}
Jyl{{\"a}}nki, P., Vanhatalo, J., and Vehtari, A. (2011).
\newblock Robust {Gaussian} process regression with a {Student-t} likelihood.
\newblock {\em Journal of Machine Learning Research}, 12(99):3227 -- 3257.

\bibitem[Keeling et~al., 2005]{keeling2005atmospheric}
Keeling, C.~D., Piper, S.~C., Bacastow, R.~B., Wahlen, M., Whorf, T.~P.,
  Heimann, M., and Meijer, H.~A. (2005).
\newblock Atmospheric {$\text{CO}_{2}$} and {$^{13}\text{CO}_{2}$} exchange
  with the terrestrial biosphere and oceans from 1978 to 2000: Observations and
  carbon cycle implications.
\newblock In {\em A history of atmospheric {$\text{CO}_{2}$} and its effects on
  plants, animals, and ecosystems}, pages 83--113. Springer.

\bibitem[Kim and Ghahramani, 2008]{KimGhahramani2008_outlier}
Kim, H.-C. and Ghahramani, Z. (2008).
\newblock Outlier robust {Gaussian} process classification.
\newblock In {\em Structural, Syntactic, and Statistical Pattern Recognition},
  pages 896 -- 905.

\bibitem[Kirchner and Bolin, 2021]{Kirchner2021_kernelEquivalence}
Kirchner, K. and Bolin, D. (2021).
\newblock Necessary and sufficient conditions for asymptotically optimal linear
  prediction of random fields on compact metric spaces.
\newblock \textit{arXiv preprint arXiv:2005.08904v4}.

\bibitem[Lee et~al., 2017]{Lee2017_rigorous}
Lee, D., Mukhopadhyay, S., Rushwork, A., and Sahu, S.~K. (2017).
\newblock A rigorous statistical framework for spatio-temporal pollution
  predition and estimation of its long-term impact on health.
\newblock {\em Biostatistics}, 2:370 -- 385.

\bibitem[Lee et~al., 2018]{Lee2018_dnn_gp}
Lee, J., Sohl-Dickstein, J., Pennington, J., Novak, R., Schoenholz, S., and
  Bahri, Y. (2018).
\newblock Deep neural networks as {Gaussian} processes.
\newblock In {\em International Conference on Learning Representations (ICLR)}.

\bibitem[Novak et~al., 2020]{neuraltangents2020}
Novak, R., Xiao, L., Hron, J., Lee, J., Alemi, A.~A., Sohl-Dickstein, J., and
  Schoenholz, S.~S. (2020).
\newblock Neural tangents: Fast and easy infinite neural networks in python.
\newblock In {\em International Conference on Learning Representations (ICLR)}.

\bibitem[Ranjan et~al., 2016]{Ranjan2016_em}
Ranjan, R., Huang, B., and Fatehi, A. (2016).
\newblock Robust {Gaussian} process modeling using {EM} algorithm.
\newblock {\em Journal of Process Control}, 42:125 -- 136.

\bibitem[Rasmussen and Williams, 2006]{RasmussenWilliams2006}
Rasmussen, C.~E. and Williams, C.~K. (2006).
\newblock {\em {Gaussian Processes for Machine Learning}}.
\newblock {The MIT Press}.

\bibitem[Ren et~al., 2021]{Ren2021_pollution}
Ren, B., Wu, X., Braun, D., Pillai, N., and Dominici, F. (2021).
\newblock Bayesian modeling for exposure response curve via {Gaussian}
  processes: causal effects of exposure to air pollution on health outcomes.
\newblock \textit{arXiv preprint arXiv:2105.0354v1}.

\bibitem[Reyna et~al., 2019]{compCardiology1}
Reyna, M., Josef, C., Jeter, R., Shashikumar, S., Moody, B., Westover, M.,
  Sharm, A., Nemati, S., and Clifford, G. (2019).
\newblock {\em Early Prediction of Sepsis from Clinical Data -- the PhysioNet
  Computing in Cardiology Challenge 2019 (version 1.0.0)}.

\bibitem[Solly, 2019]{smith19}
Solly, M. (2019).
\newblock Carbon dioxide levels reach highest point in human history.
\newblock {\em Smithsonian Magazine}.

\bibitem[Stein, 1993]{Stein1993_kernelEquivalence}
Stein, M.~L. (1993).
\newblock A simple condition for asymptotic optimality of linear predictions of
  random fields.
\newblock {\em Statistics and Probability Letters}, 17.

\bibitem[Teckentrup, 2020]{teckentrup2020_gpRates}
Teckentrup, A.~L. (2020).
\newblock Convergence of {G}aussian process regression with estimated
  hyper-parameters and applications in {B}ayesian inverse problems.
\newblock {\em SIAM/ASA Journal on Uncertainty Quantification}, 8:1310--1337.

\bibitem[van~der Vaart and van Zanten, 2011]{vaart2011_gpRates}
van~der Vaart, A. and van Zanten, H. (2011).
\newblock Information rates of nonparametric {G}aussian process methods.
\newblock {\em Journal of Machine Learning Research}, 12:2095--2119.

\bibitem[Wang and Jing, 2021]{WangJing2021_convergence}
Wang, W. and Jing, B.-Y. (2021).
\newblock Convergence of {Gaussian} process regression: Optimality, robustness,
  and relationship with kernel ridge regression.
\newblock \textit{arXiv preprint arXiv:2104.09778v1}.

\bibitem[Wilson and Adams, 2013]{WilsonAdams2013_sm}
Wilson, A.~G. and Adams, R.~P. (2013).
\newblock Gaussian process kernels for pattern discovery and extrapolation.
\newblock In {\em Proceedings of the 30th International Conference on Machine
  Learning (ICML)}.

\bibitem[Wilson et~al., 2016]{Wilson2016_dkl}
Wilson, A.~G., Hu, Z., Salakhutdinov, R., and Xing, E.~P. (2016).
\newblock Deep kernel learning.
\newblock In {\em Proceedings of the 19th International Conference on
  Artificial Intelligence and Statistics (AISTATS)}.

\bibitem[Wynne et~al., 2021]{Wynne2021_convergence}
Wynne, G., Briol, F.-X., and Girolami, M. (2021).
\newblock Convergence guarantees for {Gaussian} process means with misspecified
  likelihoods and smoothness.
\newblock {\em Journal of Machine Learning Research}, 22(123):1 -- 40.

\end{thebibliography}

\newpage

\newpage
\onecolumn
\appendix
\section{Details of spectral density constraints}
\label{app:spectral_density}
Here, we give the details of how we optimize over spectral densities to produce a stationary kernel as summarized in \cref{constraintSet:stationary}. Our goal is to optimize over the set of stationary kernels.
It is not immediately clear how to enforce this constraint; however, Bochner's theorem \cite[Thm.\ 4.1]{RasmussenWilliams2006} tells us that every stationary kernel $k(x, x') = k(\tau)$, where $\tau  = x - x'$ has a positive finite \emph{spectral measure} $\mu$ on $\R^D$ such that:
\begin{equation}
	k(\tau) = \int_{\R^D } e^{2\pi i \tau^T \omega} d\mu(\omega).
\end{equation}
A common assumption in the literature on kernel discovery \citep{WilsonAdams2013_sm,Benton2019_function,Wilson2016_dkl} is to assume that $\mu$ has a density $S$ with respect to the Lebesgue measure; that is, we can write:
\begin{equation}
	k(\tau) = \int_{\R^D} e^{2\pi i \tau^T \omega} S(\omega) d\omega. \label{spectralDensityGeneral}
\end{equation}
These works have shown that the class of stationary kernel with spectral densities is a rich, flexible class of kernels.
We thus focus on the class of stationary kernels with spectral densities as this allows us to transform the problem of optimizing over stationary kernels into the problem of optimizing over positive real valued functions.
In all of our examples optimizing over spectral densities, we have $D = 1$.
We thus assume $D = 1$ in the rest of our development here.
In this case, it must be that $S$ is symmetric around the origin to obtain a real-valued $k$.
So, we can simply \cref{spectralDensityGeneral} further as:
\begin{equation}
	k(\tau) = \int_0^\infty \cos(2\pi \tau\omega) S(\omega) d\omega. \label{finalSpectralDensity}
\end{equation}
Optimizing over positive functions $S$ on the positive real line seems at least somewhat more tractable than optimizing over stationary positive-definite functions $k(\tau)$. 
However, this is still an infinite dimensional optimization problem.
To recover a finite dimensional optimization problem, we follow \cite{Benton2019_function} and choose a grid $\omega_1, \dots, \omega_G$.
We can then optimize over the finite values $S(\omega_1), \dots, S(\omega_G)$ and use the trapezoidal rule to approximate the integral in \cref{finalSpectralDensity}.
\cite{Benton2019_function} find that $G = 100$ gives reasonable performance in their experiments; we find the same in ours, and fix $G = 100$ throughout.
\cite{Benton2019_function} recommend setting $\omega_g = 2\pi g / (8\tau_{max})$, where $\tau_{max}$ is the maximum spacing between datapoints.
We find this to sometimes give inaccurate results in the sense that using the trapezoidal rule / an exact formula to compute the density of $k_0$, $S(\omega_1), \dots, S(\omega_G)$ and then using the trapezoidal rule to recover the gram matrix $k_0(X,X)$ gives an inaccurate approximation to $k_0(X,X)$.
This is problematic in our case, as it would imply $k_0(X,X)$ is not in the constraint set for small $\eps$.
Instead, we recommend setting our $\omega_g$'s as a uniform grid from $\omega_1 = 0$ up to an $\omega_G$ such that $S_0(\omega_G)$ is equal to the floating point epsilon ($10^{-15}$ in our experiments); some manual experimentation will be required to implement this rule.

As we are only interested in kernels nearby $k_0$, we will have to put some kind of constraint on $k_1$'s spectral density,  $S_1(\omega_1), \dots, S_1(\omega_G)$.
We use a simple $\eps$-ball given by:
\begin{equation}
	 \max \big( 0, (1-\eps) S_0(\omega_g) \big)  \leq S_1(\omega_g) \leq (1+\eps) S_0(\omega_g), \quad g = 1, \dots, G,
	 \label{spectralConstraint}
\end{equation}
Because our posterior functional of interest $\Fstar$ is a differentiable function of the kernel matrix, we can compute gradients of $\Fstar$ with respect to our discretized spectral density.
Rather than manually work out the derivatives of the trapezoidal rule combined with $\Fstar$, we use the automatic differentiation package \texttt{jax}\footnote{Note that \texttt{jax} does not use 64 bit floating point numbers by default. We found that the increased precision given by 64 bit floating point arithmetic to be important in our experiments.} \citep{jax2018_github}.
Given a gradient of $\Fstar$, we take a step in the direction of the gradient and then project the current iterate onto our constraint set in \cref{spectralConstraint} by clipping the resulting spectral density.
\section{Additional details of synthetic-data experiment}
\label{app:synthetic-data}
We generated the $x$-component of the synthetic data by first drawing $25$ uniform random numbers in $[0,5].$
To investigate what happens when interpolating in a region of dense training data, we then add $3.00, 3.025, 3.075$, and $3.10$ as covariates (recall the interpolation point is $\xtest = 3.05$).
The extrapolation point $\xtest = 5.29$ lies $0.5$ to the right of the largest $x$ value drawn. 
The $y$-component is defined to be, $y_n = x_n + \epsilon_n$,
where $\epsilon_n \stackrel{iid}{\sim} \N(0, 1.5)$. 

%We chose the $\eps$ grid for extrapolation ($\xtest = 5.29$) to be $15$ evenly-spaced values between $0.2$ and $0.8.$ The grid for interpolation ($\xtest = 3.05$) is $15$ evenly-spaced values on the log grid between $10^{0.1}$ and $10^{1}.$ 

To discretize the spectral density, we follow \cref{app:spectral_density} in using $100$ frequencies evenly-spaced from $0$ to $3.5.$ To optimize over nearby spectral densities, we also perform constrained gradient descent with randomized initializations in the sense of \cref{app:spectral_density}. 
For extrapolation ($\xtest = 5.29$), using $25$ random seeds, we find non-robustness. 
For interpolation ($\xtest = 3.05$), even with $40$ random seeds, we do not find non-robustness.  

\cref{fig:synthetic_extrapolation_histograms} compares the distance between $k_{1}$ and $k_{0}$ to the distances between $k_{0}$ and $k^{(r)},$ where $k^{(r)}$ is a bootstrapped version of $k_{0}.$

Computation for the synthetic experiments is done using a computing cluster, which has xeon-p8 computing cores. We request $7$ nodes, each using $15$ cores to run parallel experiments across both $\epsilon$ and the random seed for initialization. Total wall-clock time comes to roughly $10$ minutes. 
\section{Additional details for the heart rate example}
\label{app:heartRate}

\begin{figure*}
	\centering
	\begin{tabular}{cc}
		\includegraphics[width=0.45\textwidth]{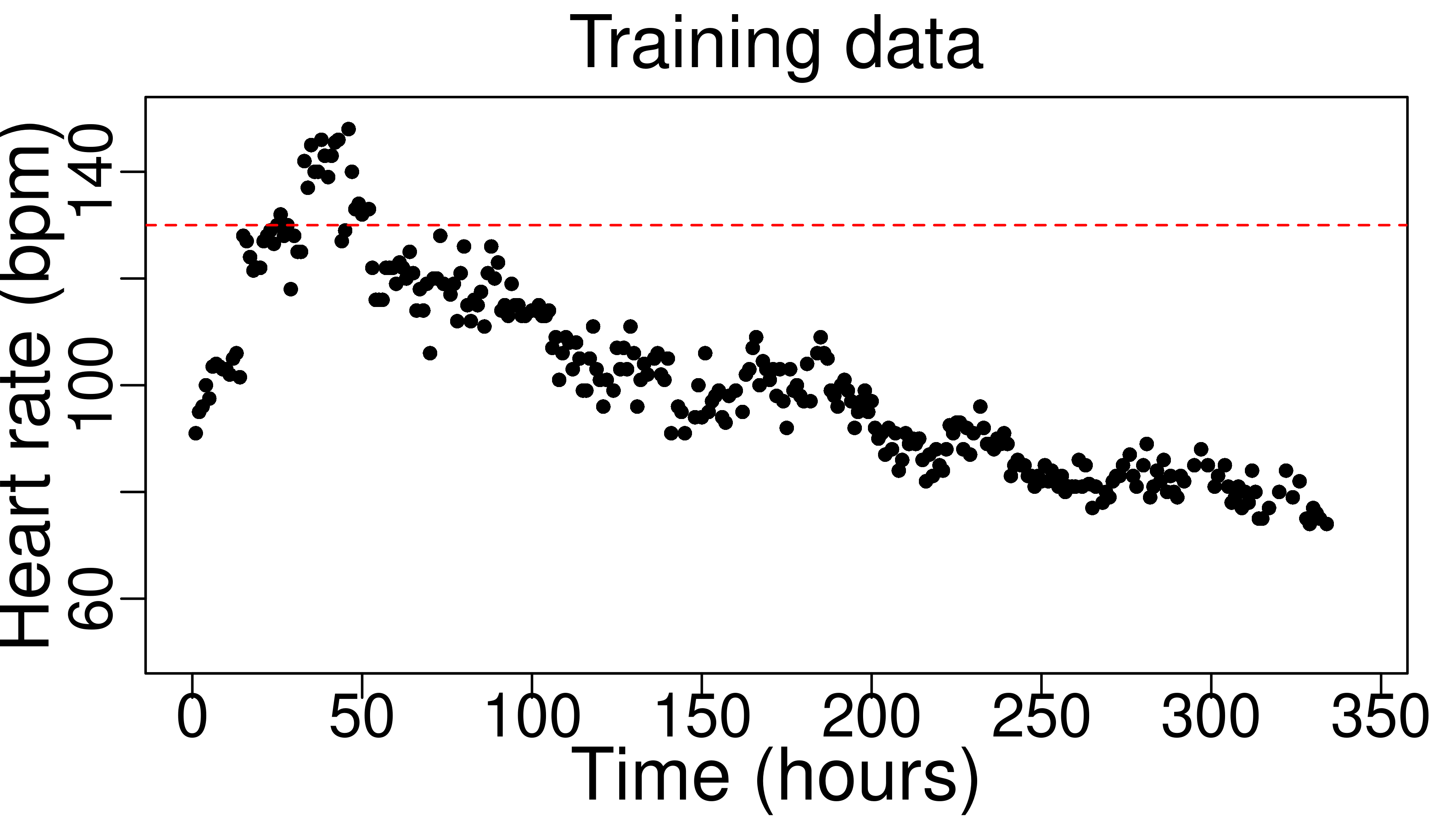} &
		\includegraphics[width=0.45\textwidth]{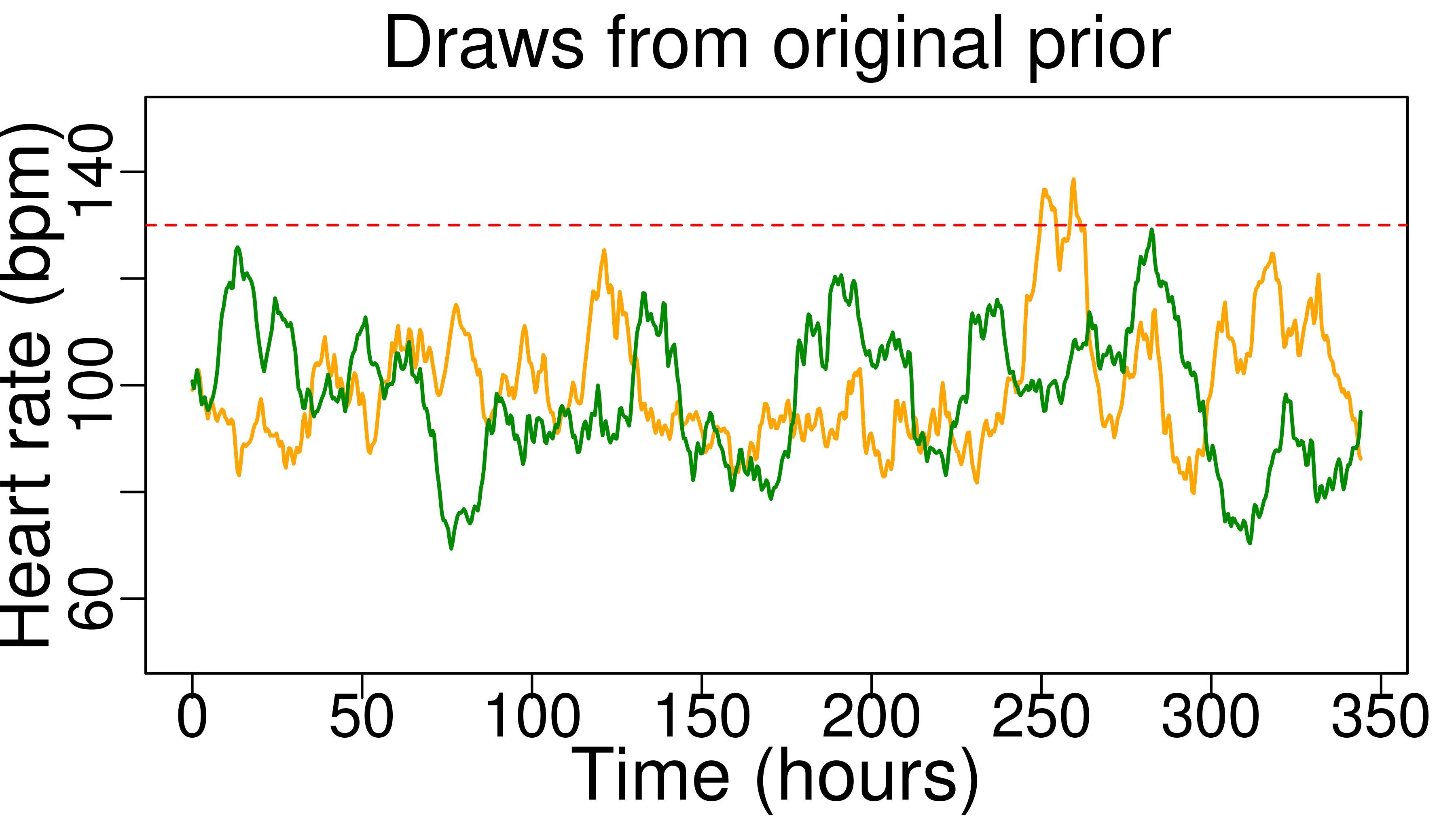} \\
		\includegraphics[width=0.45\textwidth]{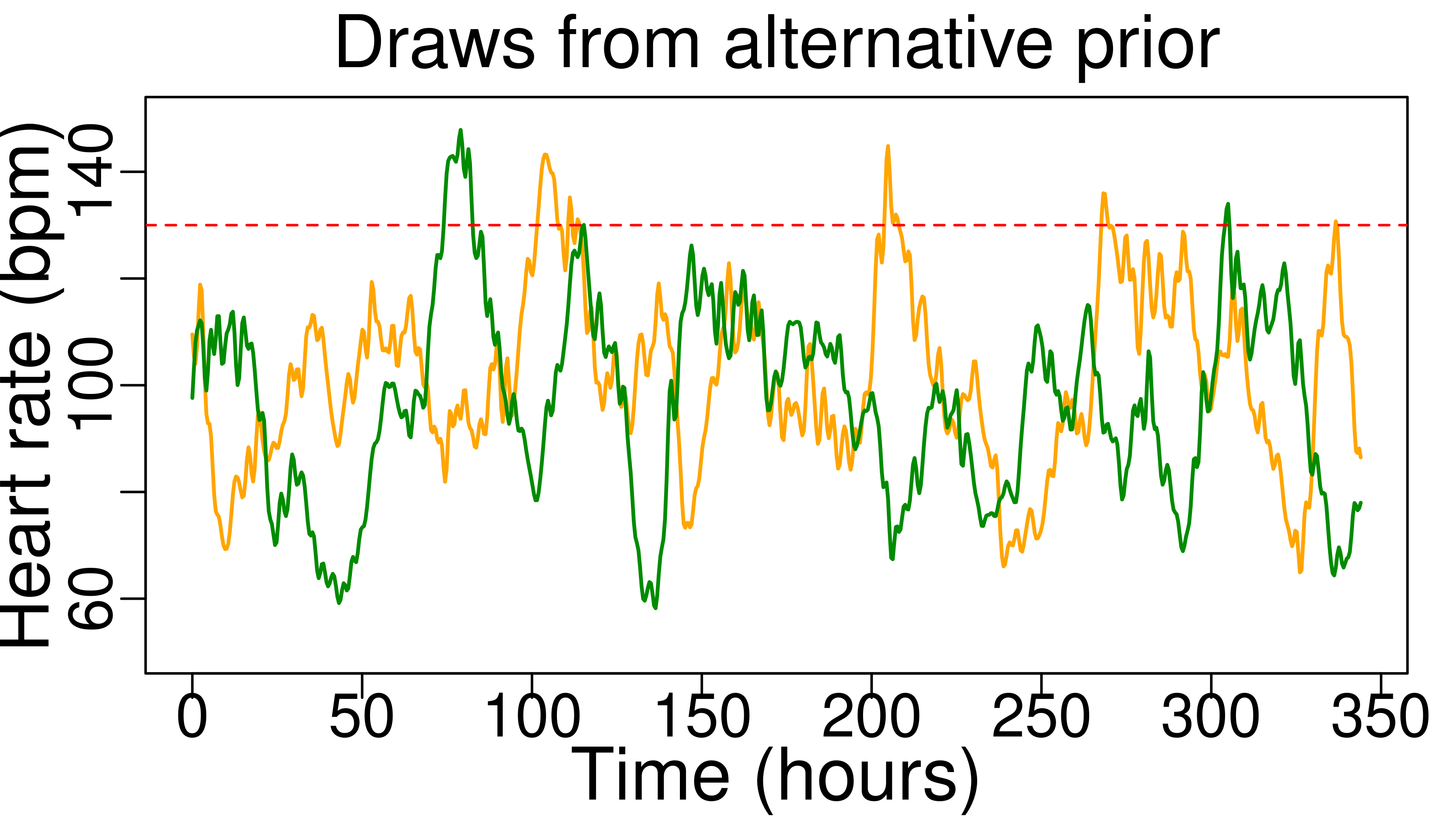} &
		\includegraphics[width=0.45\textwidth]{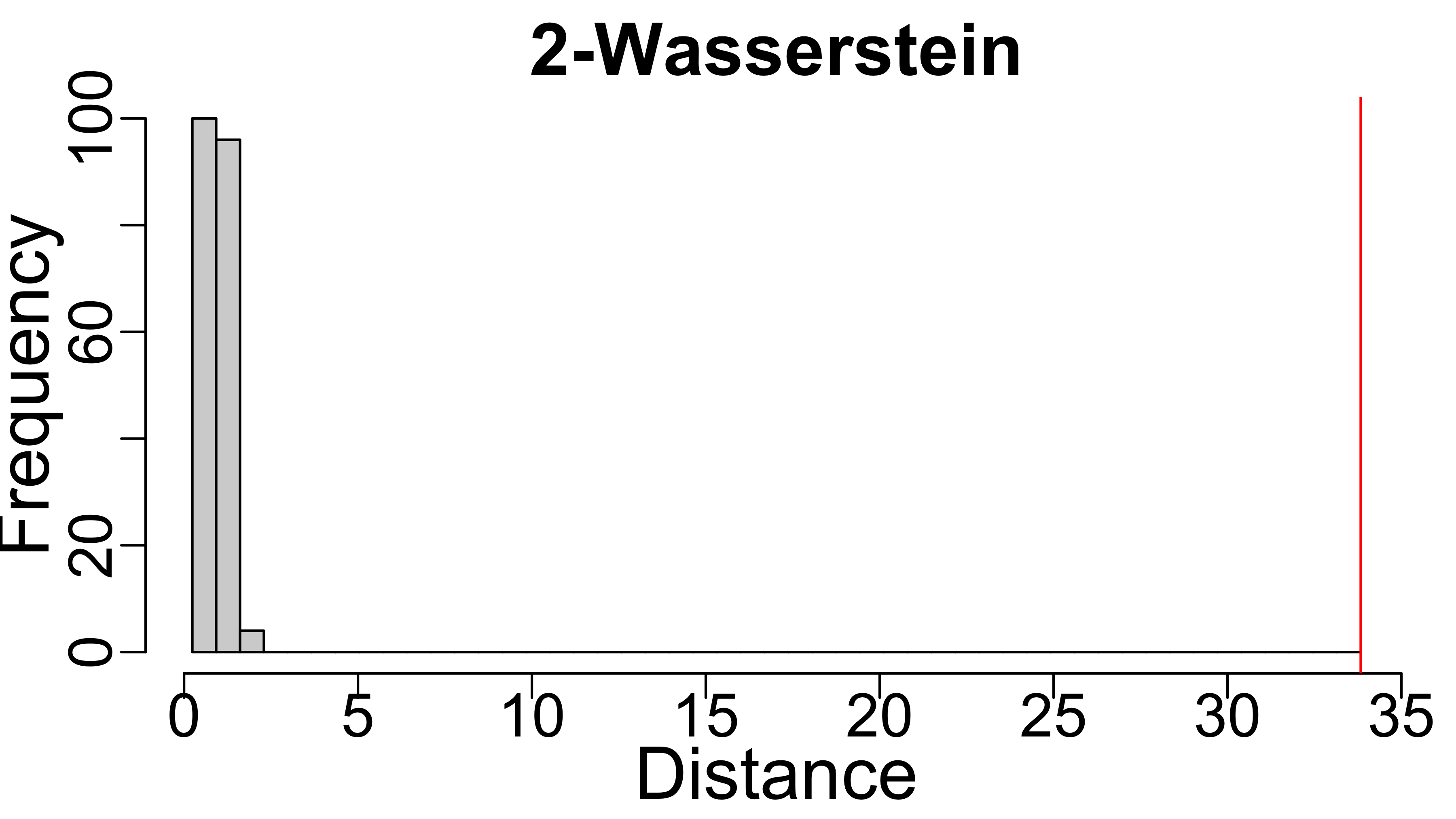}
	\end{tabular}
	\caption{Sensitivity of heart rate analysis in \cref{app:heartRate} for an example where we do not find non-robustness.
	(\emph{Top-left}): Heart rate data; notice the data is trending downwards at the end of the time series.
	(\emph{Top-right}): Prior draws from our original kernel $k_0$ from \cref{heartRateKernel}). (\emph{Bottom-left}): Prior draws from our decision-changing kernel $k_1$ that achieves $\Fstar = \bigChange$, noise matched by color to the draws from $k_0$.
	(\emph{Bottom-right}): Comparison of the difference between $k_0$ and $k_1$ (red line) to posterior hyperparameter uncertainty (histogram).}
	\label{fig:app-nonNonRobustHR}
\end{figure*}

Here, we give additional details for our heart rate modeling example from \cref{sec:heartRate}.
According to \cite{compCardiology1}, the data was collected under the approval of appropriate institutional review boards, and personal identifiers were removed. 
Following \cite{Colopy2016_identifying}, we first take the log transform of our heart rate observations $y_n$.
We then zero-mean the observations ($\sum_{n=1}^N y_n = 0$) and set them to have unit variance ($\sum_{n=1}^N y_n^2 = 1$).
The kernel used by \cite{Colopy2016_identifying} to model the resulting data log-scaled standardized data is a Mat\'{e}rn $5/2$ kernel plus a squared exponential kernel:
\begin{equation}
	k_0(x,x') = h_1^2 \left( 1 + \frac{\sqrt{5}\abs{x-x'}}{\lambda_1} + \frac{5\abs{x-x'}^2}{3\lambda_1} \right) \exp \left[ - \frac{\sqrt{5}\abs{x-x'}}{\lambda_1} \right]  
	+ h_2^2 \exp\left[ -\frac{\abs{x-x'}^2}{2\lambda_2^2} \right],
	\label{heartRateKernel}
\end{equation}
where $h_1, h_2, \lambda_1, \lambda_2 > 0$ are kernel hyperparameters, which we set via MMLE.
While all inferences are done on the zero-mean, unit-variance log-scaled data, all of our plots and discussion are given in the untransformed (i.e.\ raw bpm) scale for ease of interpretability.

In the main text, we showed an example where our workflow in \cref{alg:workflow} discovered non-robustness in predicting whether a patient's heart rate would be likely to be above 130 BPM or not 1.5 hours in the future.
We noted that there was some evidence in the data supporting this finding: the patient's heart rate was trending upward towards the end of the observed data, so we might expect that small changes to the prior could result in significant posterior mass being placed on high heart rates.
To demonstrate that we do not always find GP analyses non-robust to the choice of the prior, we give an example here where we do not find non-robustness.
For our example, we use a different patient from the Computing in Cardiology challenge \cite{compCardiology1,compCardiology2}.
The heart rate for this patient is plotted in \cref{fig:app-nonNonRobustHR}; notice that their heart rate is trending down at the end of the observed data.

As in \cref{sec:heartRate}, we use the constraint set and objective specified by \cref{constraintSet:stationary} (i.e.\ we constrain ourselves to stationary kernels with spectral densities close to the density of $k_0$). Following \cref{alg:workflow}, we solve \cref{mainOptimizationProblem} to find $\kperturb$ such that $\Fstar(\kperturb) = 
\bigChange$.
We then assess whether the recovered $\kperturb$ is qualitatively qualitatively interchangeable with $k_0$.
We plot noise-matched prior draws from $k_0$ and $\kperturb$ in \cref{fig:app-nonNonRobustHR}.
We see that $\kperturb$ has obvious qualitative deviations from $k_0$; the functions drawn from $\kperturb$ have noticably larger variance (count the number of times the functions from $\kperturb$ pass 130 bpm).
Additionally, we see in \cref{fig:app-nonNonRobustHR} that the 2-Wasserstein distance between $k_0(X,X)$ and $k_1(X,X)$ is much larger than the typical deviations around $k_0$ due to hyperparameter uncertainty.
We conclude that $k_0$ and $\kperturb$ are not qualitatively interchangeable.
Thus, we say that we do not find non-robustness in the sense of \cref{def:nonRobust}.
Again, this conclusion is fairly sensible: at the final observation, the patient's heart rate is below 80 BPM and is trending downwards.
It thus seems reasonable that it would take a somewhat unusual prior to predict that the patient's heart rate would suddenly spike to 130.

The two heart rate experiments in were run on a laptop with a six-core i7-9750H processor. 
The experiments took roughly five minutes each to complete.

\section{Additional details for \co2 experiment}
\label{app:co2}

\begin{figure*}
	\includegraphics[width = 0.485\textwidth]{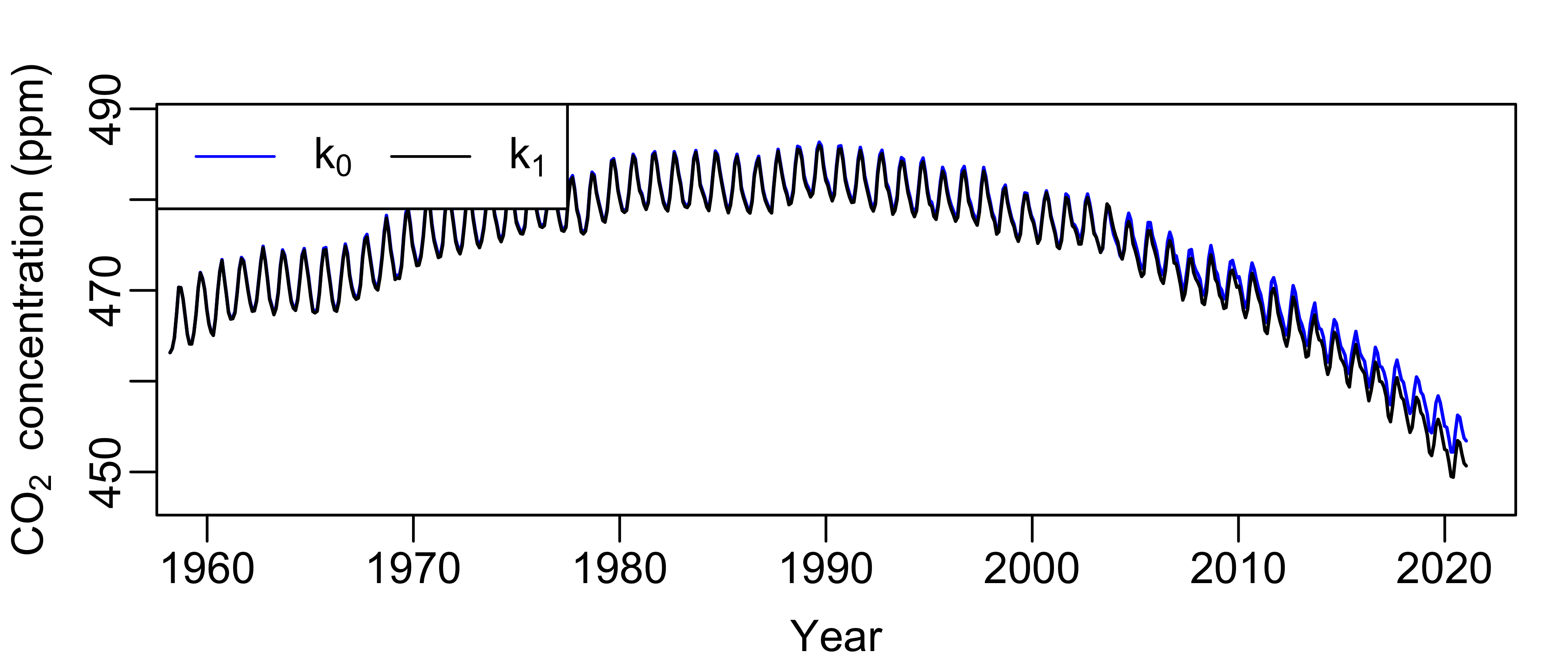}
	\includegraphics[width = 0.485\textwidth]{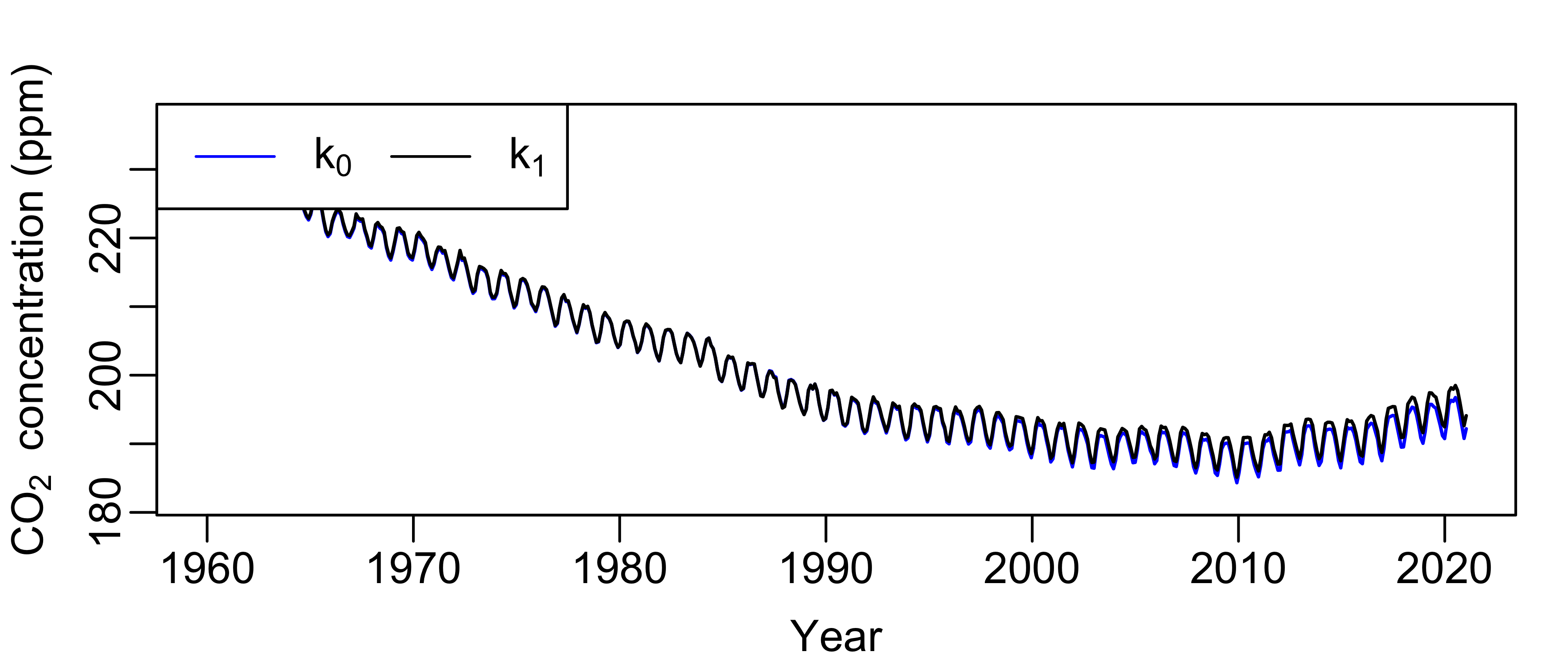}\\
	\includegraphics[width = 0.485\textwidth]{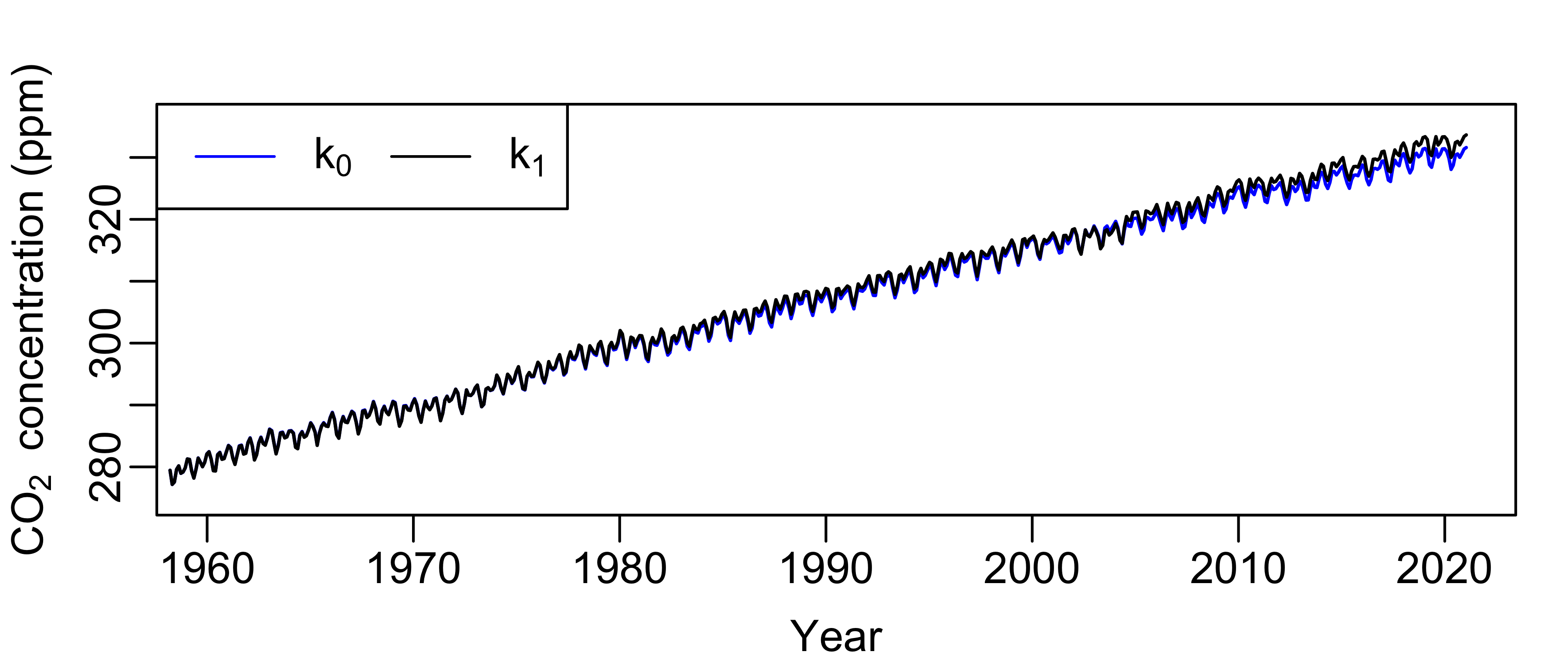}
	\includegraphics[width = 0.485\textwidth]{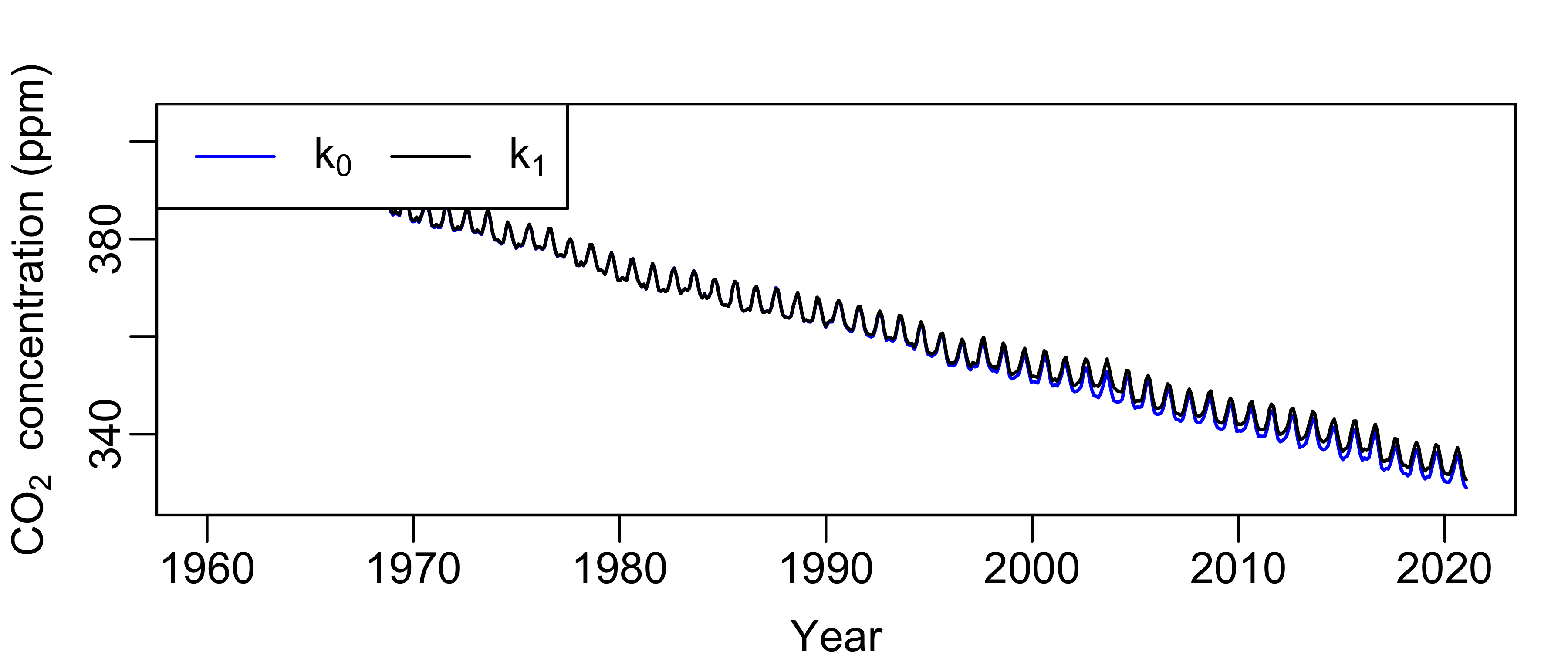} \\
	\includegraphics[width = 0.485\textwidth]{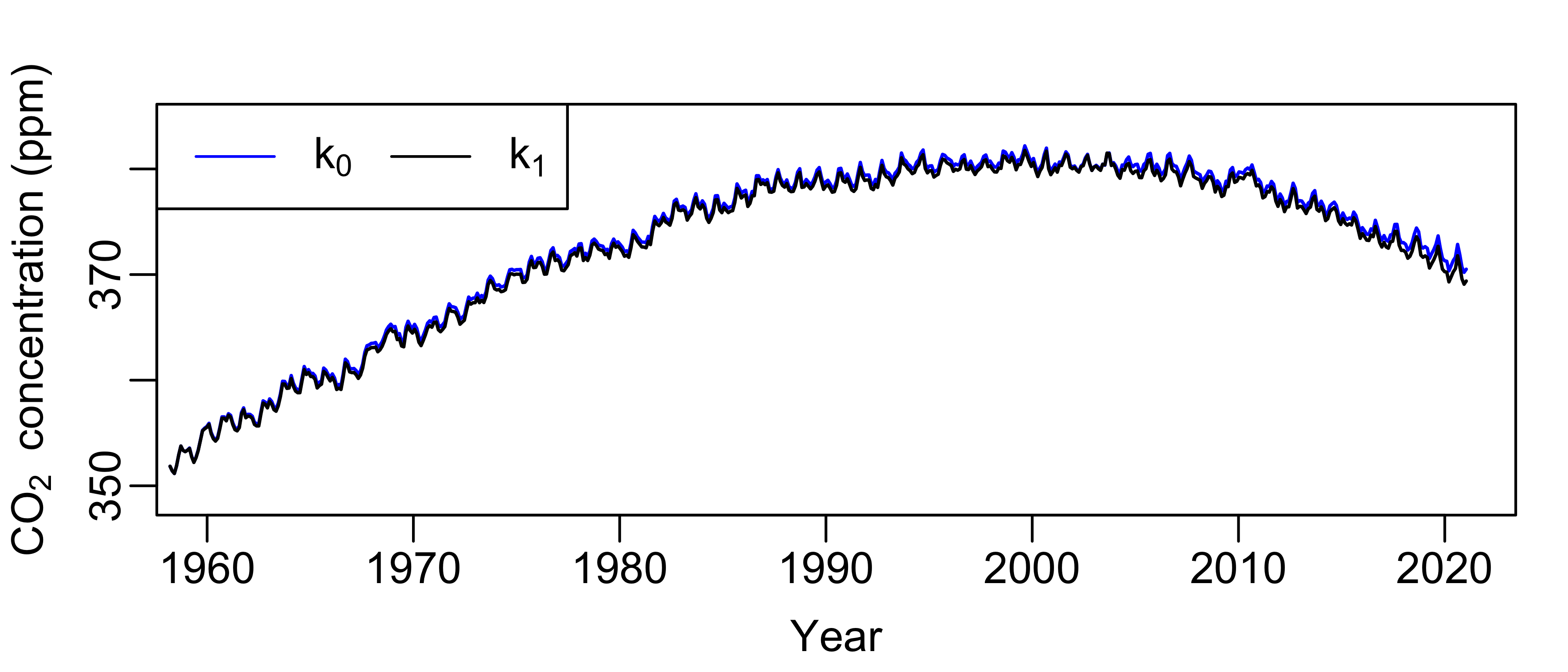}
	\includegraphics[width = 0.485\textwidth]{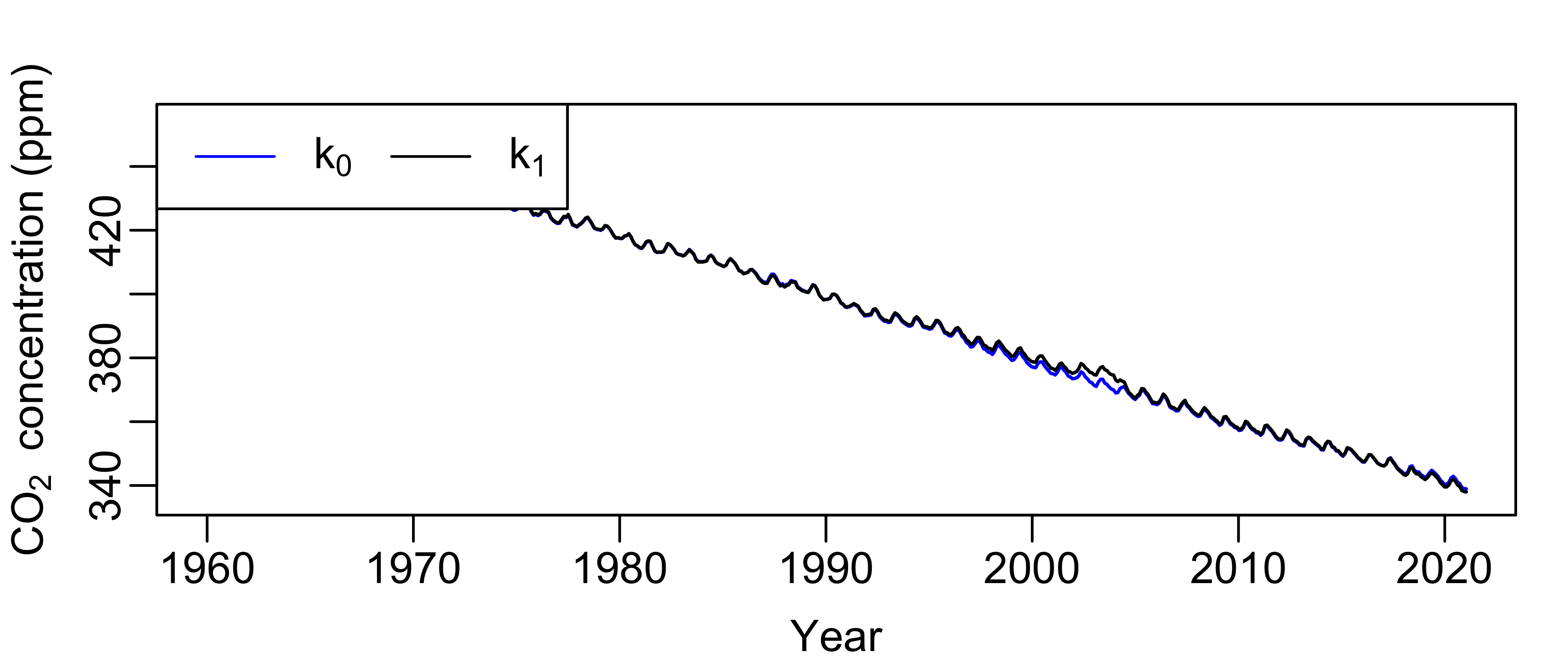} \\
	\includegraphics[width = 0.485\textwidth]{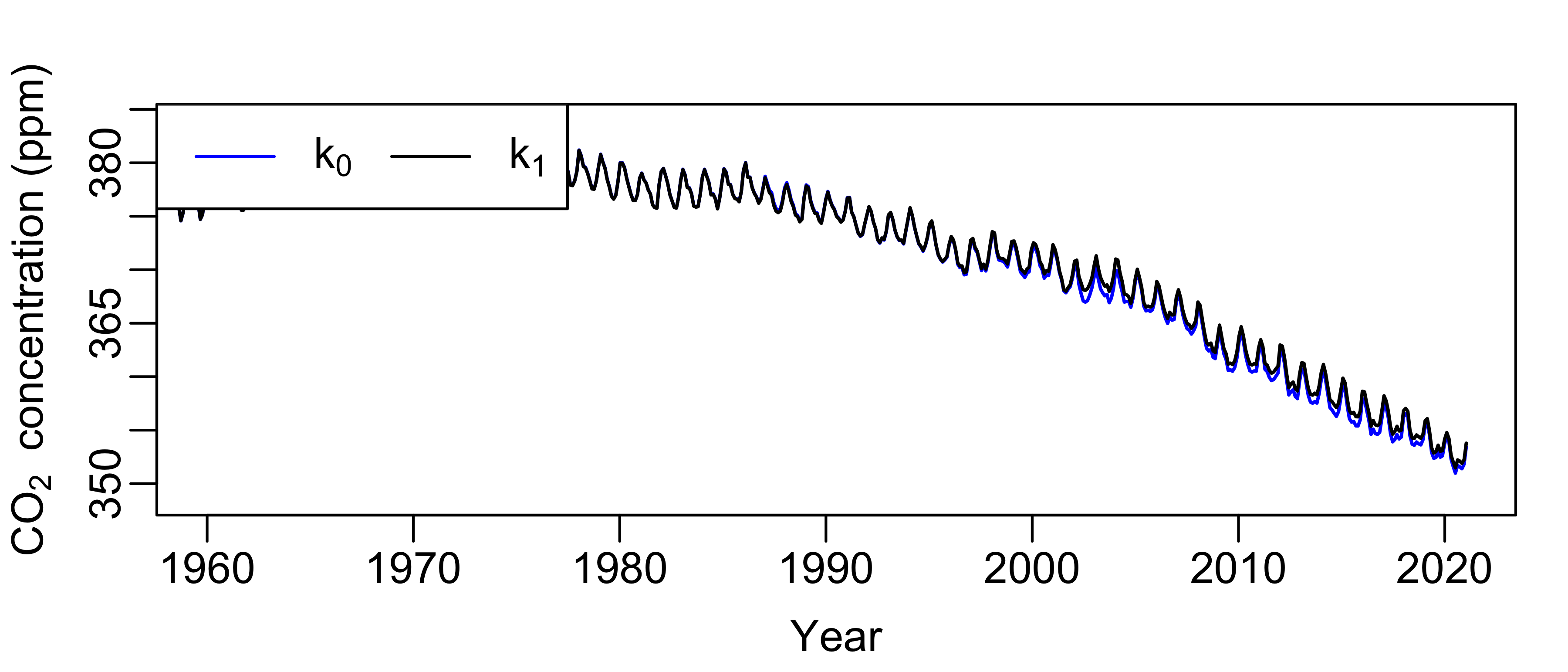}
	\includegraphics[width = 0.485\textwidth]{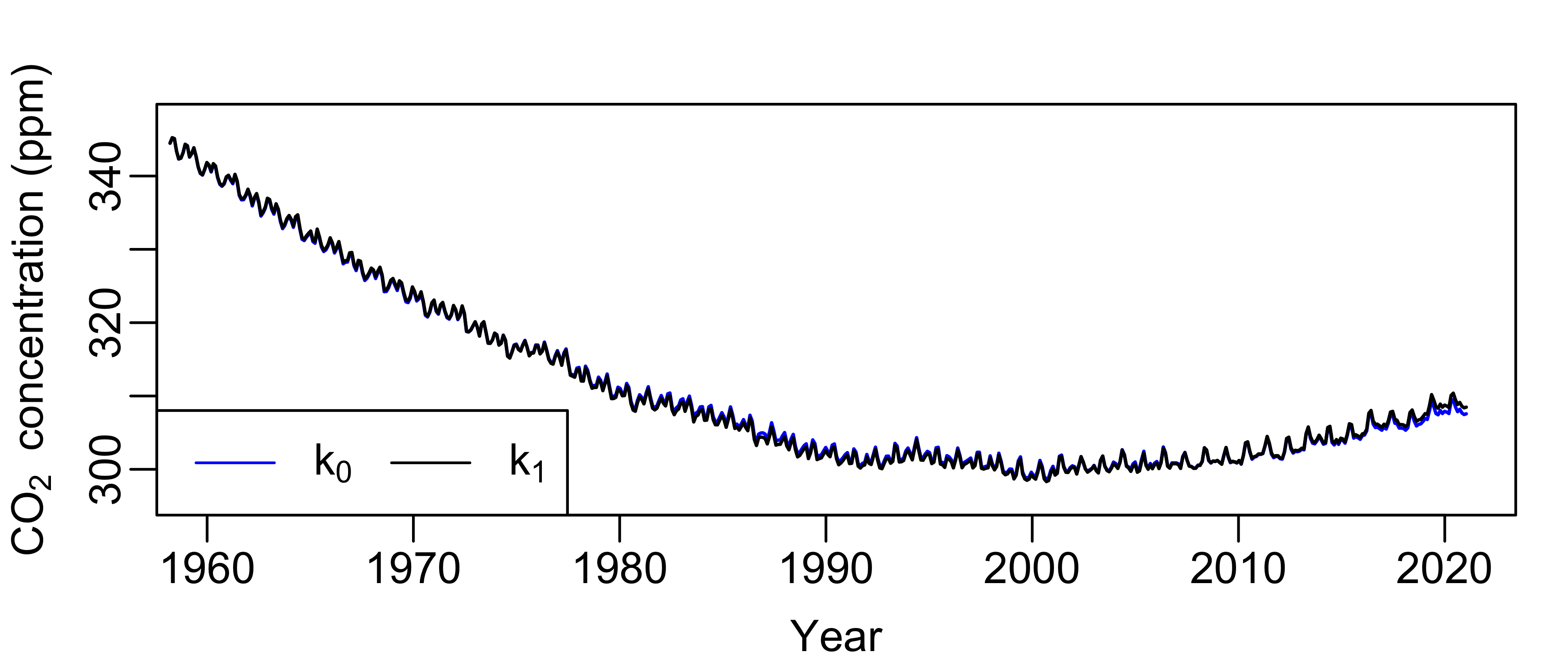} \\
	\includegraphics[width = 0.485\textwidth]{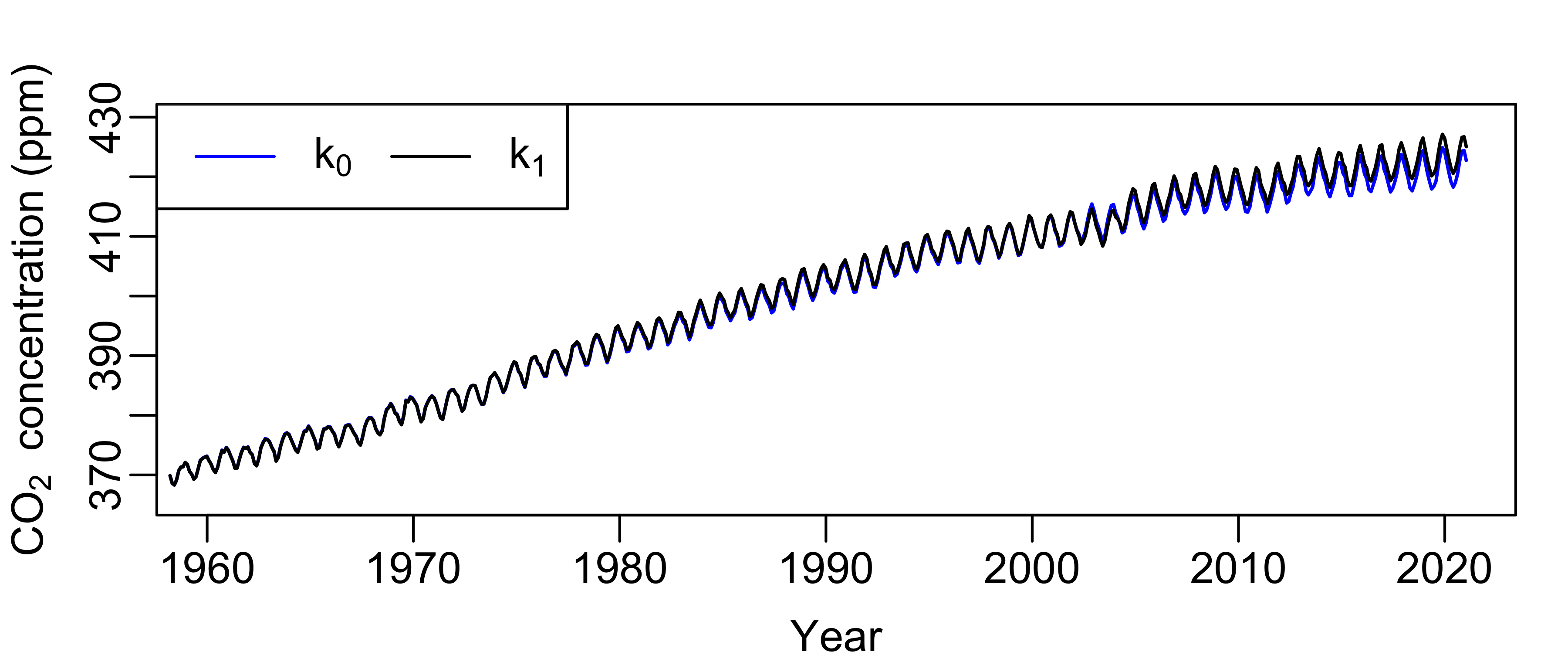}
	\includegraphics[width = 0.485\textwidth]{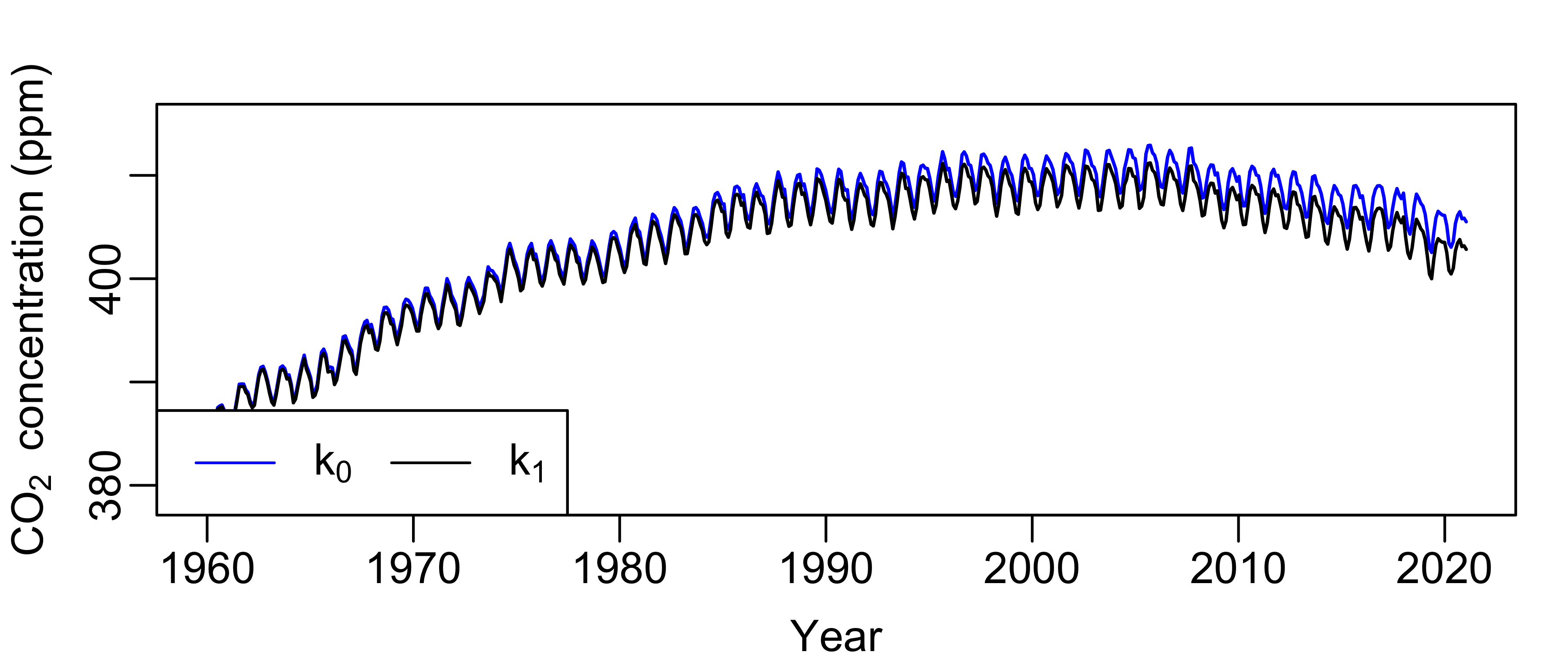}
	\caption{Sensitivity analysis of Mauna Loa. Each plot shows noise matched samples from a zero mean Gaussian process with original and perturbed kernel functions. These plots provide a zoomed in view of the prior samples shown in \cref{fig:MaunaLoa}. We note that draws from $\kperturb$ are in-phase with those of $k_0$ (i.e.\ $\kperturb$ captures the seasonal maxima and minima of \co2 just as well as $k_0$ does). Overall, there is high agreement between functions sampled from the two GPs.}
	\label{fig:app-maunaLoaPriors}
\end{figure*}

Here, we give additional details on the \co2 experiment from \cref{sec:co2}.
Our dataset is a series of monthly \co2 levels taken from Mauna Loa in Hawaii between 1958 and 2021 \citep{keeling2005atmospheric}; we download our data from \href{https://scrippsco2.ucsd.edu/assets/data/atmospheric/stations/in\_situ\_co2/monthly/monthly\_in\_situ\_co2\_mlo.csv}{https://scrippsco2.ucsd.edu/assets/data/atmospheric/stations/in\_situ\_co2/monthly/monthly\_in\_situ\_co2\_mlo.csv}.
\citet[Section 5.4.3]{RasmussenWilliams2006} predict future \co2 levels using a GP.
Their kernel is the sum of four terms:
\begin{align}
	k_0(x_1, x_2) &= \theta_1^2 \exp\left( -\frac{(x_1 - x_2)^2}{2\theta_2^2} \right) \\
	& + \theta_3^2 \exp\left( - \frac{(x_1-x_2)^2}{2\theta_4^2} - \frac{2\sin^2(\pi (x_1 - x_2))}{\theta_5^2} \right) \\
	& + \theta_6^2 \left( 1 + \frac{(x_1 - x_2)^2}{2\theta_7^2 \theta_8} \right)^{-\theta_8} \label{co2RationalQuadratic}\\
	& + \theta_9^2 \exp\left( - \frac{(x_1-x_2)^2)}{2\theta_{10}^2} \right),
\end{align}
where the $\theta_i$ comprise the kernel hyperparameters (in addition to the noise variance $\sigma^2$).
The different components of this kernel encode different pieces of prior knowledge.
The two squared exponentials encode long-term trends and small-scale noise, respectively.
The rational quadratic kernel (\cref{co2RationalQuadratic}) encodes small seasonal variability in \co2 levels between different years.
The periodic kernel captures the periodic trend in \co2 levels, which peak in the summer and reach their minimum in the winter.
This periodic is multiplied by a squared exponential to allow deviations away from exact periodicity.

Similar to \cite[Section 5.4.3]{RasmussenWilliams2006}, we first transform the training data by making the \co2 levels have zero mean.
To set the GP hyperparameters, we find that the hyperparameters values reported in \citet[Section 5.4.3]{RasmussenWilliams2006} are close, but not exactly, the MMLE solution on our data set (the gradient of the marginal log-likelihood has an entry substantively different from zero under the parameters from \citet{RasmussenWilliams2006}).\footnote{This problem might be due to the existence of slightly different versions of the Mauna Loa data set. 
The originally link for the data \url{http://cdiac.esd.ornl.gov/ftp/trends/co2/maunaloa.co2} is no longer responsive, for instance.} We set hyperparameters by $10$ random restarts of MMLE, where the solution iterates are initialized at the values reported in \cite[Section 5.4.3]{RasmussenWilliams2006}. The fitted values are $\theta_1 = 68.58$, $\theta_2 = 69.09$,  $\theta_3 = 2.55$, $\theta_4 = 87.60$, $\theta_5 = 1.44$,  $\theta_6 = 0.66$, $\theta_7 = 1.18$, $\theta_8 = 0.74$, $\theta_9 = 0.18$, $\theta_{10} = 0.13$, $\theta_{11} = 0.19.$ They are, for the most part, within $5\%$ of the values reported in \citet[Section 5.4.3]{RasmussenWilliams2006}. 

When \citet{RasmussenWilliams2006} ran their analysis, only data up to 2003 were available.
As it turns out, their analysis significantly underestimates current \co2 levels.
In particular, they fail to predict the fact that \co2 levels hit 415 ppm for the first time in human history in 2019; in fact, the maximum of the predicted \co2 levels in 2019 is over three posterior standard deviations away from 415 ppm.
We ask if a qualitatively interchangeable kernel could have changed this result.
Ideally, we would set $\Fstar$ to be the max of all posterior predictions in 2019. 
However, this is not a smooth function of the kernel. 
So we instead let $\Fstar$ to be the smooth max of the posterior means of all the test points in 2019, $\{\mu(x_t)\}_{t=1}^T$. The smooth-max we use is a scaled log-sum-exp, with a scale $\alpha > 0$:
$
	\Fstar(k) = \log \left( \sum_{t=1}^T e^{\alpha \mu(x_t)} \right) / \alpha.
$
Larger values of $\alpha$ provide a better approximation to the actual max function but may cause numerical difficulties; we choose $\alpha = 10$ as it seems to provide a reasonable approximation to the max function without 	introducing numerical problems. 
While we optimize using this approximation to the max, our experiments show that the recovered $\kperturb$ have an exact max prediction in 2019 of 415 ppm.

$k_0$ is stationary, so we could search for alternative stationary kernels using our spectral density framework from \cref{sec:nearbyKernels} (\cref{constraintSet:stationary}).
However, there is good reason to think we might want to consider non-stationary prior beliefs.
Developments in technology and/or global policy could have a large impact on \co2 levels.
Thus, we might encode past / expected future changes in technology and policy into our prior beliefs, making our prior beliefs non-stationary.

Thus, we use the input warping approach from \cref{sec:nearbyKernels} (\cref{constraintSet:nonStationary}).
However, we do not input warp the entirety of $k_0$.
As we know \co2 data has a regular periodicity, we leave the periodic component of the kernel, $\exp[ - 2\sin^2(\pi(x_1 - x_2)) / \theta^2_5 ]$ unwarped.
In preliminary experiments, we input warped the entirety of $k_0$; the resulting prior draws sometimes had minima in the summer and maxima in the winter, a clear violation of our prior knowledge about \co2 levels.
We input warp all other parts of $k_0$ using a use a two hidden layer fully connected network, with $50$ units and ReLU nonlinearities to parameterize $h$.
Finally, to ensure the optimal $k_1(\eps)$ is finite, we use $\ell(k; \Fstar, \bigChange) = (\Fstar(k) - \bigChange)^2$ in \cref{constraintSet:nonStationary}, which guarantees that our objective is bounded below.

We plot noise matched prior draws for $k_0$ and $\kperturb$ in \cref{fig:app-maunaLoaPriors}.
The samples from $\kperturb$ appropriately line up with the expected maxima and minima of \co2 levels (to see this, note that the draws from $\kperturb$ are in-phase with those from $k_0$, which correctly captures the seasonal maxima and minima).
The deviations between the noise-matched samples do not seem significant, so we say that $\kperturb$ and $k_0$ are qualitatively interchangeable.
Further, we find that the distance between $\kperturb$ and $k_{0}$ is smaller than what we might expect to arise from sampling uncertainty about $k_{0}$'s hyperparameters (see \cref{fig:mauna_loa_histograms} in \cref{app:histograms}).
We therefore conclude that the prediction of \co2 levels under $k_0$ is non-robust to the choice of the kernel in the sense of \cref{def:nonRobust}.

The Mauna Loa experiments were run on a laptop with a 2.3 GHz 8-Core Intel Core i9, with 64 GB of RAM. The experiment (which optimizes from five random initialization) took about 15 minutes to run, with each seed taking about 3 mins.

\section{\co2 experiments using the automatic statistician}
\label{app:co2_ac}
Here we use the kernel learned by the automatic statistician~\cite{Duvenaud2013_compositional} to model the Mauna-Loa \co2 data. The automatic statistician kernel is:
\begin{align}
	k_0(x_1, x_2) &= \left(\theta_1^2 + \theta_2^2(x_1 - \theta_3)(x_2 - \theta_3)\right) \times \theta_4^2 \exp\left( -\frac{(x_1 - x_2)^2}{2\theta_5^2} \right) \\
	& + \theta_6^2 \exp\left(- \frac{2\sin^2(\pi (x_1 - x_2)/\theta_7)}{\theta_8^2}\right)  \times \theta_9^2 \exp\left( -\frac{(x_1 - x_2)^2}{2\theta_{10}^2} \right)\\
	& + \theta_{11}^2 \left( 1 + \frac{(x_1 - x_2)^2}{2\theta_{12}^2 \theta_{13}} \right)^{-\theta_{14}} \times \theta_{15}^2 \exp\left( - \frac{(x_1-x_2)^2)}{2\theta_{16}^2} \right).\label{co2RationalQuadratic}\end{align}

\begin{figure*}
\centering
%	\begin{tabular}{lll}
		\includegraphics[width=0.485\textwidth]{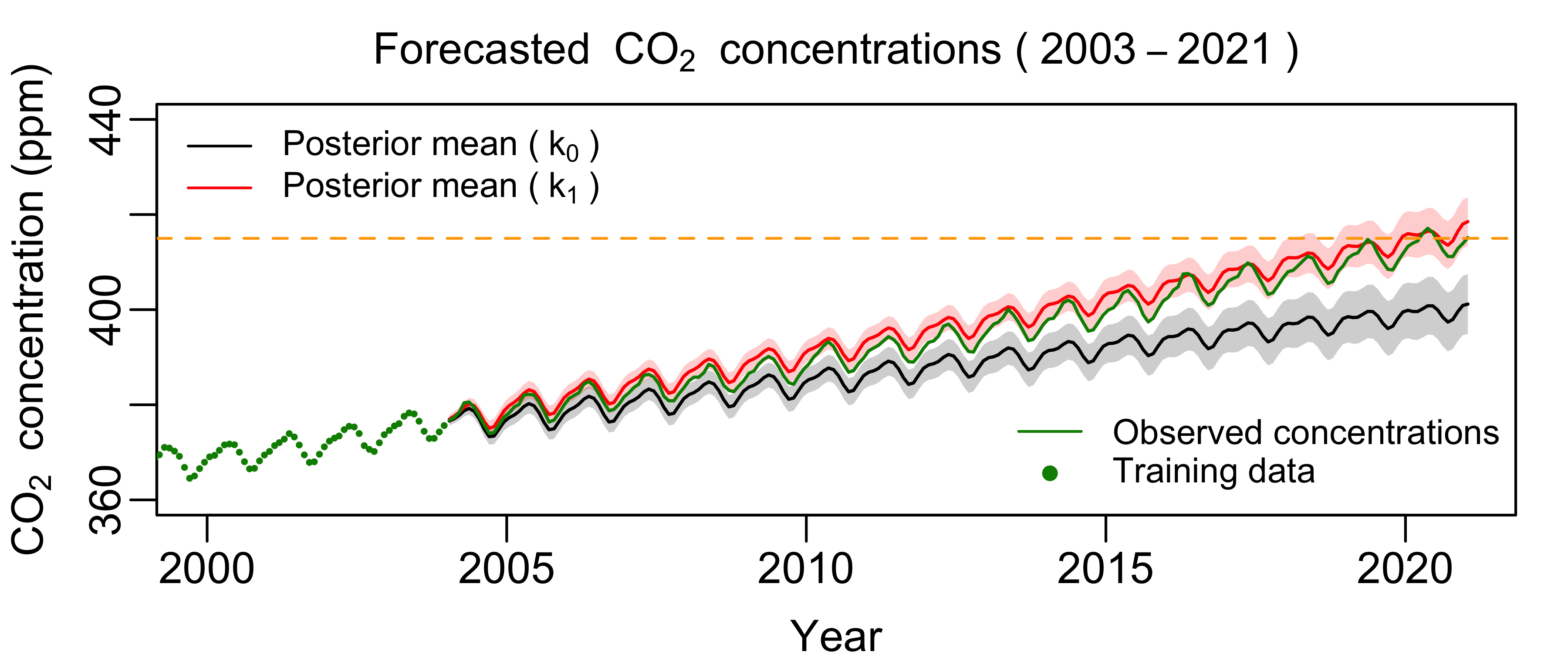}
		\includegraphics[width = 0.485\textwidth]{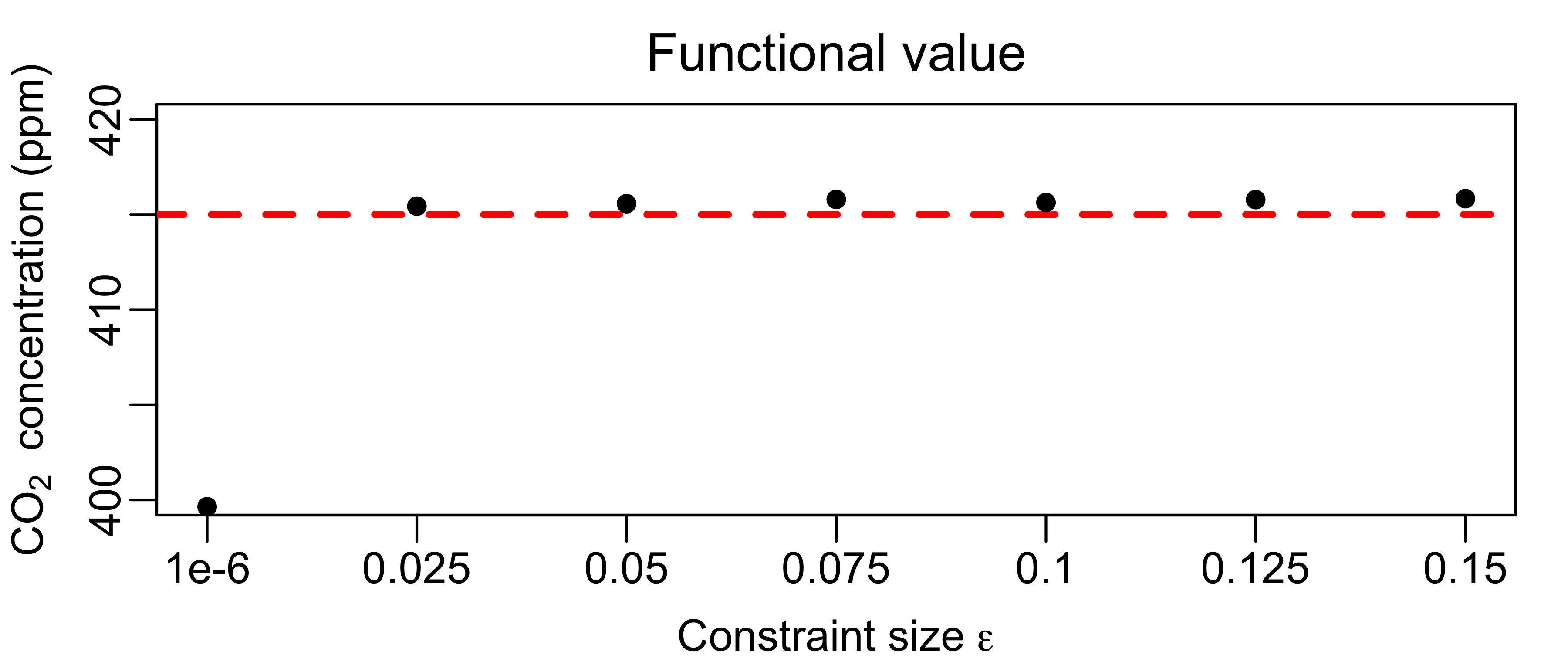}

		\includegraphics[width=0.485\textwidth]{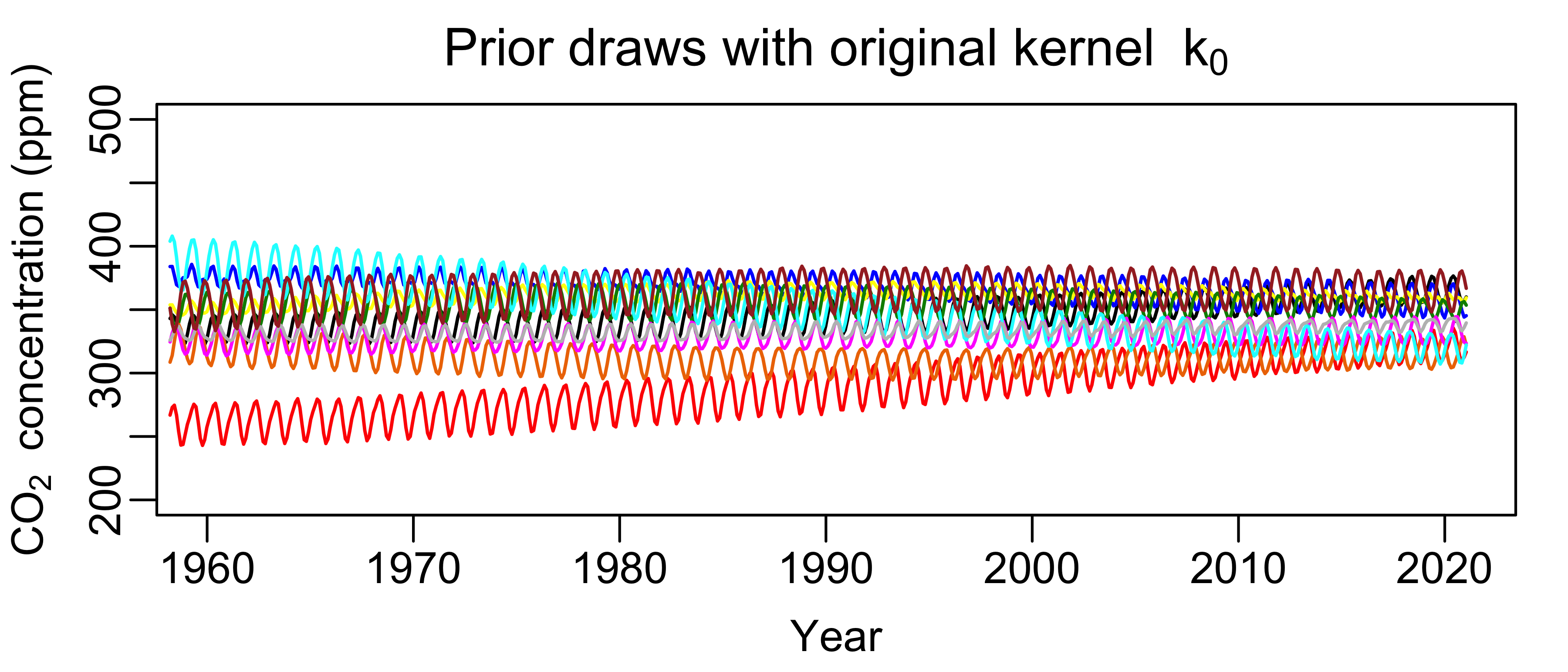}
		\includegraphics[width=0.485\textwidth]{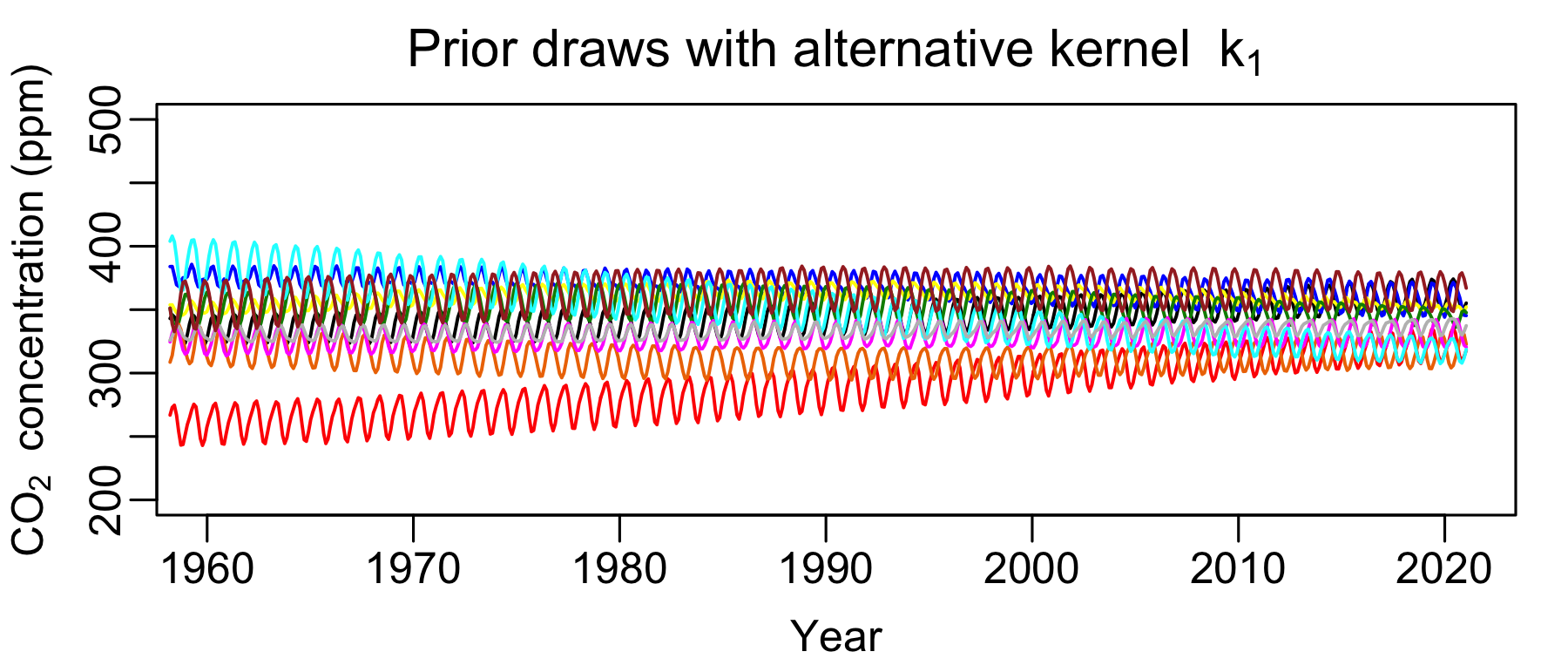}
%	\end{tabular}
\caption{\small{Sensitivity of the Mauna Loa analysis in \cref{app:co2_ac}. (\emph{Top-left}): Predictions made with the automatic statistician kernel $k_{0}$ (black) and a qualitatively interchangeable kernel $k_{1}$ (red). (\emph{Top-right}): $\Fstar$, the mean \co2 level in June 2020, as a function of $\eps$. (\emph{Bottom}): Noise-matched draws from a $\GP(0,k_{0})$ (\emph{left}) and $\GP(0,\kperturb)$ (\emph{right}) prior. See \cref{fig:automatic_app-maunaLoaPriors} for a closer inspection of each prior draw.}}
	%\caption{\wts{Get these panels to be labeled w/ a, b, c, d like in other figs} Sensitivity analysis of the Mauna Loa  \cref{sec:co2}. (\emph{Top}): Predictions under the original (black) and the perturbed (red) kernels from 2003 to present. (\emph{Bottom-left and Bottom-center}): Noise-matched prior draws from $k_0$ and $k_1$, respectively. See \wts{appendix} for a closer inspection of each prior draw. (\emph{Bottom-right}): $\Fstar$ -- the mean \co2 level in June 2020 -- as a function of constraint set size and prior draws from the original and perturbed kernels.}
	\label{fig:automatic_MaunaLoa}
\end{figure*}

We learned the hyper-parameters of the kernel by maximizing the marginal likelihood on data up till $2003$, mimicking the process used for learning the parameters of the hand designed kernel described in \cref{app:co2}. \cref{fig:automatic_MaunaLoa} presents analogous results to those presented in \cref{fig:MaunaLoa} for the hand designed kernel. 

\begin{figure*}
	\includegraphics[width = 0.485\textwidth]{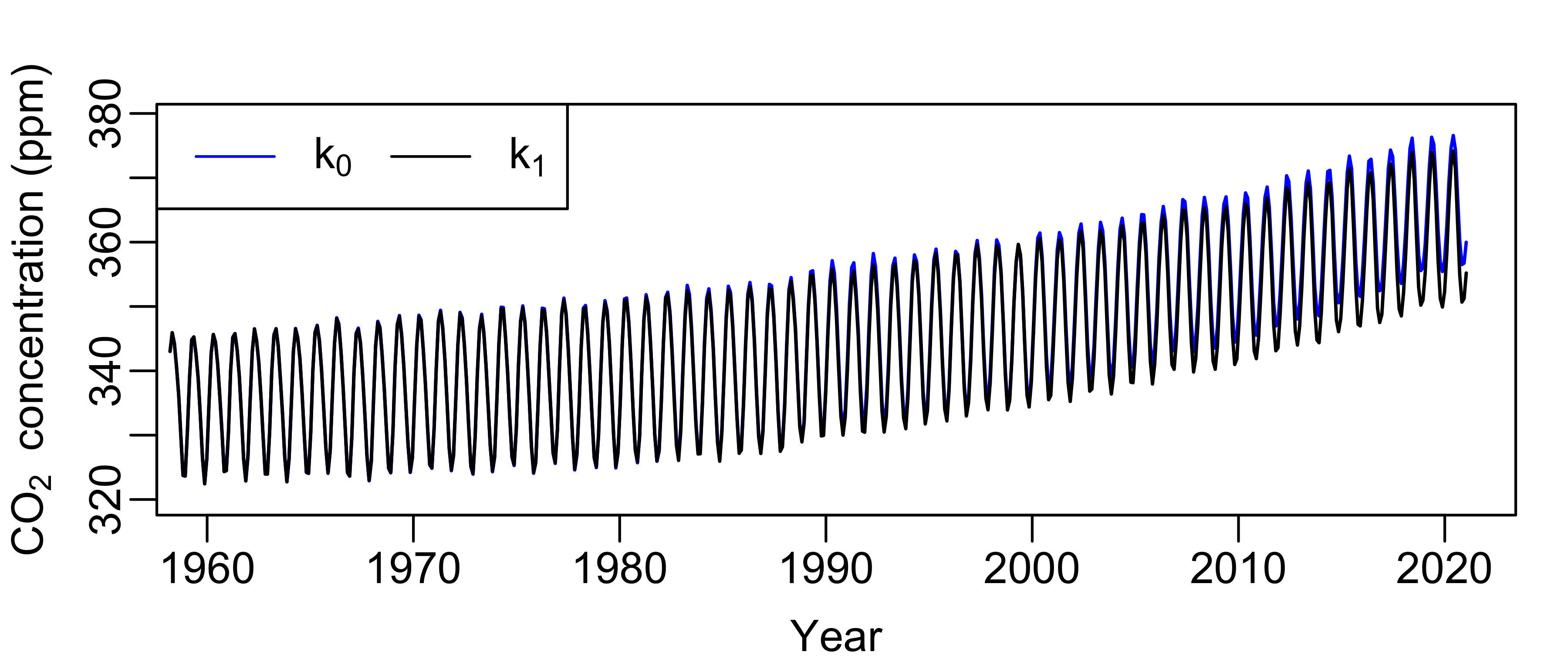}
	\includegraphics[width = 0.485\textwidth]{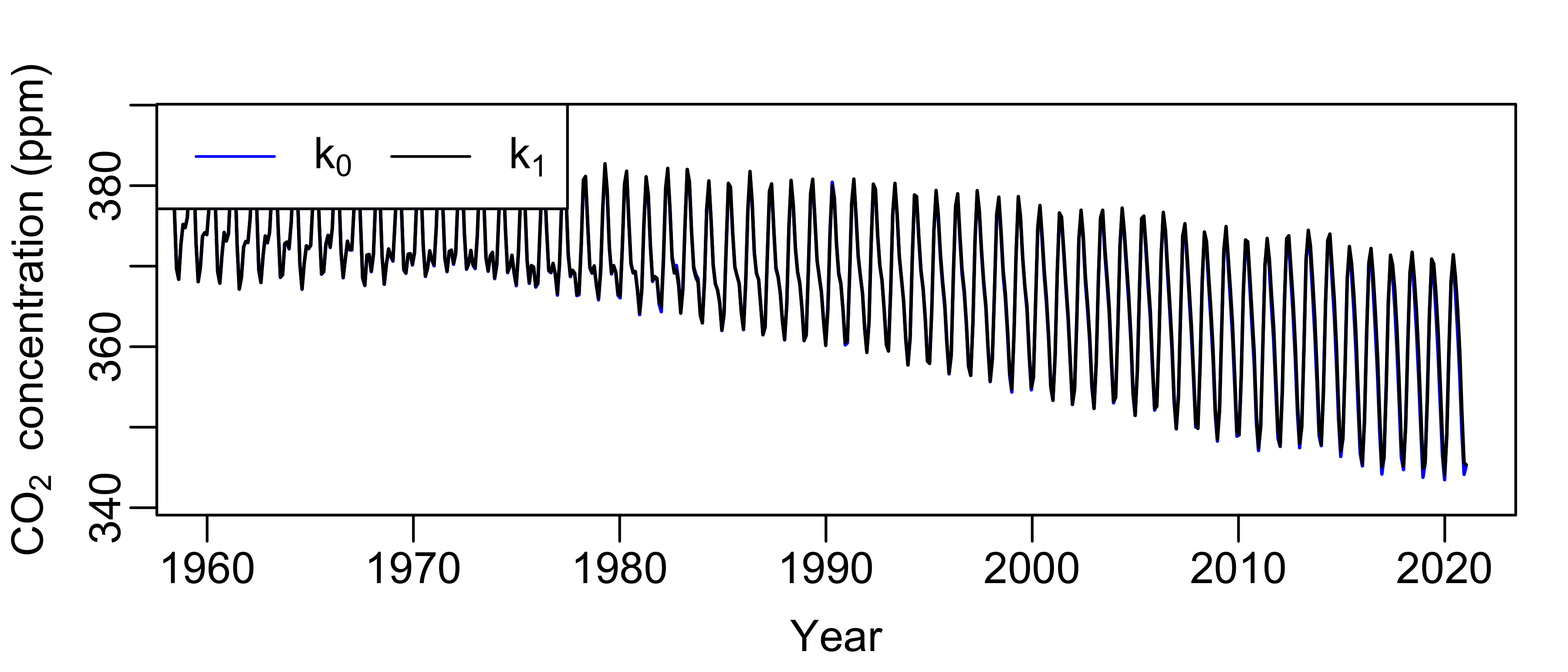}\\
	\includegraphics[width = 0.485\textwidth]{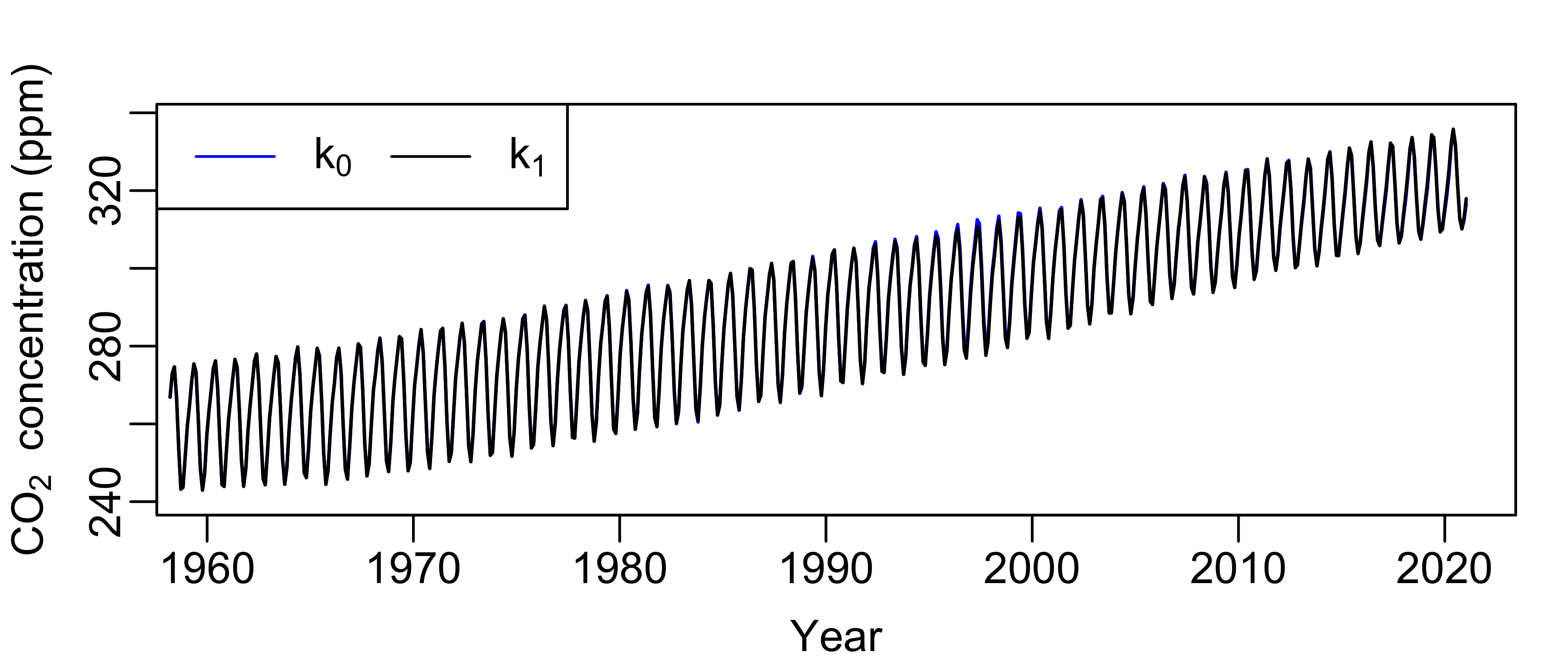}
	\includegraphics[width = 0.485\textwidth]{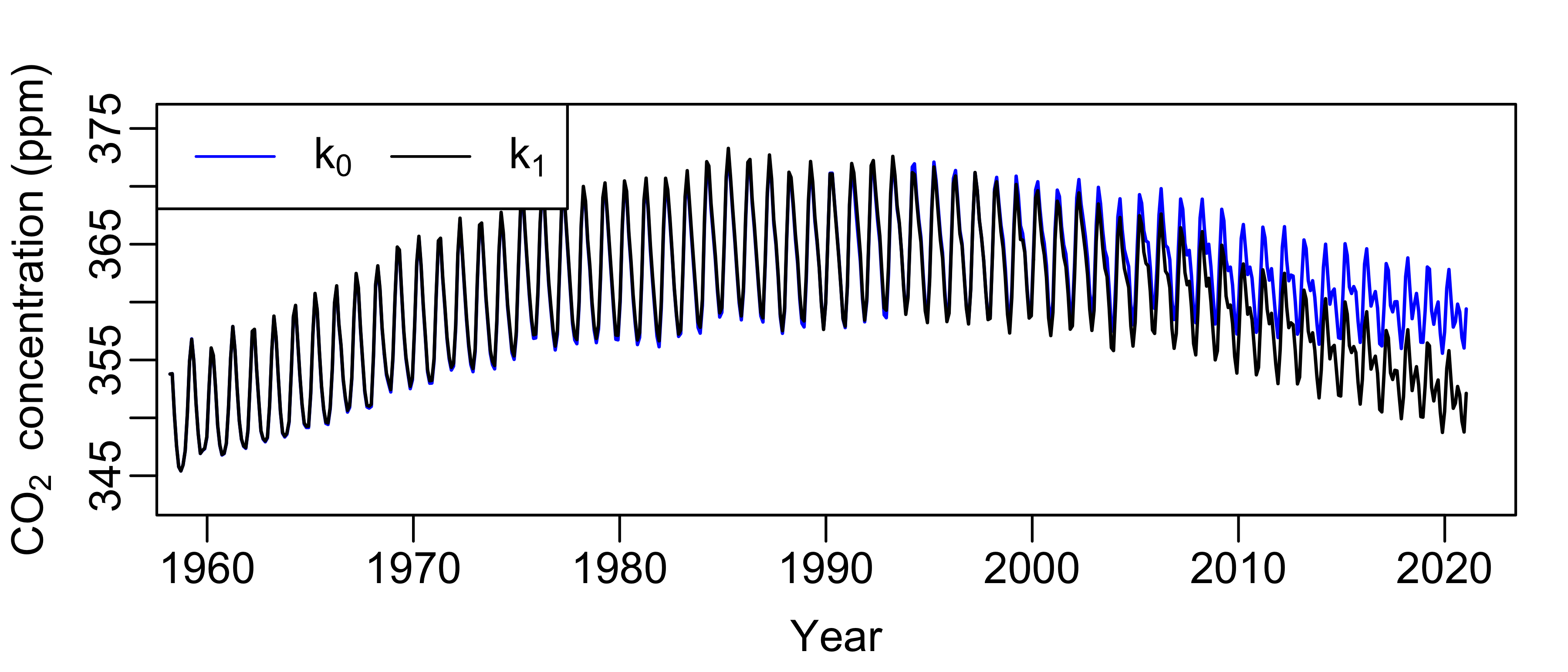} \\
	\includegraphics[width = 0.485\textwidth]{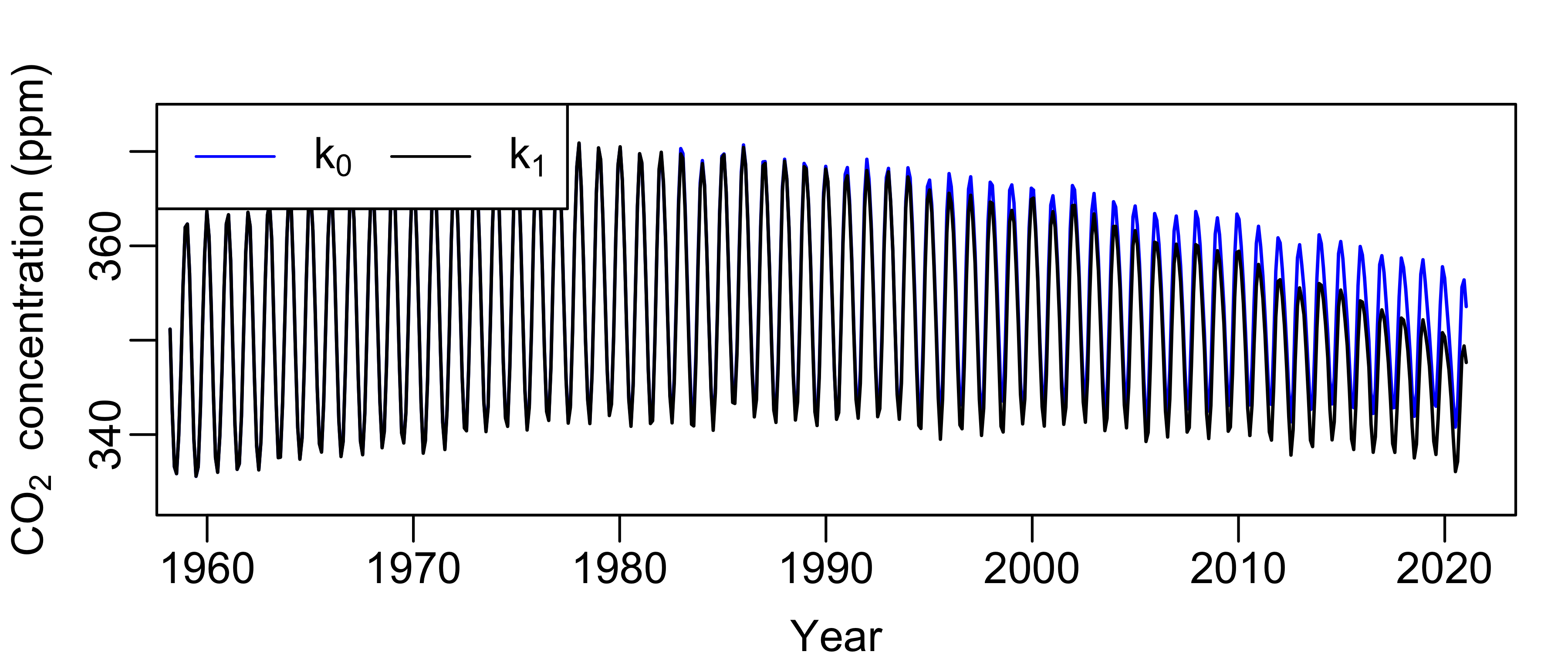}
	\includegraphics[width = 0.485\textwidth]{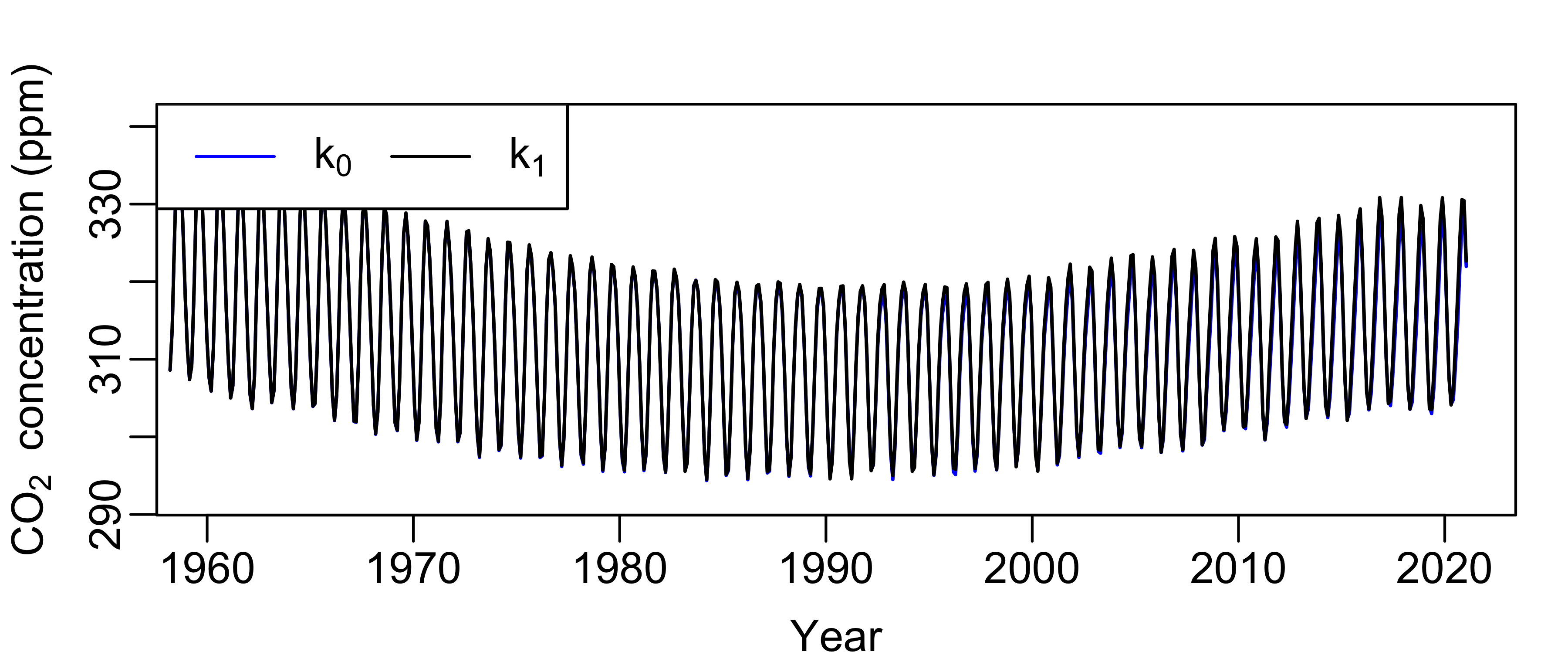} \\
	\includegraphics[width = 0.485\textwidth]{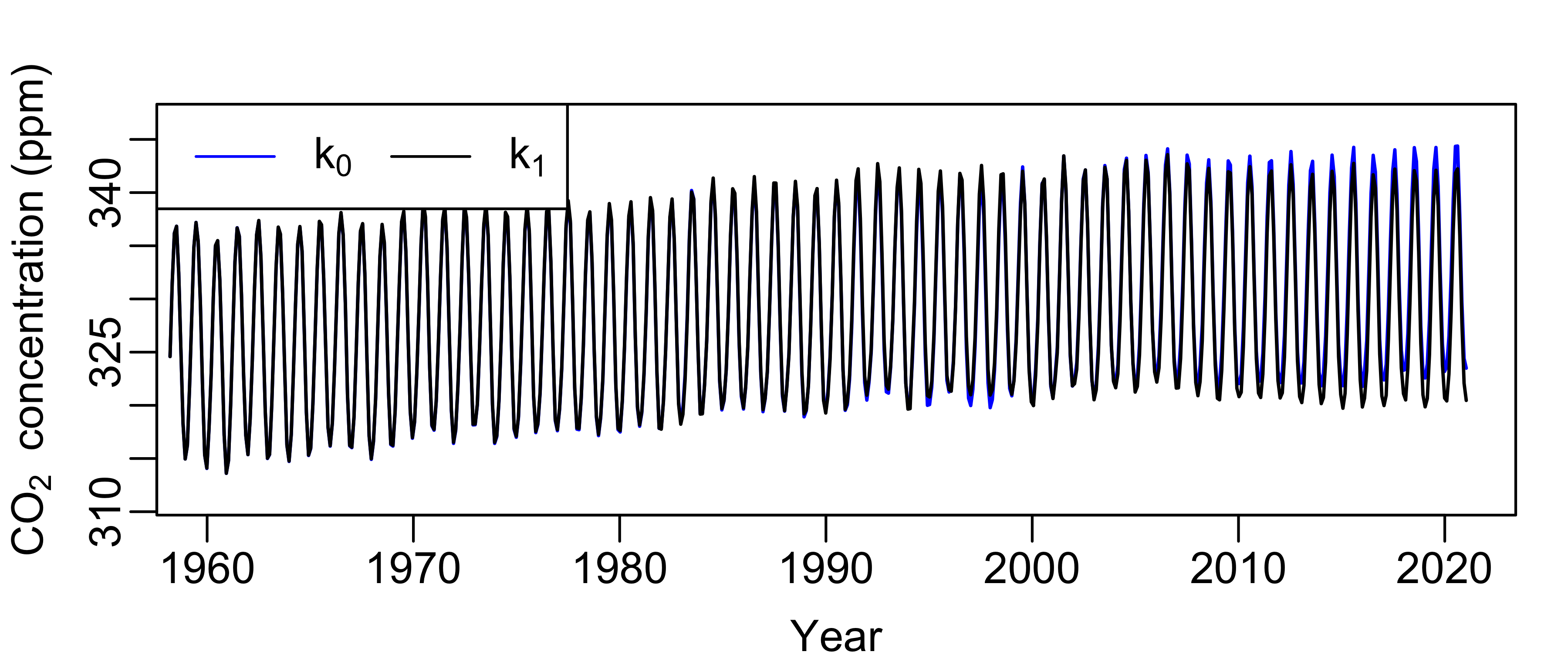}
	\includegraphics[width = 0.485\textwidth]{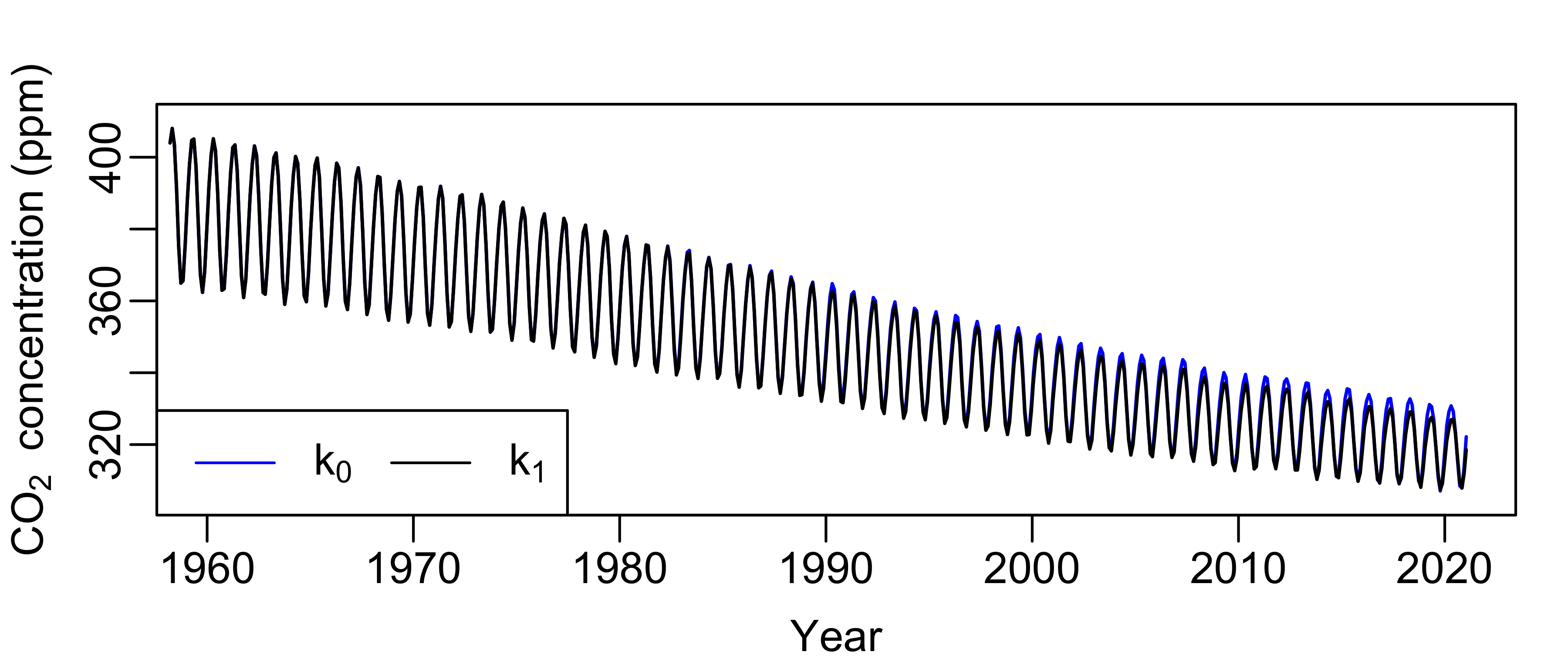} \\
	\includegraphics[width = 0.485\textwidth]{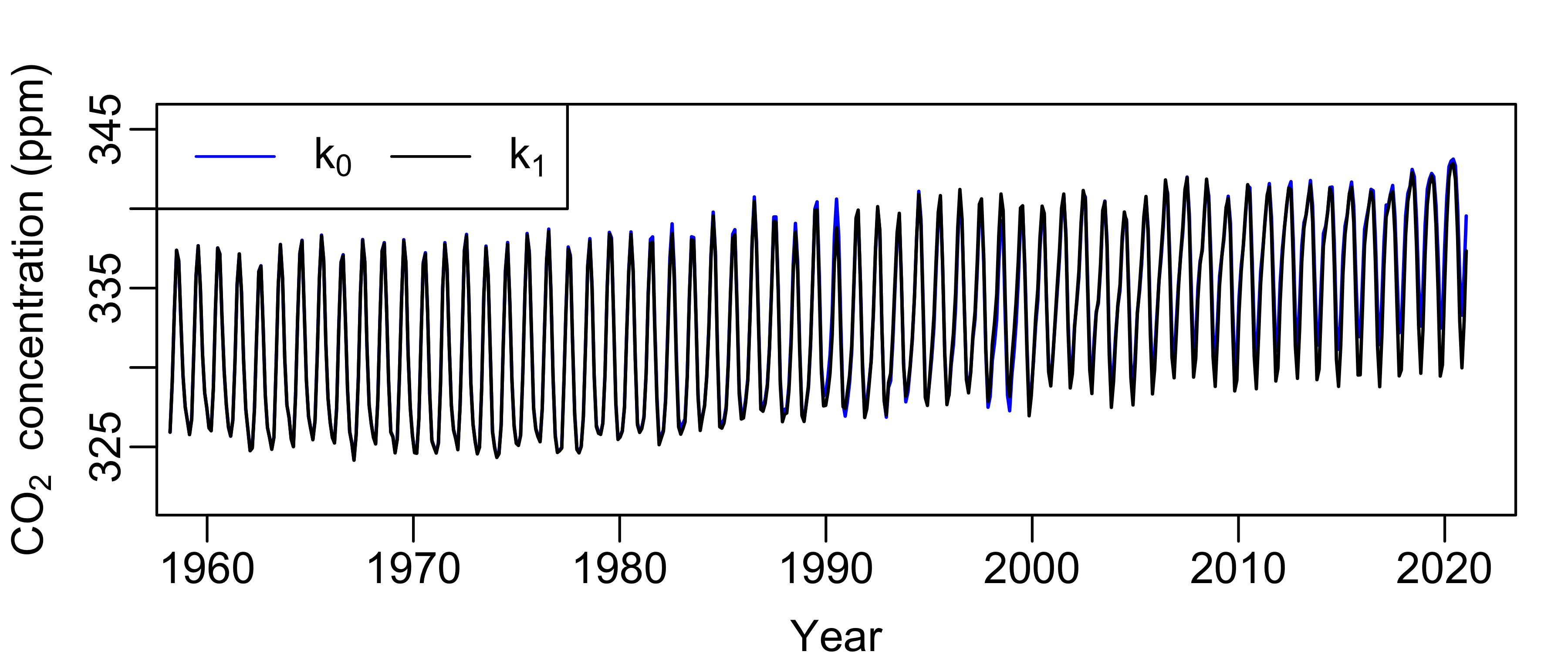}
	\includegraphics[width = 0.485\textwidth]{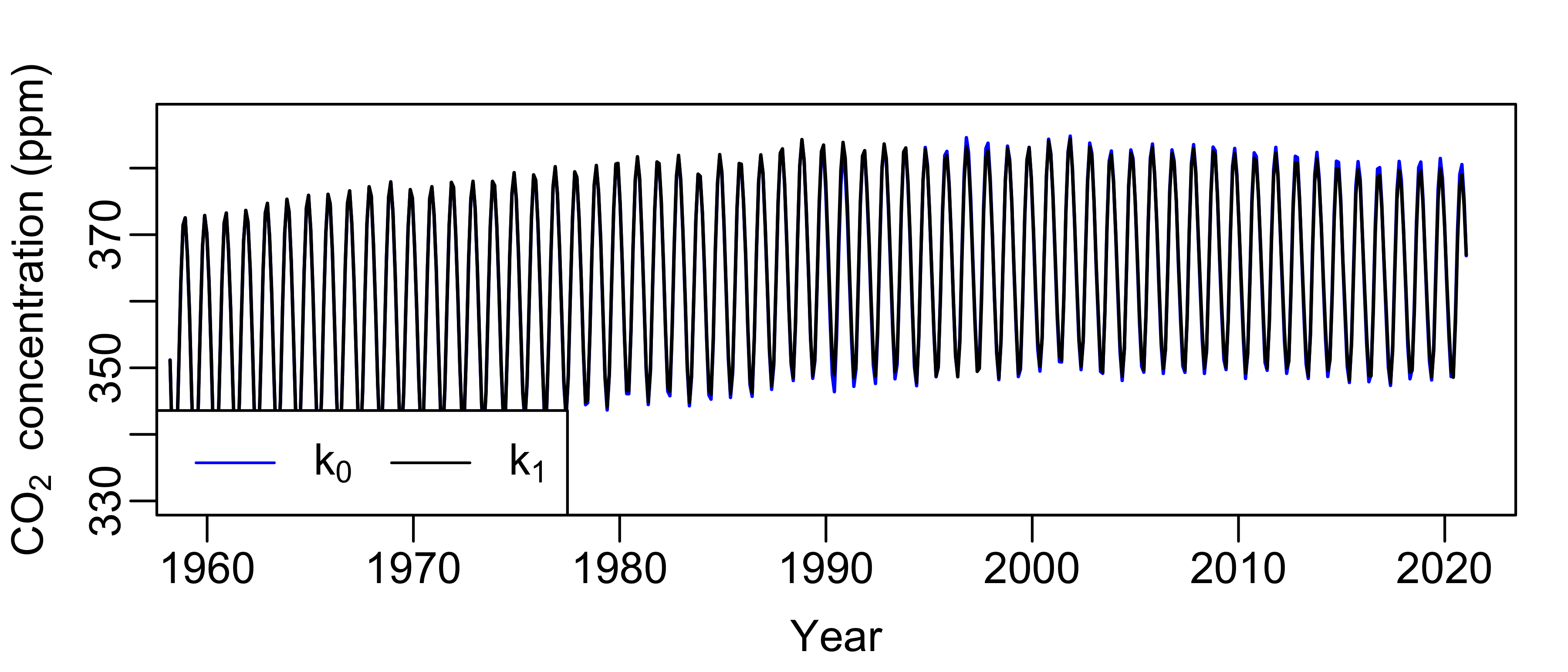}
	\caption{Sensitivity analysis of Mauna Loa. Each plot shows noise matched samples from a zero mean Gaussian process with original and perturbed kernel functions. These plots provide a zoomed in view of the prior samples shown in \cref{fig:automatic_MaunaLoa}. We note that draws from $\kperturb$ are in-phase with those of $k_0$ (i.e.\ $\kperturb$ captures the seasonal maxima and minima of \co2 just as well as $k_0$ does). Overall, there is high agreement between functions sampled from the two GPs.}
	\label{fig:automatic_app-maunaLoaPriors}
\end{figure*}

\section{More details on MNIST experiments}
\label{app:MNIST}

\begin{figure*}[t]
\includegraphics[width=\textwidth]{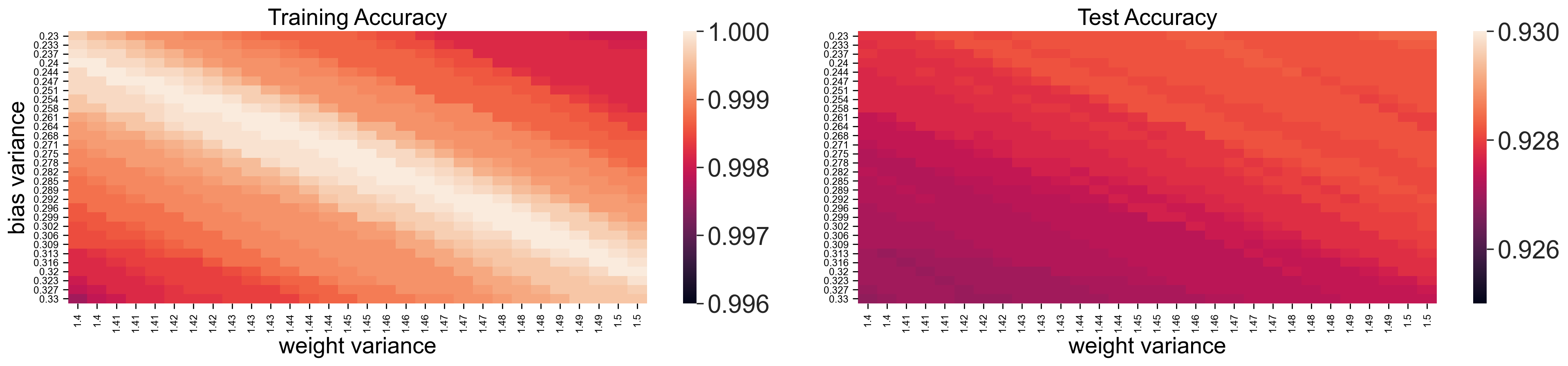}
\caption{Additional MNIST experiments. Here we visualize the training and test set performances along the hyperparamter grid used for assessing qualitative interchangeability. The train and test accuracies exhibit high performance and low variability across the grid.}	
\label{fig:app_minist_grid}
\end{figure*}
\begin{figure}
\centering
\includegraphics[width=0.45\textwidth]{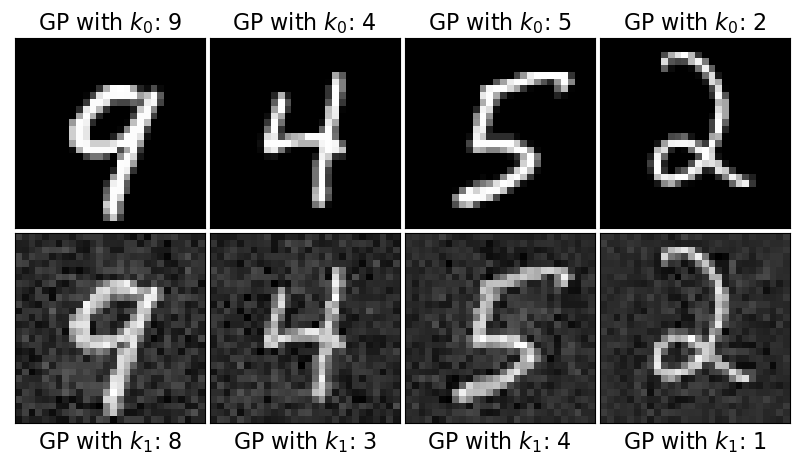}
\caption{Example test images from \cref{sec:MNIST}. $\xtest$ (upper) and their warps $g(\xtest)$ (lower) and predicted class labels (above and below).}
\label{fig:MNIST2}
\end{figure}

We use the publicly available \texttt{neural-tangents}~\citep{neuraltangents2020} package for constructing the kernels in our MNIST experiments.
We follow the experimental setup of \cite{Lee2018_dnn_gp} where the authors use a Gaussian process with a kernel corresponding to a $20$ layer, infinitely wide, fully connected, deep neural network with ReLU non-linearities. They place zero mean Gaussian priors over the weights, $\mathcal{N}(0, \sigma^2_w)$, and biases, $\mathcal{N}(0, \sigma^2_b)$, and set the hyper-parameters $\sigma^2_w = 1.45$ and $\sigma^2_b = 0.28$ via a grid search over parameters to maximize held-out predictive performance. \citet{Lee2018_dnn_gp} use a GP with $C=10$ outputs (classes). They pre-process one-hot encoded output vectors to have zero mean, i.e. $y_{ic} = 0.9$ if $c$ is the correct class for the $i$th training point, and $y_{ic} = -0.1$ for all incorrect classes; input images are flattened and an overall mean is subtracted from every image. Test prediction is made by selecting a class corresponding to the GP output with mean closest to $0.9$. The resulting GP trained on one thousand images from the MNIST training set and evaluated on the MNIST test set achieves an accuracy of $92.79\%$. 

In our experiments we assess the robustness of their kernel. The $28 \times 28$ MNIST images require a warping function $g : \R^{784} \rightarrow \R^{784}$. We use a fully connected multi-layer perceptron with one $784$ unit hidden layer, $784$ input, and $784$ output units with ReLU non-linearities to parametrize $g$. Let $c_0$ be the prediction under the original kernel at a target test image $\xtest$. We define $c_1 := |c_0 - 1|$ and create a ``fake'' output $y^*$ with $y^*_{c_1} = 0.9$ and $y^*_{c} = -0.1$ for $c \neq c_1$. We find parameters of $g$ by minimizing the objective in Algorithm \ref{alg:non-stationary} plugging in

\begin{equation}
\label{eq:sup:l-mnist}
\ell(k; \Fstar, \bigChange) = -\frac{1}{C}\sum_{c=1}^{10} \log p(y^*_c | X, \xtest, Y),
\end{equation}

i.e. the negative log-likelihood of the ``fake'' output at a particular test image $\xtest$ under the perturbed kernel; $X$ and $Y$ are the train inputs and outputs. As we discussed in the main text, directly optimizing the posterior quantity of interest $\Fstar = |\mu_{c_{0}}(\xtest) - 0.9| - |\mu_{c_{1}}(\xtest) - 0.9|$ produces unrealistic outputs, e.g.\ $\mu_{c_{0}}(\xtest) \ll -0.1$.
Such predictions would look obviously suspicious to a user, so we would say that our supposed malicious actor has not achieved their goal in this case.
Instead, we optimize the surrogate loss in \cref{eq:sup:l-mnist}. With this surrogate loss we are able to find kernel perturbations yielding benign-looking outputs and achieving the goal of the malicious actor to change the prediction at $\xtest$ to $c_1$, i.e.\ $\mu_{c_{1}}(\xtest) \approx 0.9$ and $\mu_{c}(\xtest) \approx -0.1$ for all $c \neq c_1$.
In this case, we feel that a user would not be able to identify these predictions as obviously wrong, and so we say that the malicious actor has achieved their goal of changing the predictions of $k_0$ without detection.

\textbf{Hyperparameter sensitivity.} To quantify variability in the Gram matrices arising from hyperparameter uncertainty, we vary $\sigma^2_w$ over $30$ uniformly spaced points between $1.4$ and $1.5$, and  $\sigma^2_b$ over $30$ uniformly spaced points between $0.23$ and $0.33$.
This defines a grid that is ten times smaller than the grid \citet{Lee2018_dnn_gp} optimize their hyperparameters over when searching for $\hat\theta$.
Thus we have no reason to pick $\hat\theta$ as the true optimum over any of our grid points; that is, our grid points provide a natural (conservative) notion of uncertainty in $\hat\theta$.
\cref{fig:app_minist_grid} shows that over the $900$ possible hyperparameter combinations the train and test accuracies remain high and exhibit low variability. 

The experiment took approximately 55 minutes to run for a single test image. We ran the computations for the 1000 test images in parallel on a compute cluster with Intel Xeon E5-2667 v2, 3.30GHz cores, requesting one core each time. 
\section{Additional Gram matrix comparisons}
\label{app:histograms}

To assess qualitative interchangeability, we compared the 2-Wasserstein distance $d$ between the Gram matrices $k_{0}(X,X)$ and $\kperturb(X,X)$ to the 2-Wasserstein distance between $k_{0}(X,X)$ and $k^{(r)}(X,X),$ where $k^{(r)}$ had the same functional form as $k_{0}$ but different hyperparameters.
In \cref{sec:QI}, we argue that the 2-Wasserstein distance is a good default choice, as it corresponds to coordinate-wise differences in standard deviations.
However, we emphasize that if a user has problem-specific knowledge that would make another distance $d$ more suitable, then our workflow can use this $d$ just as well.
For example, if a user thinks that for any points $x_1, x_2 \in \R^D$, deviation in the covariance $\abs{k_1(x_1, x_2) - k_0(x_1, x_2)}$ is meaningful in their problem, then they may want to consider $d$ as the infinity norm between Gram matrices. 
Here, we examine what what happens in all of our experiments when considering a number of matrix norms and statistical distances for $d$; in particular, we consider the Frobenius norm, nuclear norm, spectral norm, infinity norm, and symmetrized Kullback-Leibler distance.

\begin{figure*}
\includegraphics[width = 0.325\textwidth]{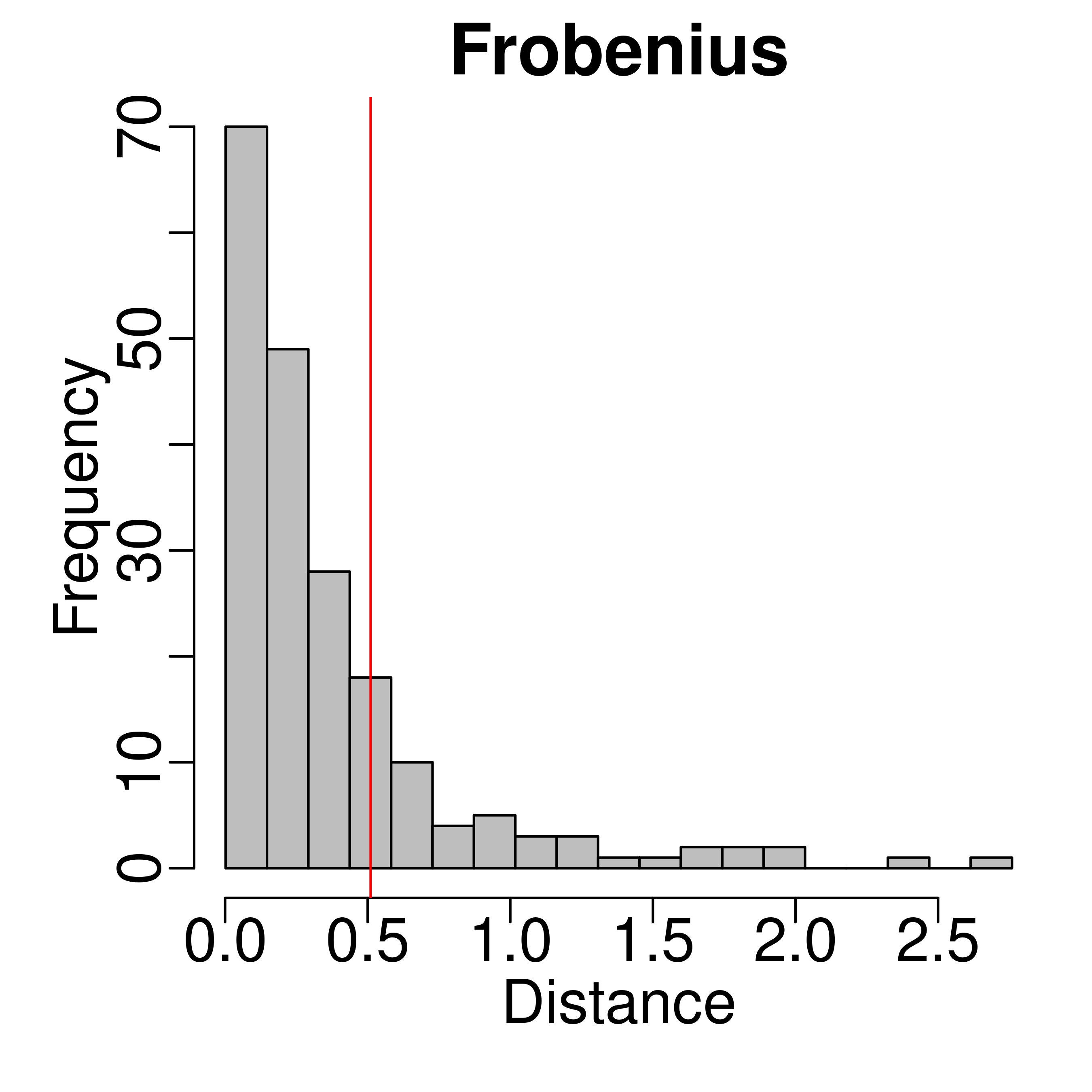}
\includegraphics[width = 0.325\textwidth]{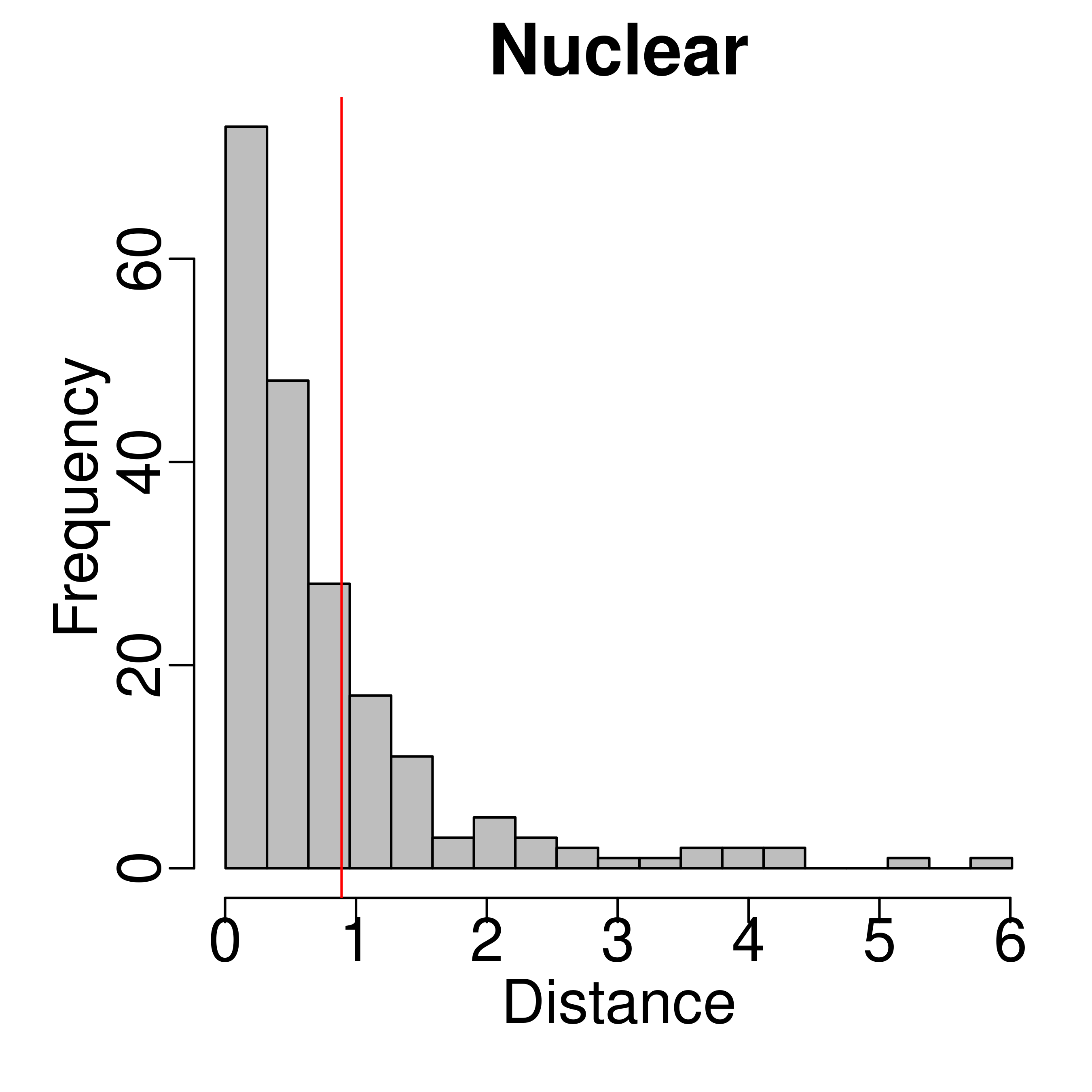}
\includegraphics[width = 0.32\textwidth]{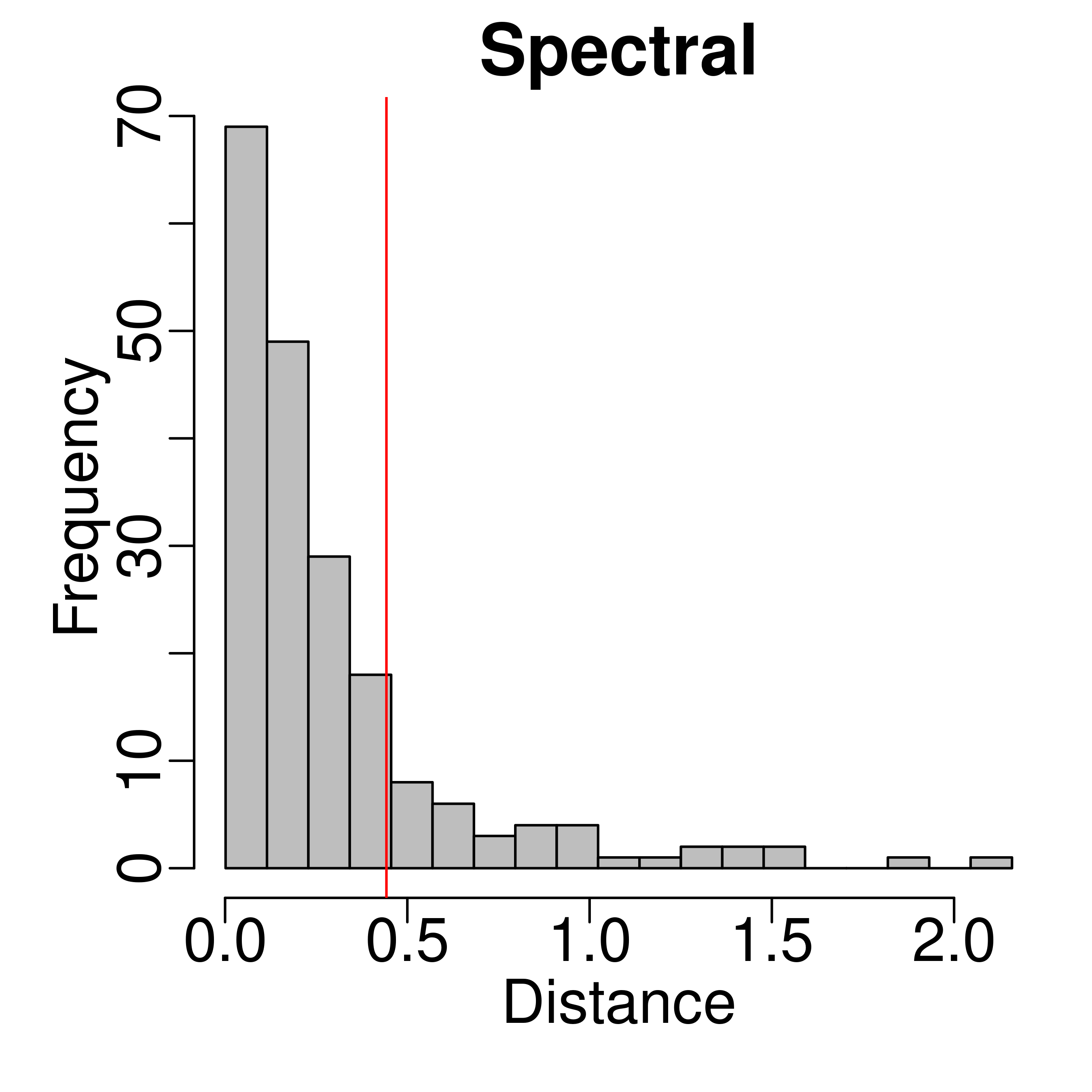}
\includegraphics[width = 0.325\textwidth]{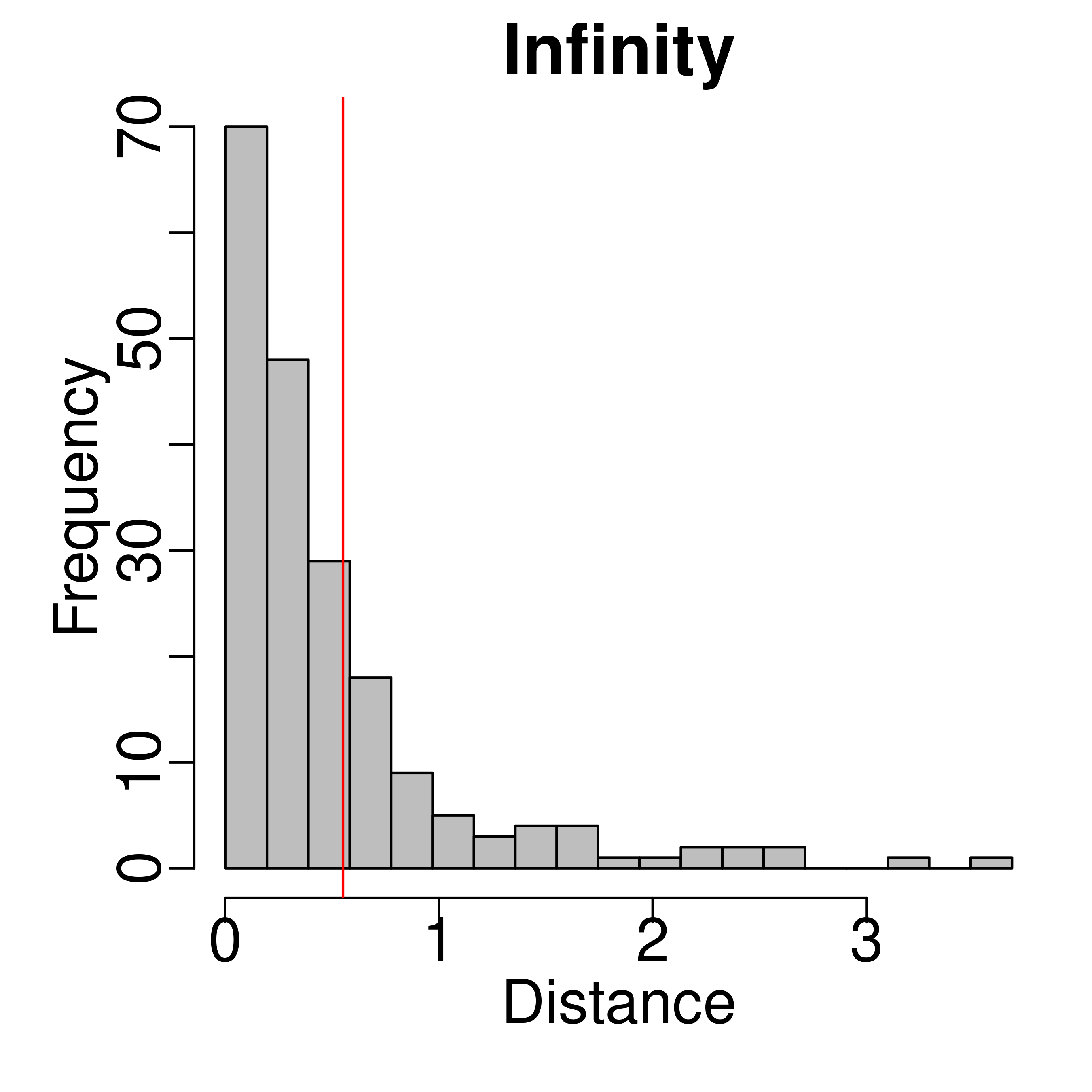}
\includegraphics[width = 0.325\textwidth]{figures/synthetic_extrapolation_wasserstein}
\includegraphics[width = 0.32\textwidth]{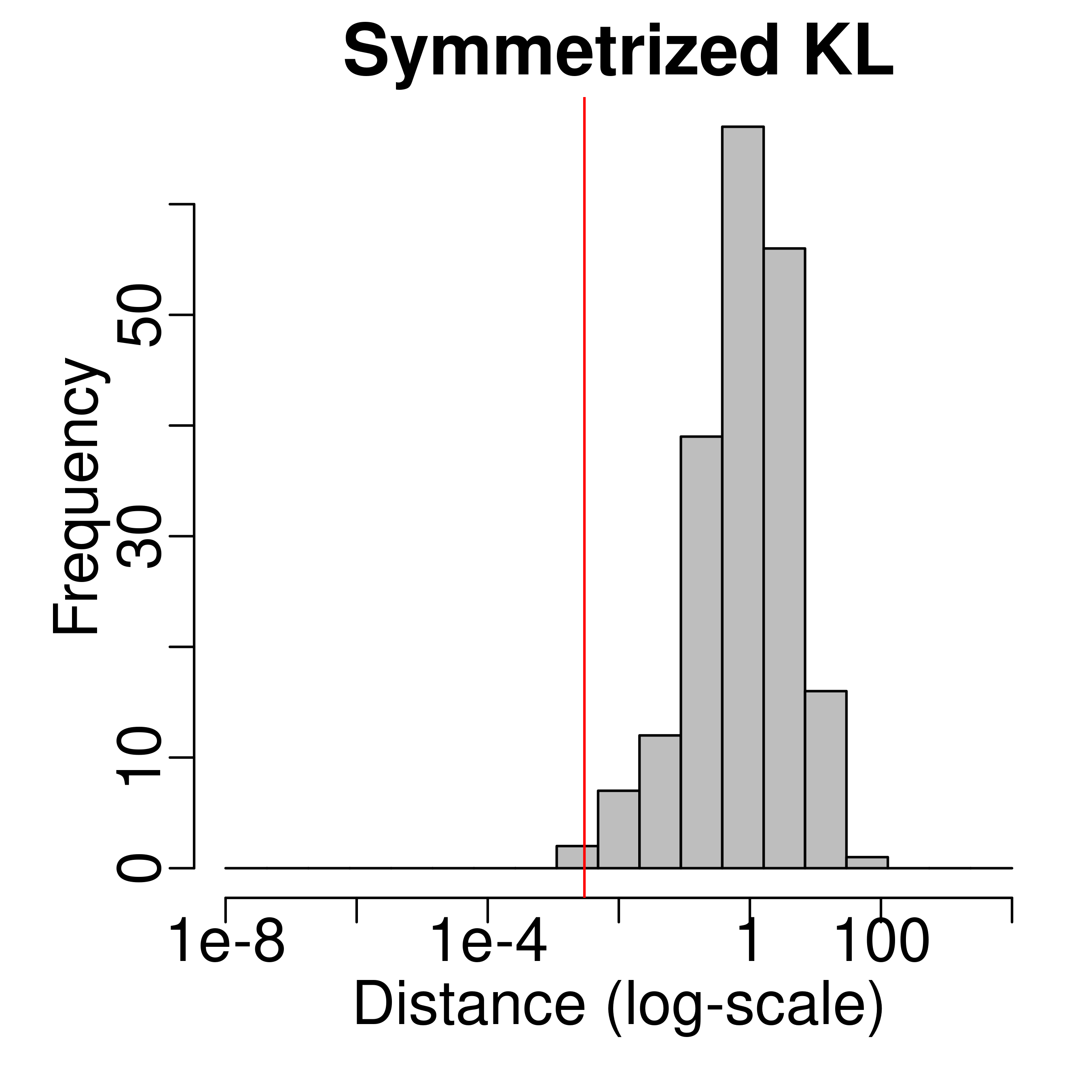}
\caption{Extra hyperparameter uncertainty histograms for our synthetic extrapolation example in \cref{sec:syntheticExample} in which we find find non-robustness. We compare the difference between $k_{0}$ and $\kperturb$ (red) to bootstrapped hyperparameter uncertainty (gray) in several distances.}
\label{fig:synthetic_extrapolation_histograms}
\end{figure*}

\begin{figure*}
\includegraphics[width = 0.325\textwidth]{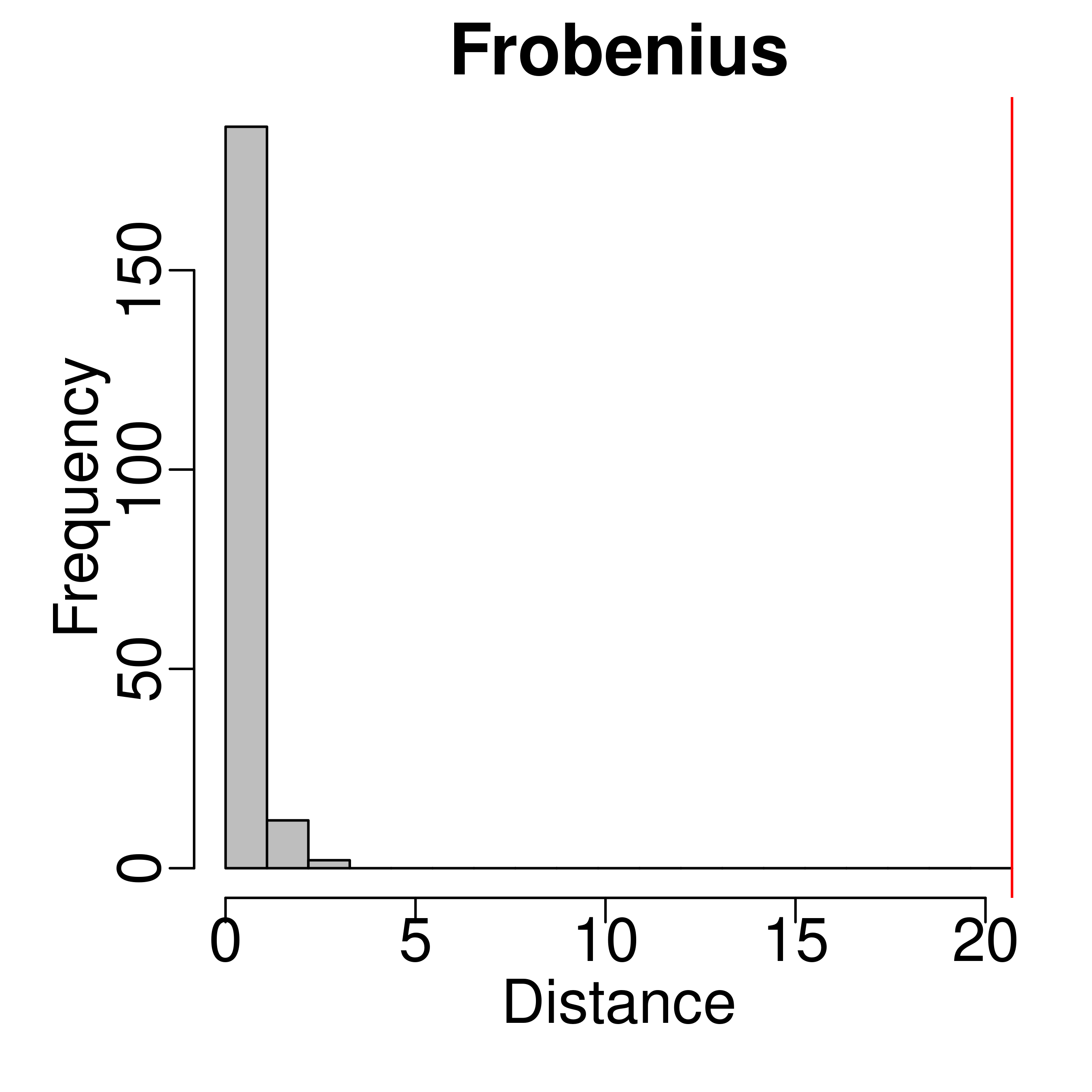}
\includegraphics[width = 0.325\textwidth]{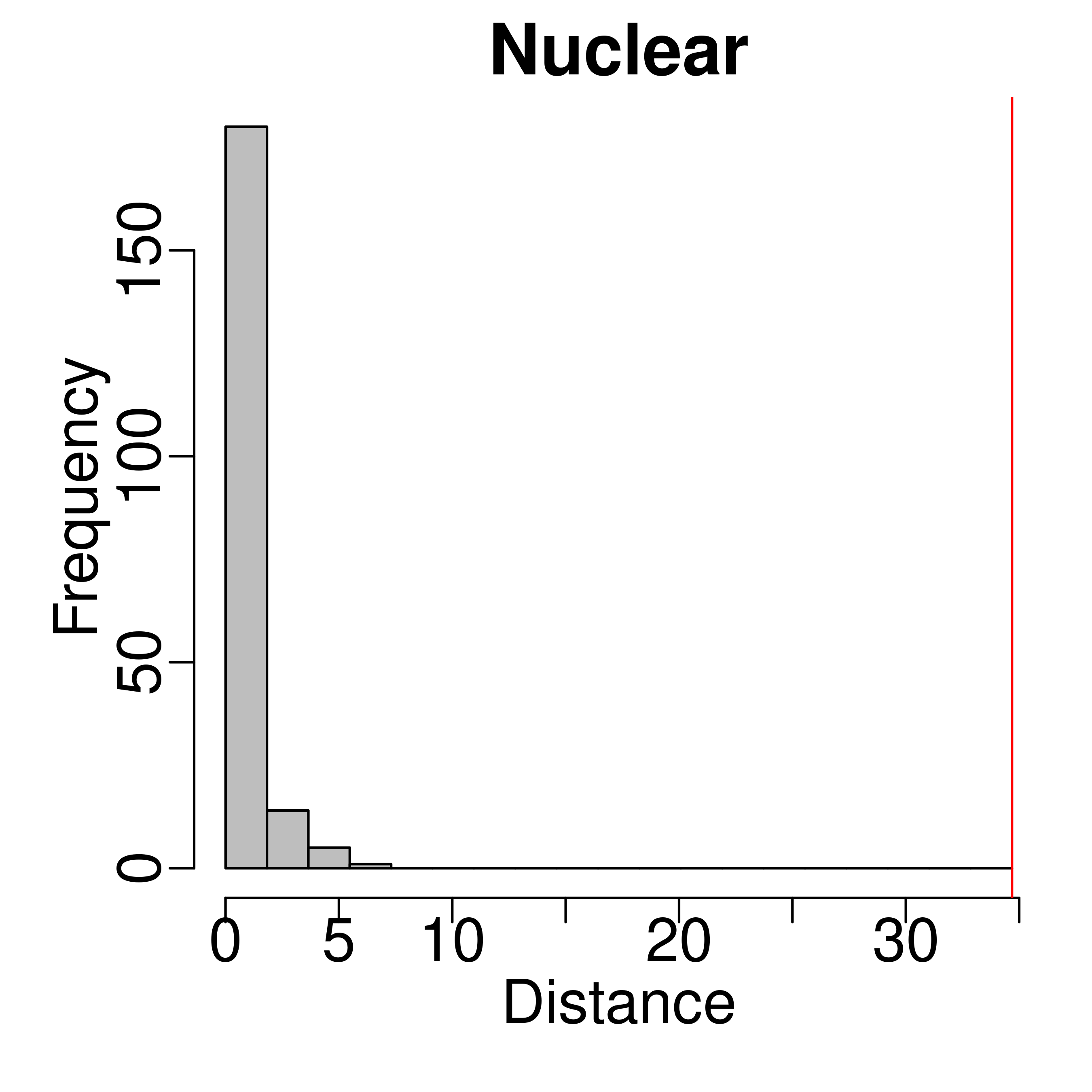}
\includegraphics[width = 0.32\textwidth]{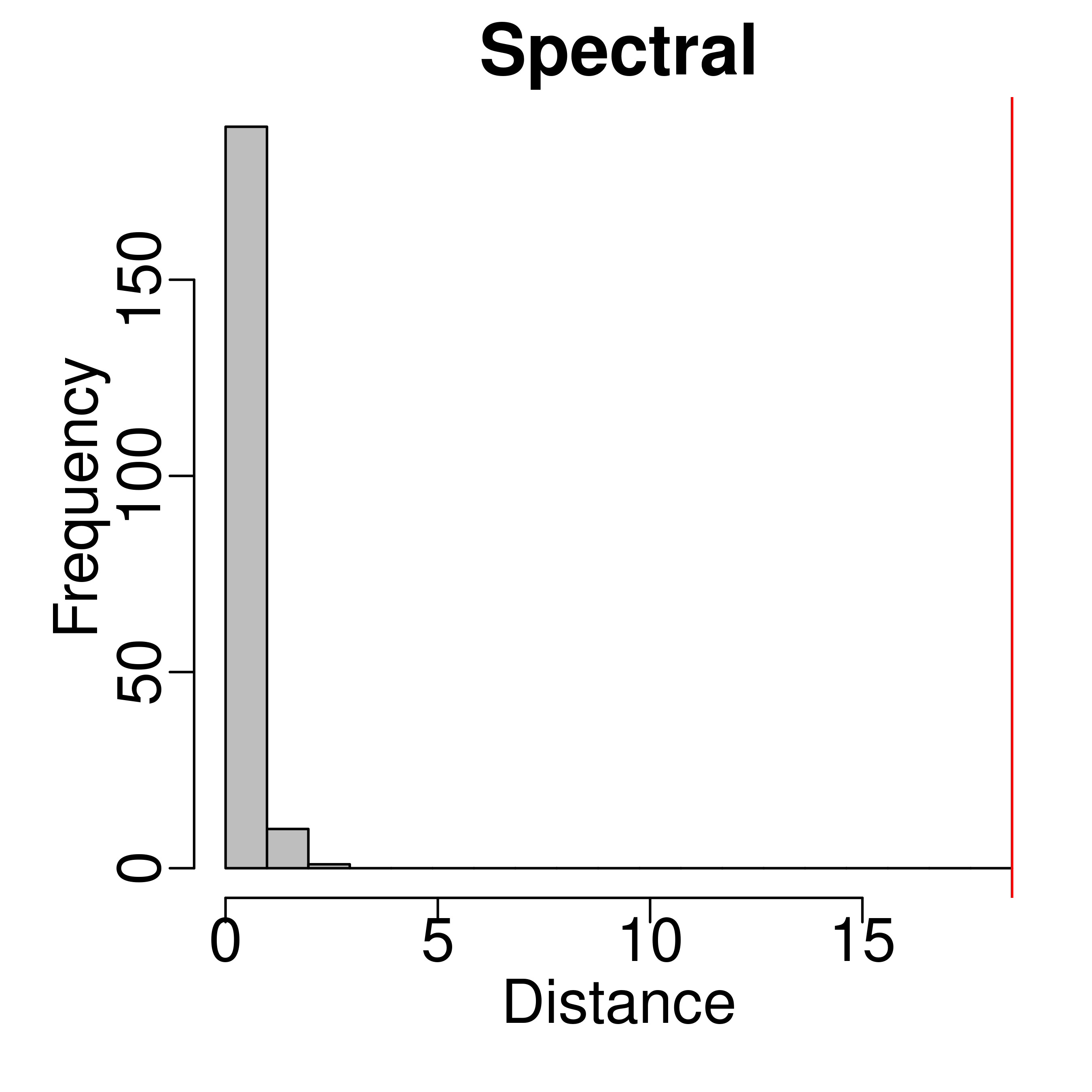}
\includegraphics[width = 0.325\textwidth]{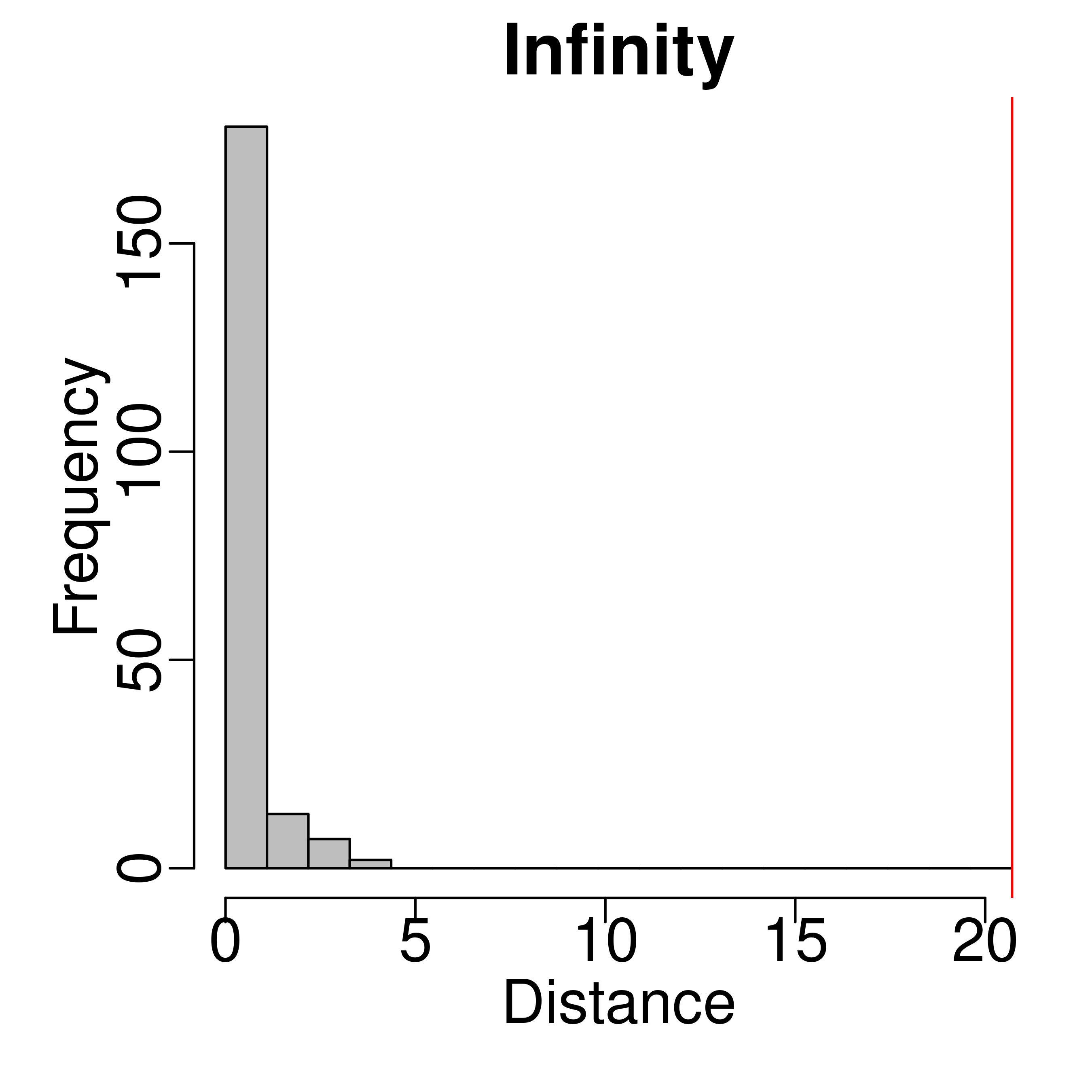}
\includegraphics[width = 0.325\textwidth]{figures/synthetic_interpolation_wasserstein}
\includegraphics[width = 0.32\textwidth]{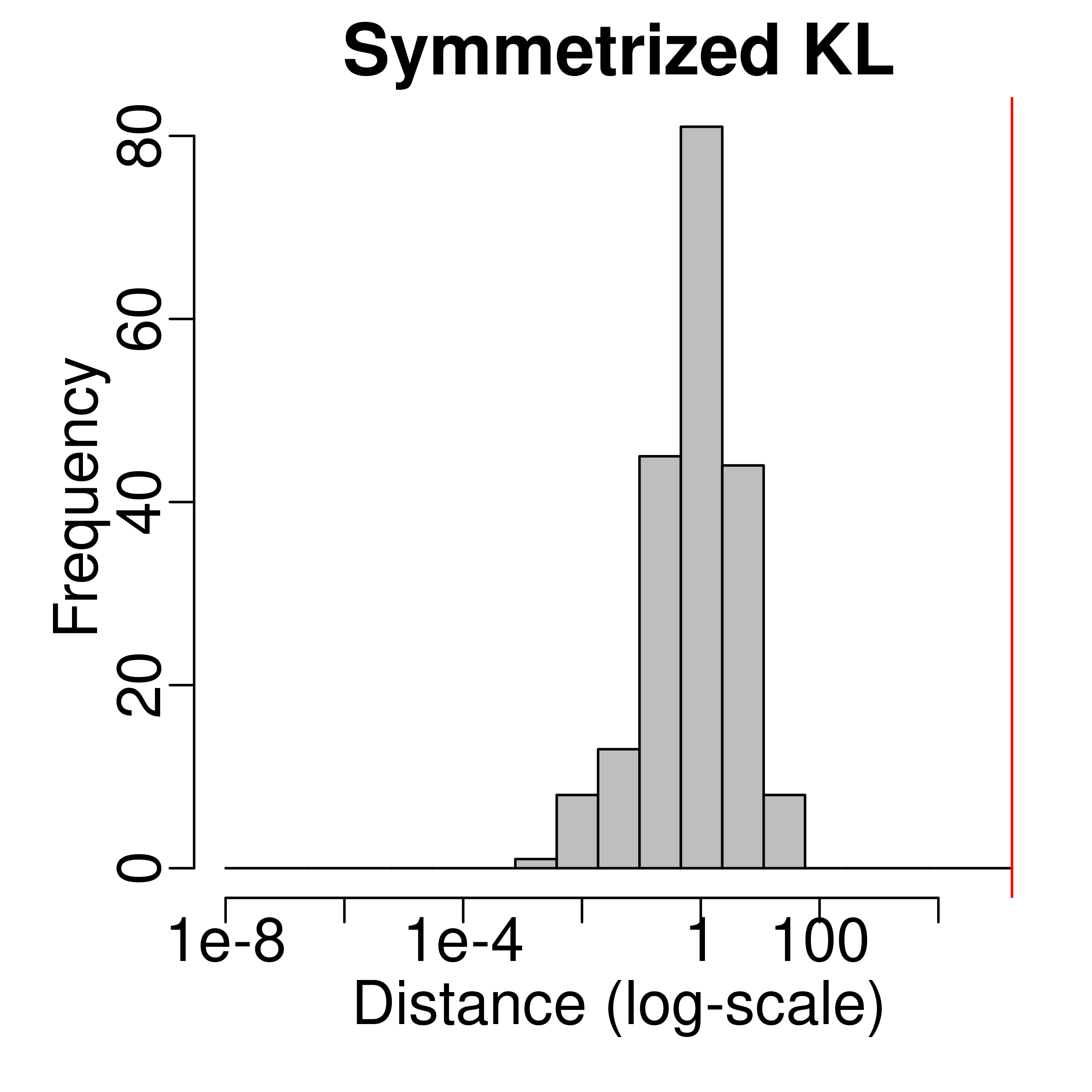}
\caption{Extra hyperparameter uncertainty histograms for our synthetic interpolation example in \cref{sec:syntheticExample} in which we find do not find non-robustness. We compare the difference between $k_{0}$ and $\kperturb$ (red) to bootstrapped hyperparameter uncertainty (gray) in several distances.}
\label{fig:synthetic_interpolation_histograms}
\end{figure*}

\begin{figure*}
\includegraphics[width = 0.325\textwidth]{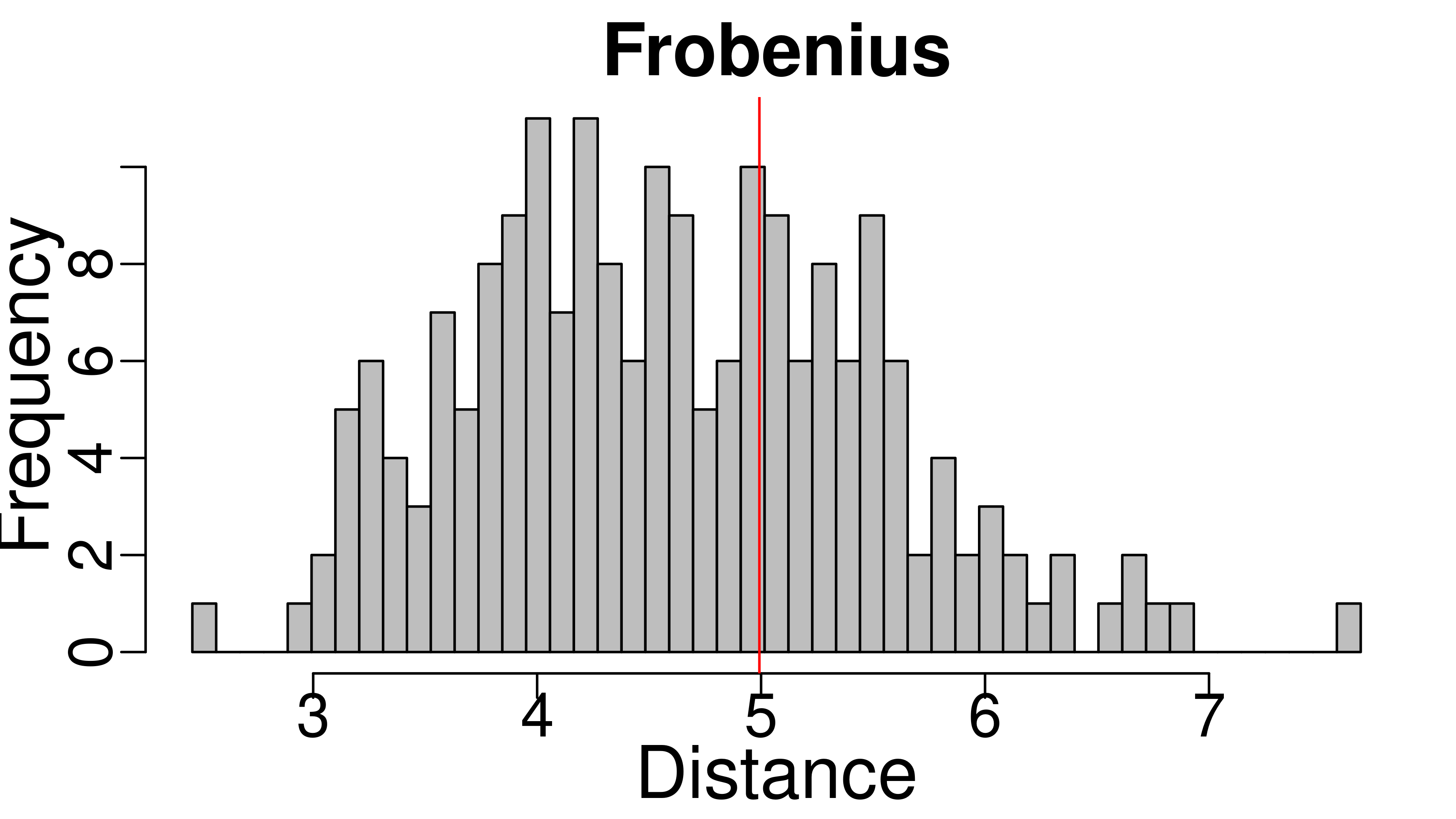}
\includegraphics[width = 0.325\textwidth]{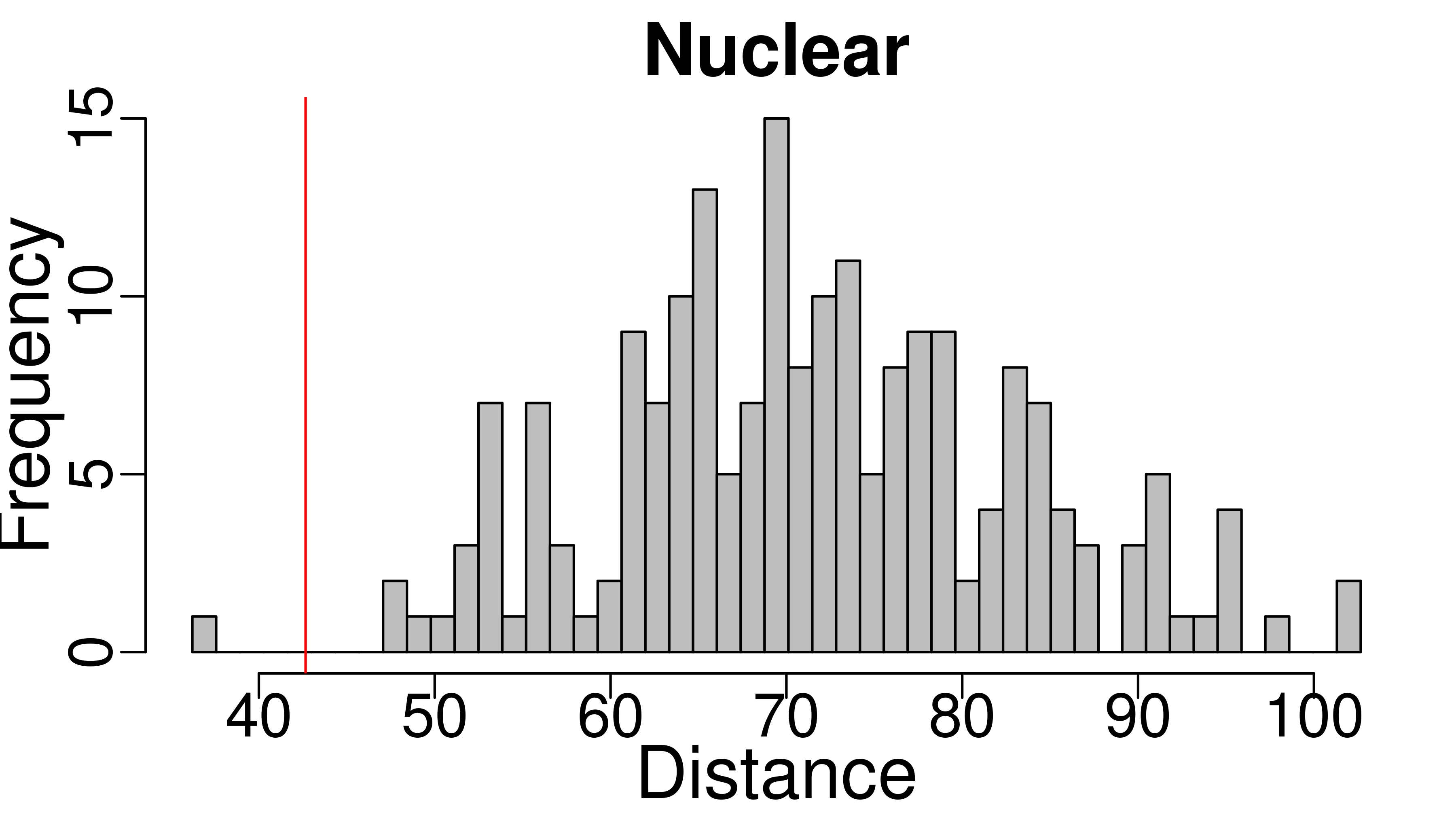}
\includegraphics[width = 0.325\textwidth]{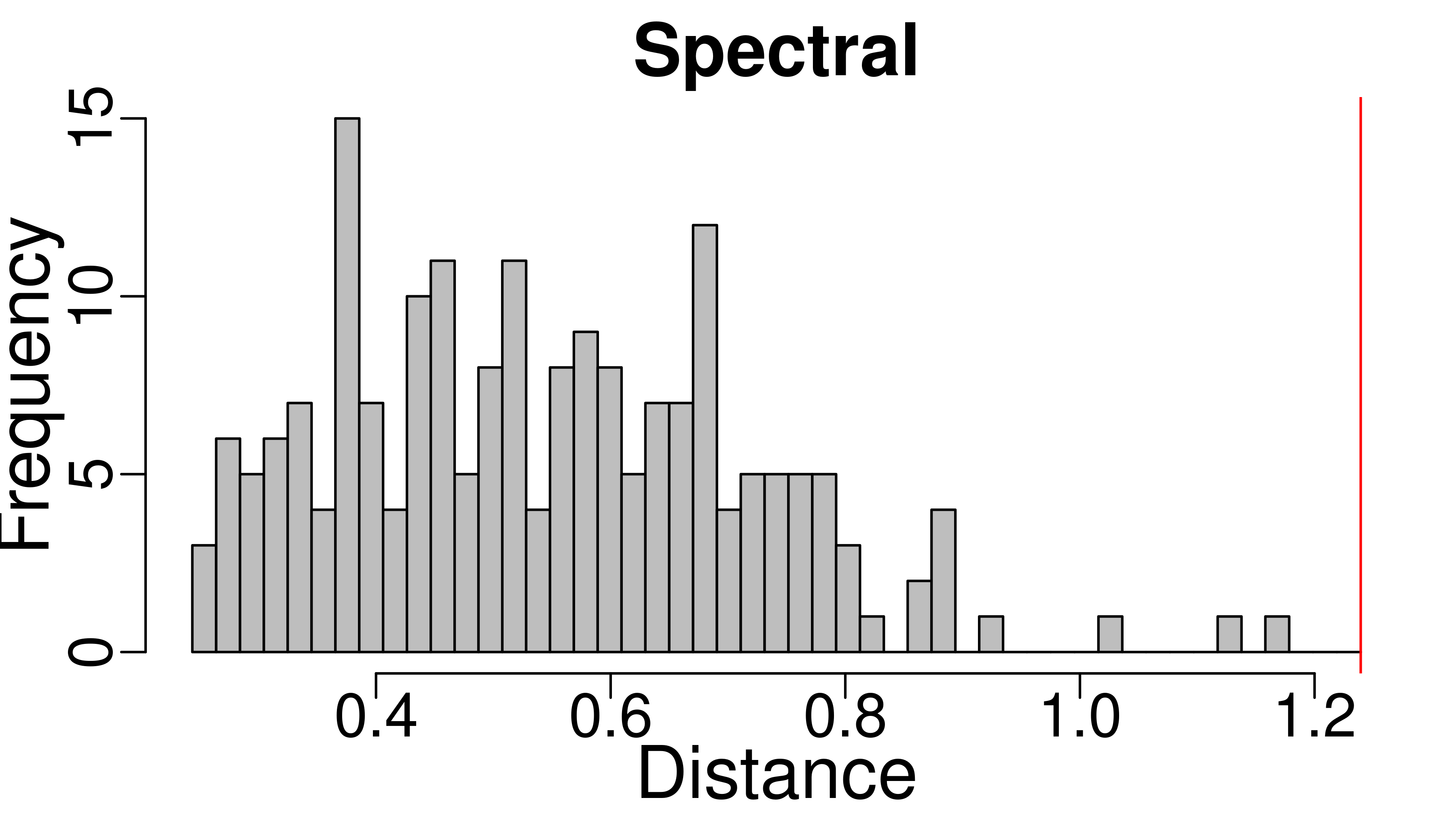}

\includegraphics[width = 0.325\textwidth]{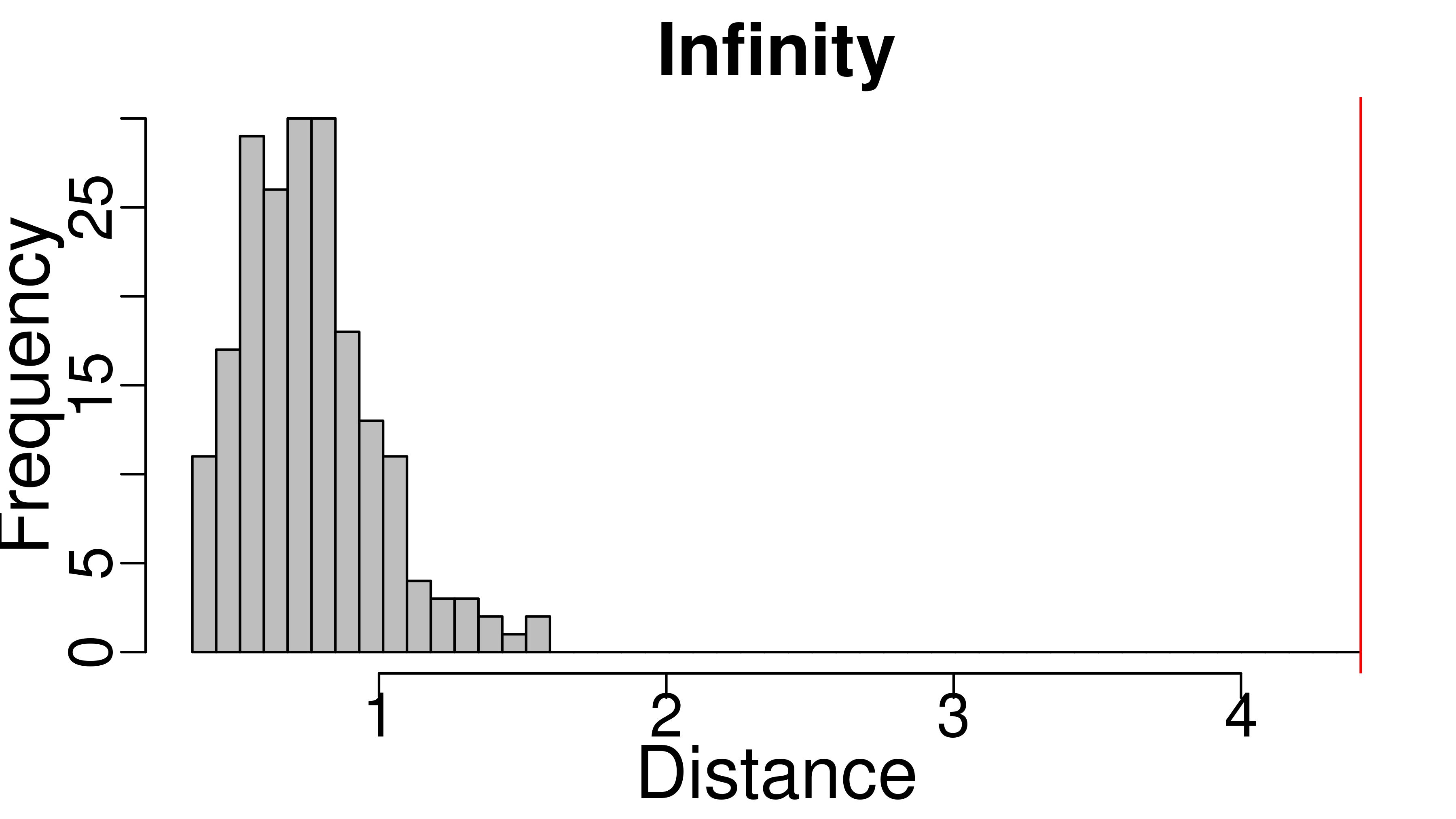}
\includegraphics[width = 0.325\textwidth]{figures/HR_nonrobust_wasserstein}
\includegraphics[width = 0.325\textwidth]{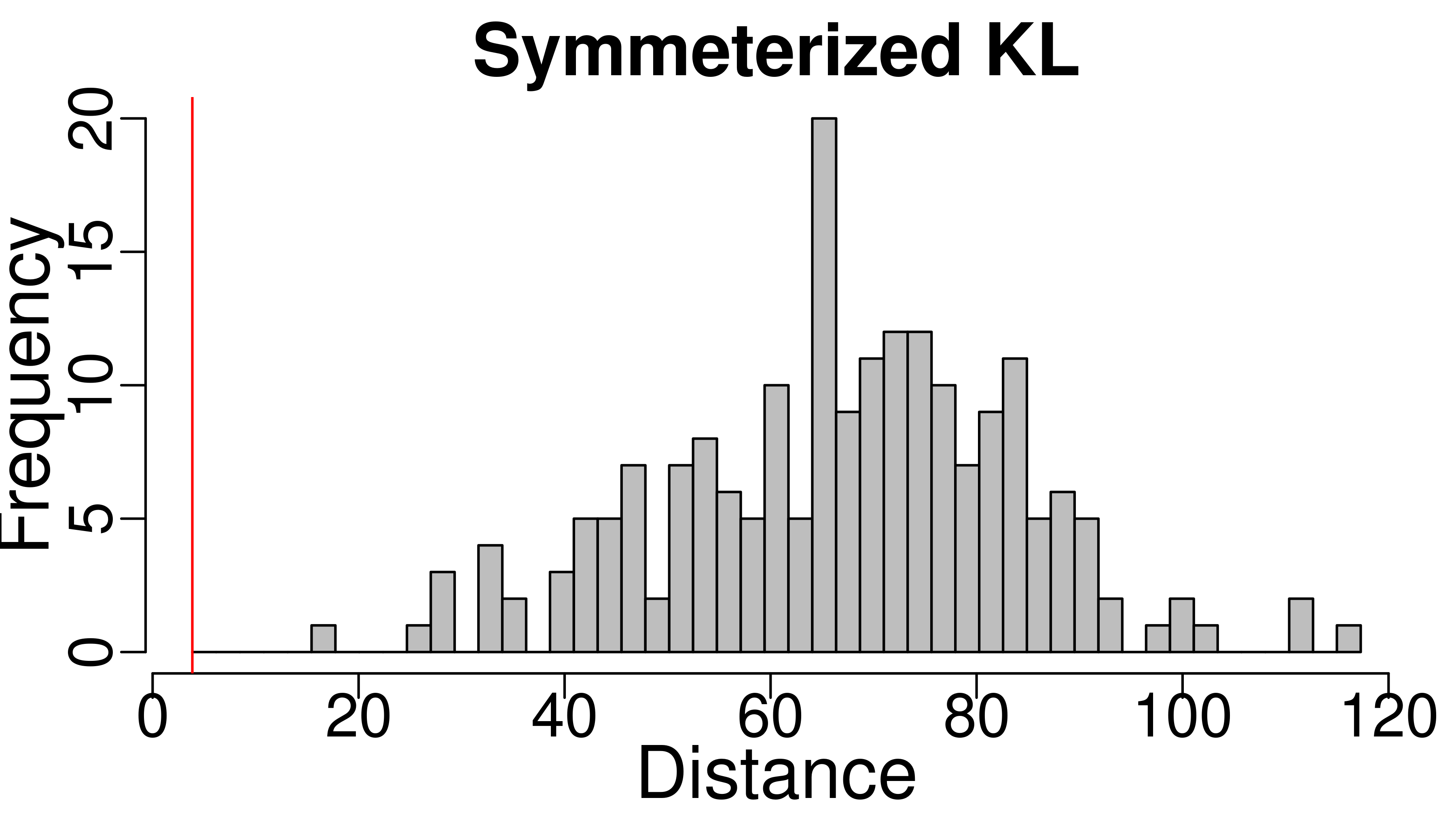}
\caption{Extra hyperparameter uncertainty histograms for our heart rate experiment in \cref{sec:heartRate} in which we find non-robustness. We compare the difference between $k_{0}$ and $\kperturb$ (red) to bootstrapped hyperparameter uncertainty (gray) in several distances.}
\label{fig:heartRate_histograms}
\end{figure*}

\begin{figure*}[t]
\includegraphics[width = 0.325\textwidth]{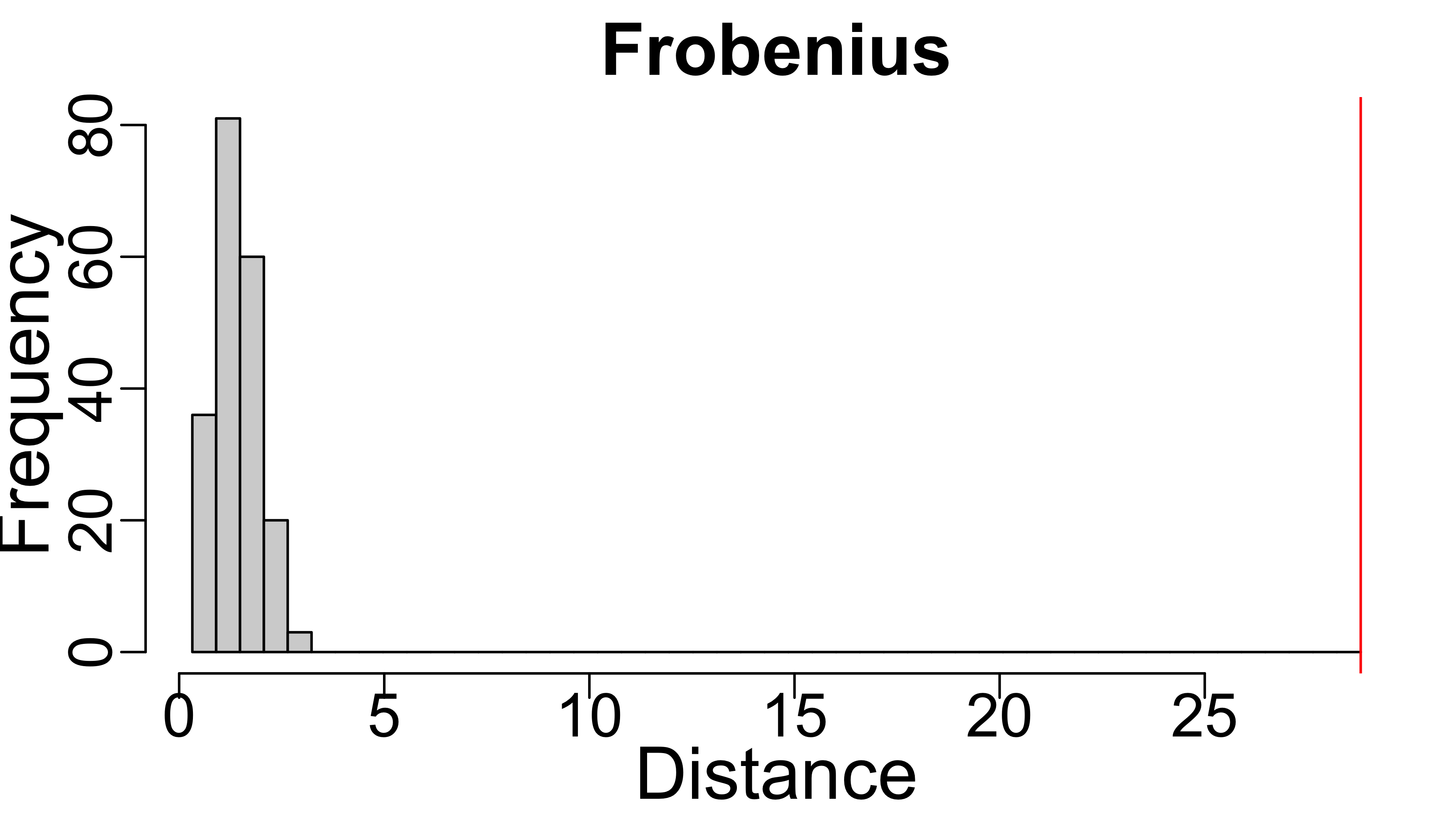}
\includegraphics[width = 0.325\textwidth]{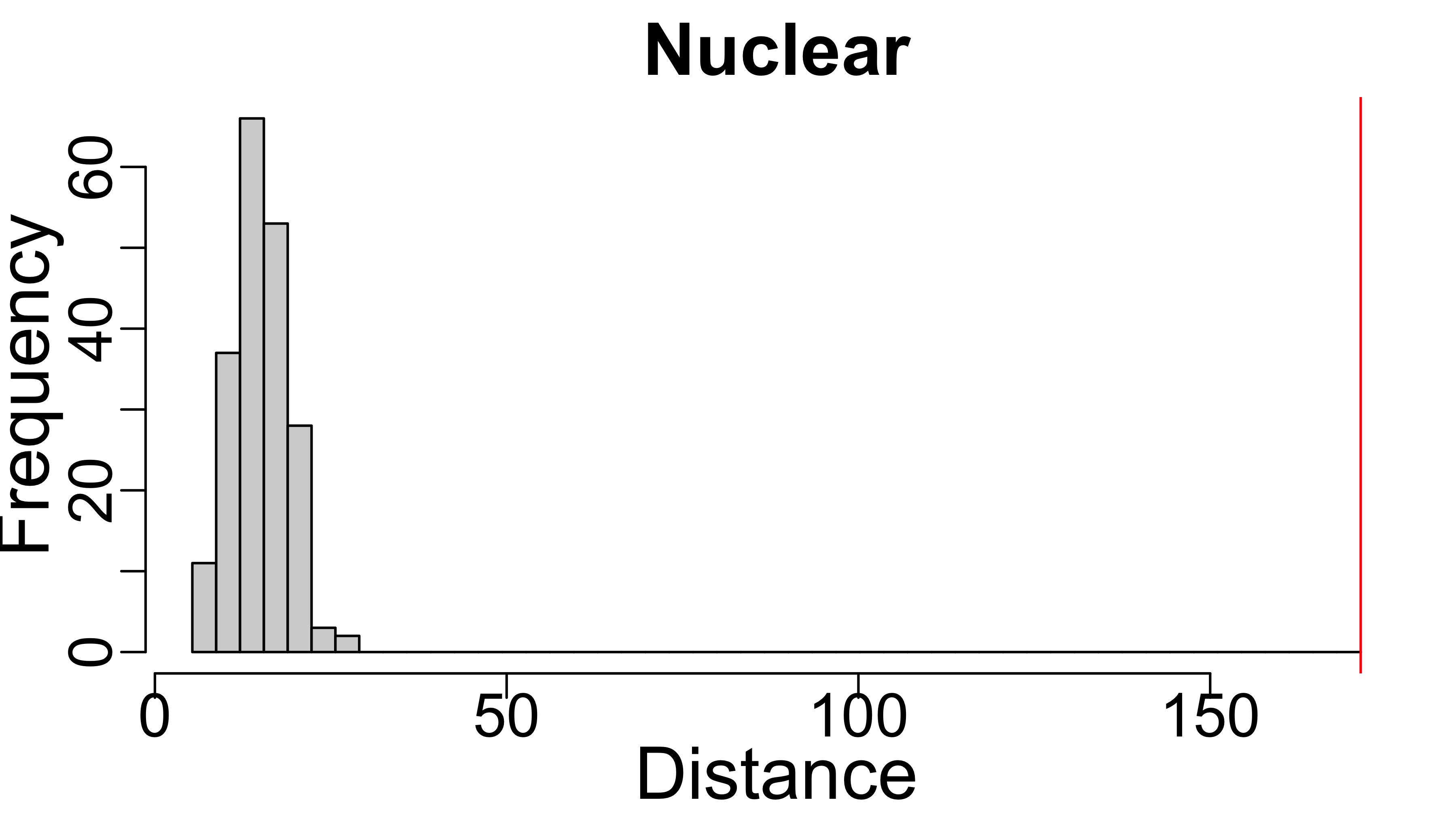}
\includegraphics[width = 0.325\textwidth]{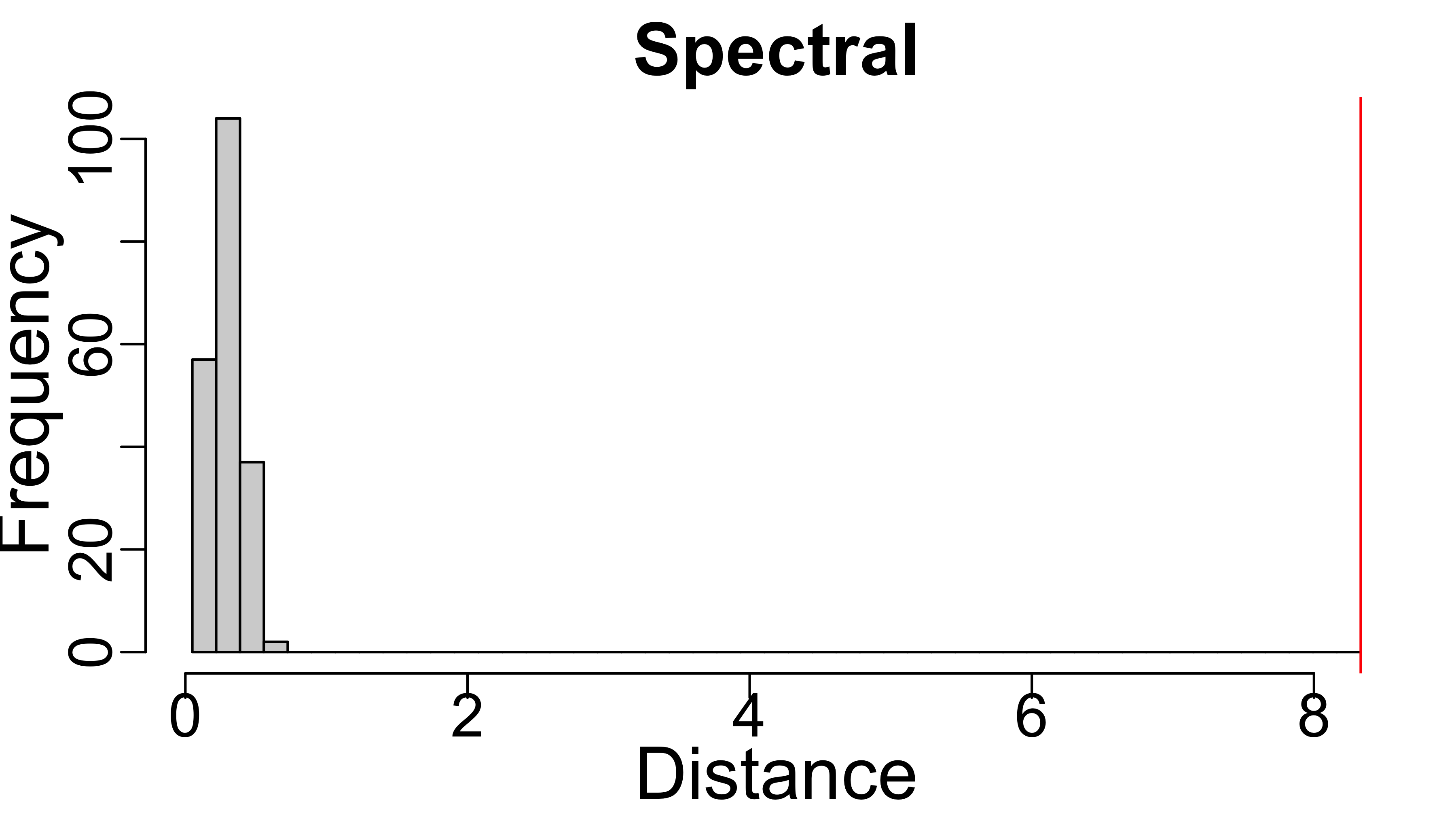}

\includegraphics[width = 0.325\textwidth]{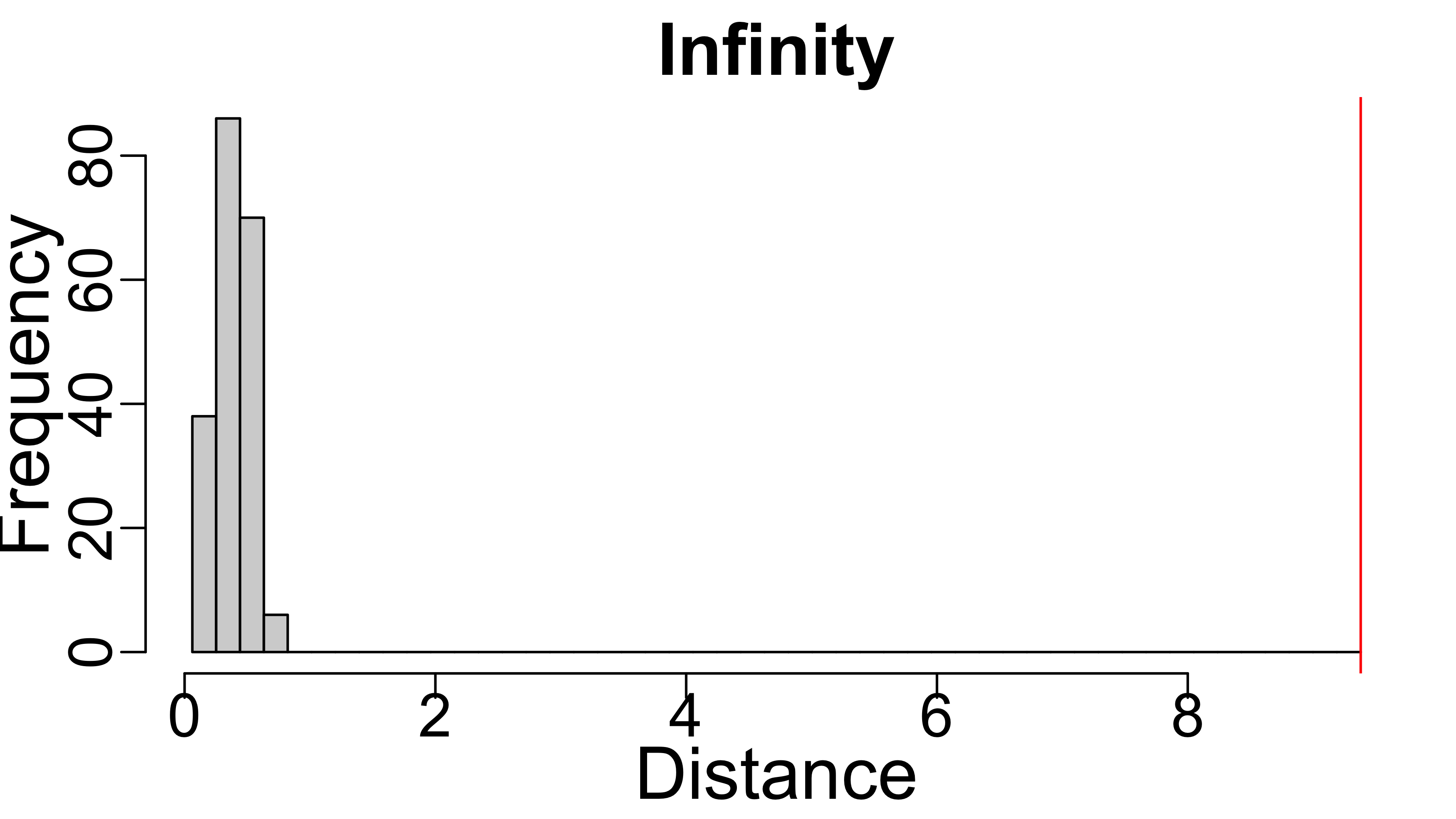}
\includegraphics[width = 0.325\textwidth]{figures/HR_non_nonrobust_wasserstein}
\includegraphics[width = 0.325\textwidth]{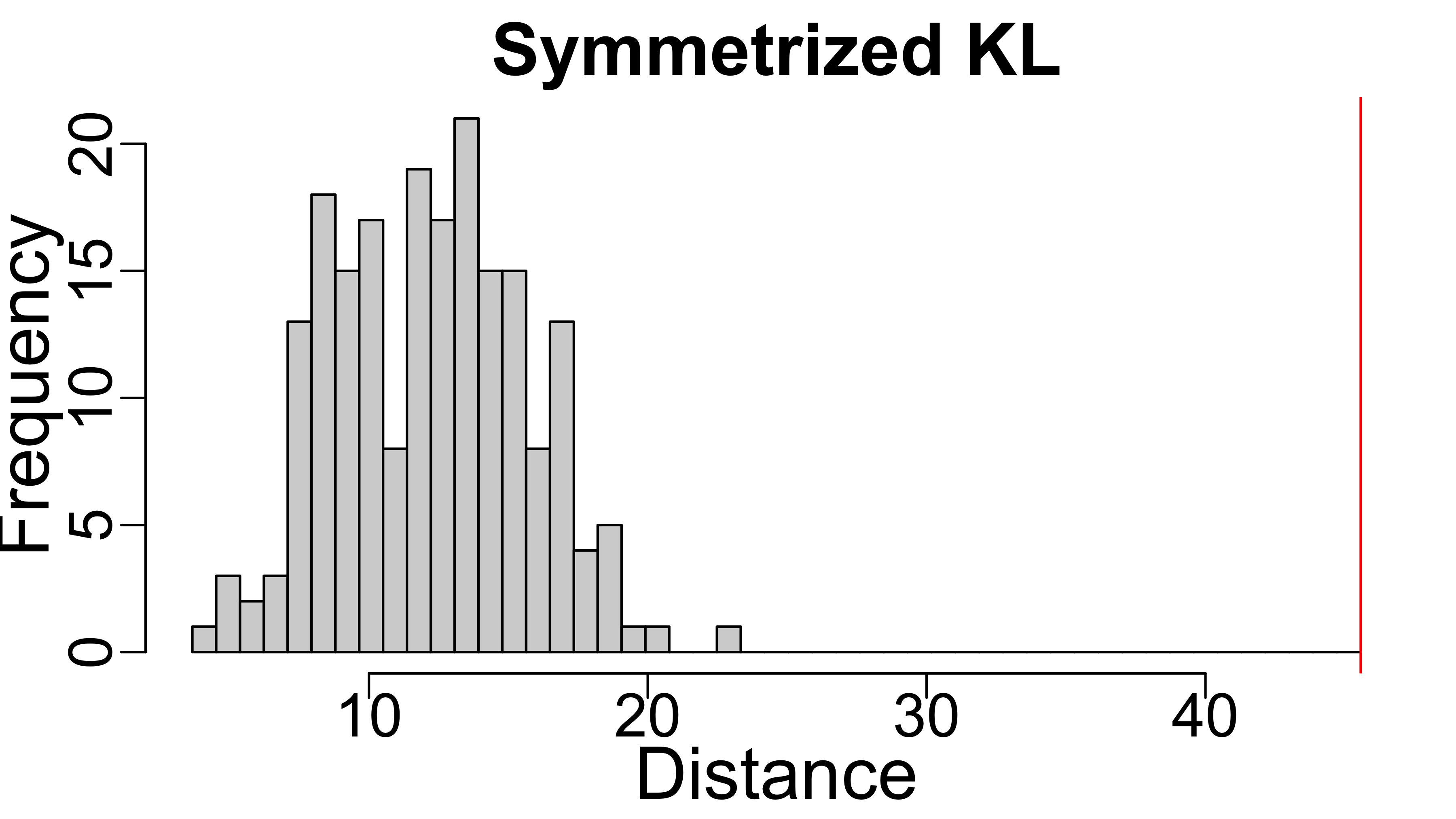}
\caption{Extra hyperparameter uncertainty histograms for our additional heart rate experiment in \cref{app:heartRate} in which we to find non-robustness. We compare the difference between $k_{0}$ and $\kperturb$ (red) to bootstrapped hyperparameter uncertainty (gray) in several distances.}
\label{fig:non_nonrobust_heartRate_histograms}
\end{figure*}

In \cref{fig:synthetic_extrapolation_histograms,fig:synthetic_interpolation_histograms,fig:heartRate_histograms,fig:non_nonrobust_heartRate_histograms},
we show the results of our histogram tests for qualitative interchangeability under these alternative $d$'s for our synthetic and heart-rate experiments.
While the use of some $d$'s leads to the same conclusion as our use of the 2-Wasserstein distance in the main text, the use of the spectral norm or infinity norm can result in a different conclusion. In particular, in our synthetic extrapolation experiment and our heart rate example from the main text, we concluded that $\kperturb$ and $k_0$ were qualitatively interchangeable (the red line sat to the left of the grey histograms); however, in \cref{fig:synthetic_extrapolation_histograms,fig:heartRate_histograms}, we see that the red line lies to the \emph{right} of the grey histograms, which would lead us to reject qualitative interchangeability.
We note that it is not surprising that kernels optimized according to \cref{constraintSet:stationary} deviation significantly in the spectral norm or infinity norm.
This is because \cref{constraintSet:stationary} only constraints the spectral density of $\kperturb$ to be close in a percentage-wise sense to the spectral density of $k_0$.
If the spectral density of $k_0$ is large in an absolute sense -- and it typically is for lower frequencies -- then $\kperturb$'s spectral density can have large deviations in an absolute sense.
Large absolute deviations in the spectral density allow for large absolute deviations in the Gram matrices.
As large absolute deviations in the Gram matrices is what the spectral and infinity norm measure, it is unsurprising that $\| \kperturb(X,X) - k_0(X,X) \|_{spectral}$ and $\| \kperturb(X,X) - k_0(X,X) \|_{infinity}$ are somewhat large.
So, if a user decides that the spectral or infinity norm are appropriate in their context, we recommend using a constraint set in \cref{constraintSet:stationary} that better reflects this choice.
E.g.\ one might constrain the density of $\kperturb$ to be close in both a percentage and absolute sense.

\begin{figure*}[t]
\includegraphics[width = 0.325\textwidth]{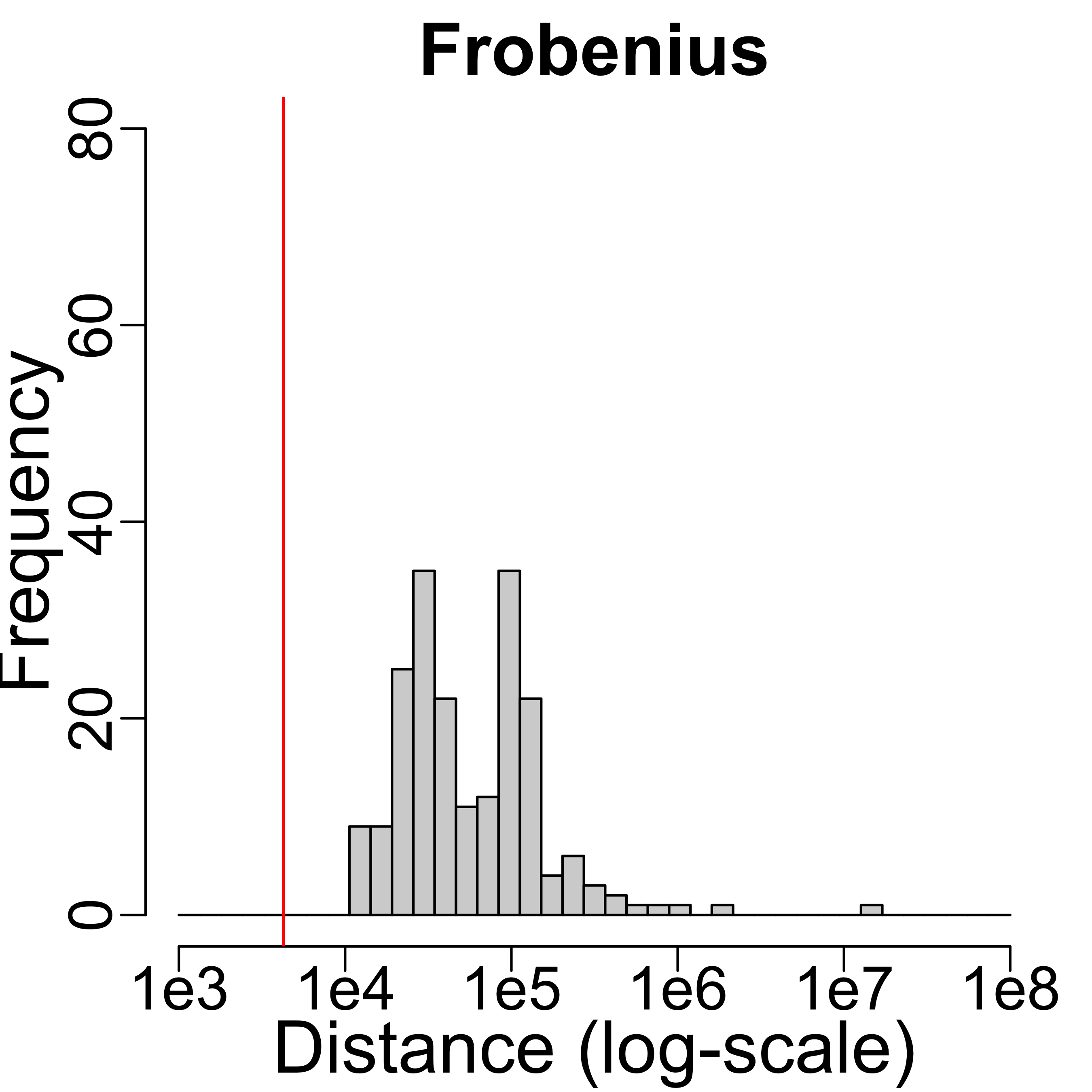}
\includegraphics[width = 0.325\textwidth]{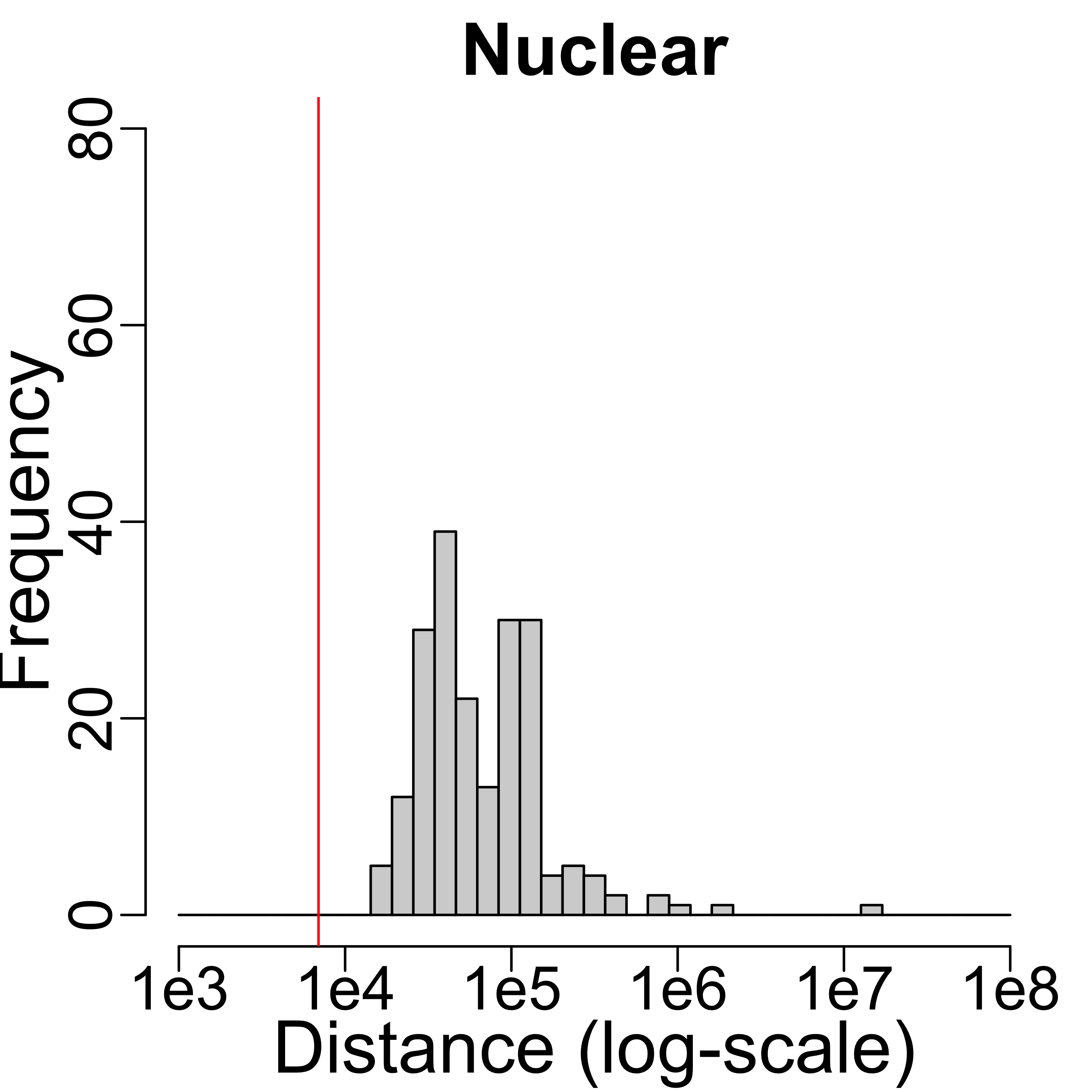}
\includegraphics[width = 0.325\textwidth]{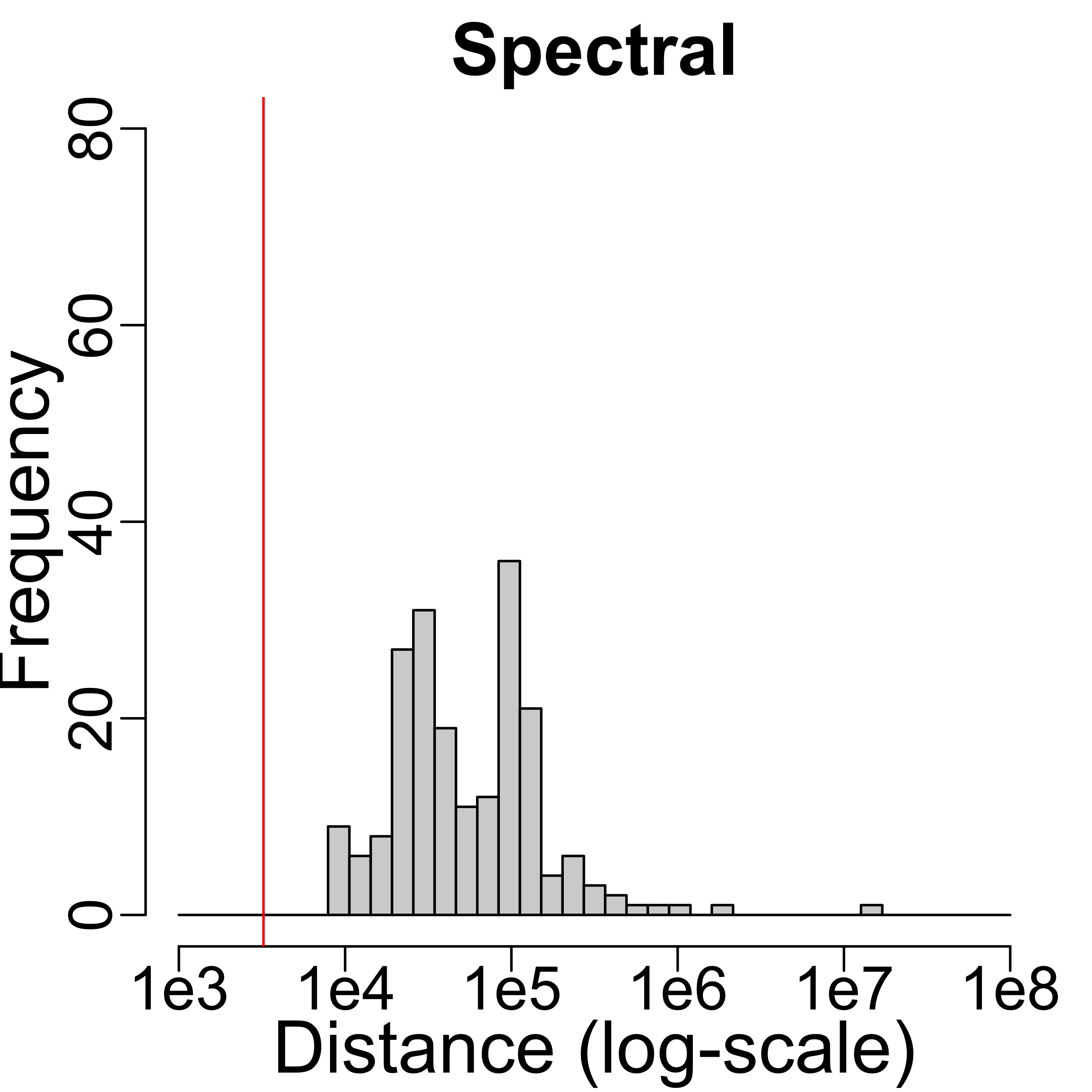}

\includegraphics[width = 0.325\textwidth]{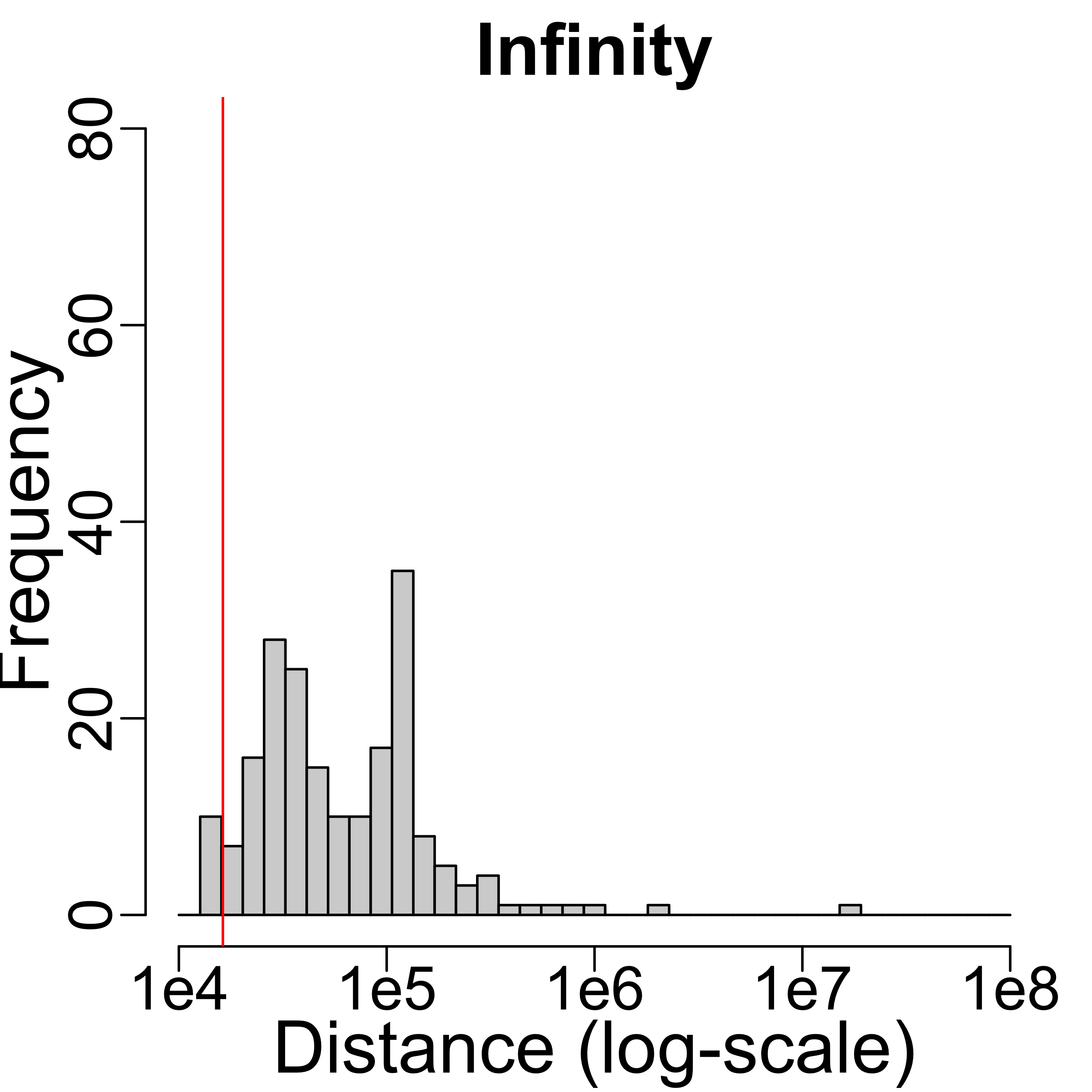}
\includegraphics[width = 0.325\textwidth]{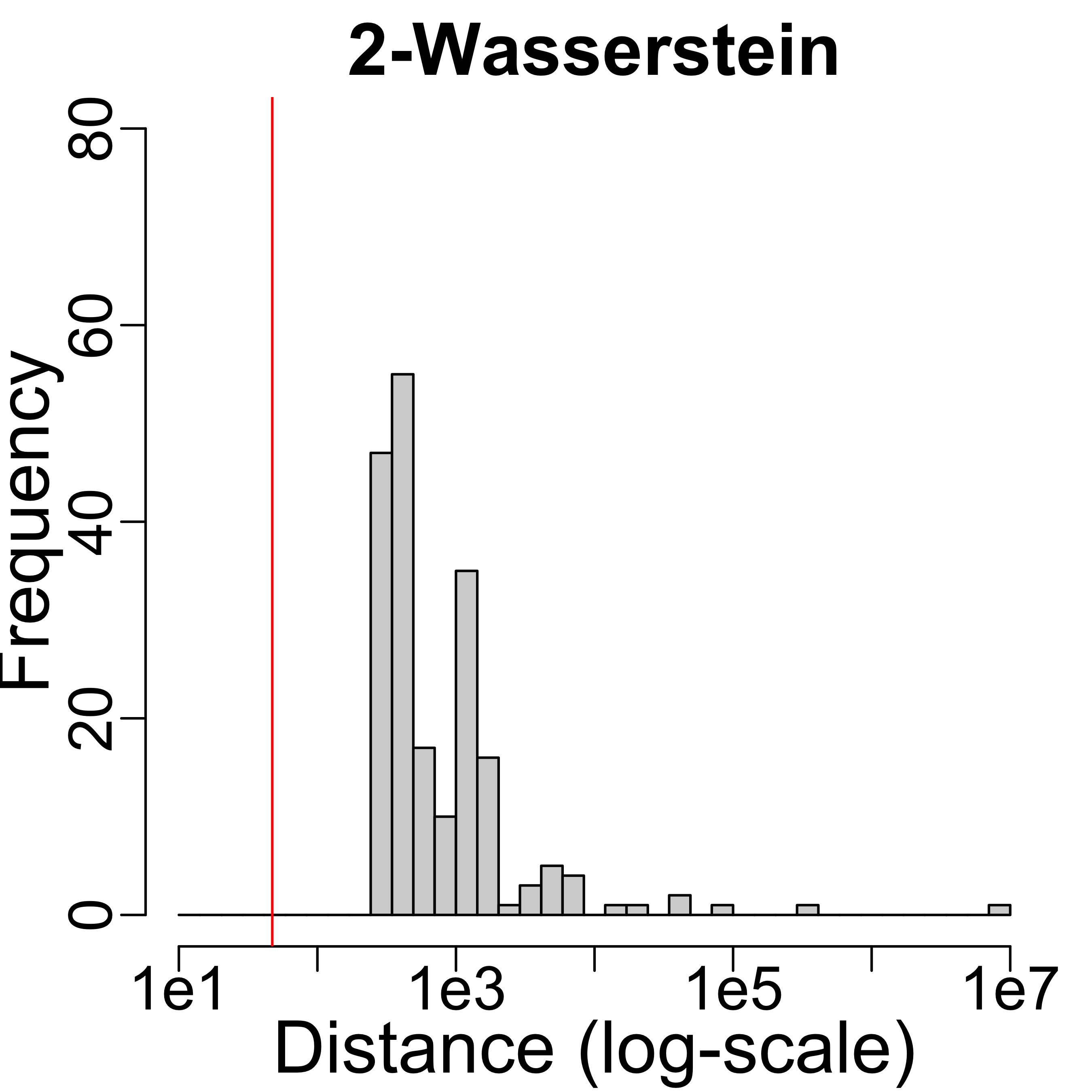}
\includegraphics[width = 0.325\textwidth]{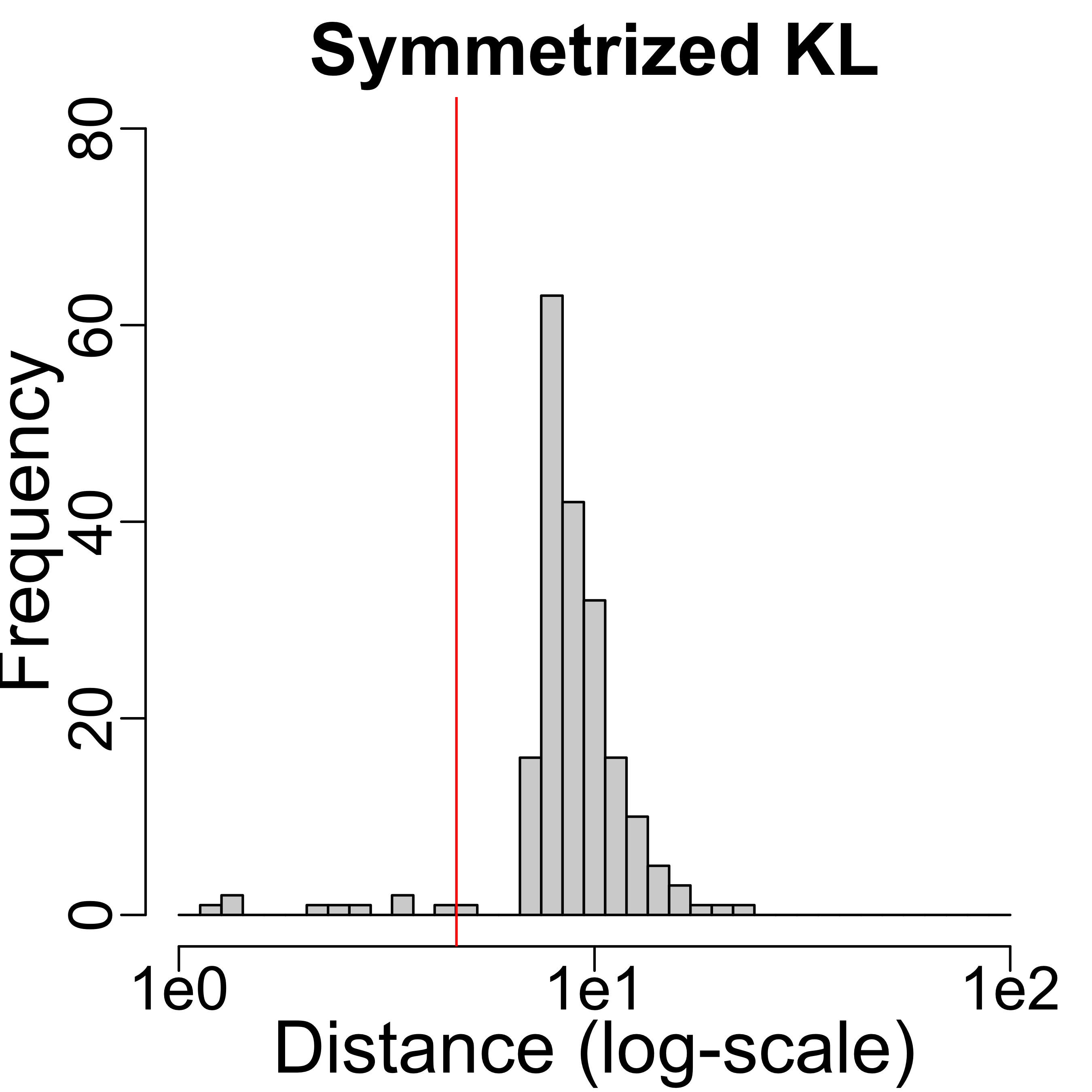}
\caption{Extra hyperparameter uncertainty histograms for our Mauna Loa experiment in \cref{sec:co2} in which we find non-robustness. We compare the difference between $k_{0}$ and $\kperturb$ (red) to bootstrapped hyperparameter uncertainty (gray) in several distances.}
\label{fig:mauna_loa_histograms}
\end{figure*}

\begin{figure*}[t]
\includegraphics[width = 0.325\textwidth]{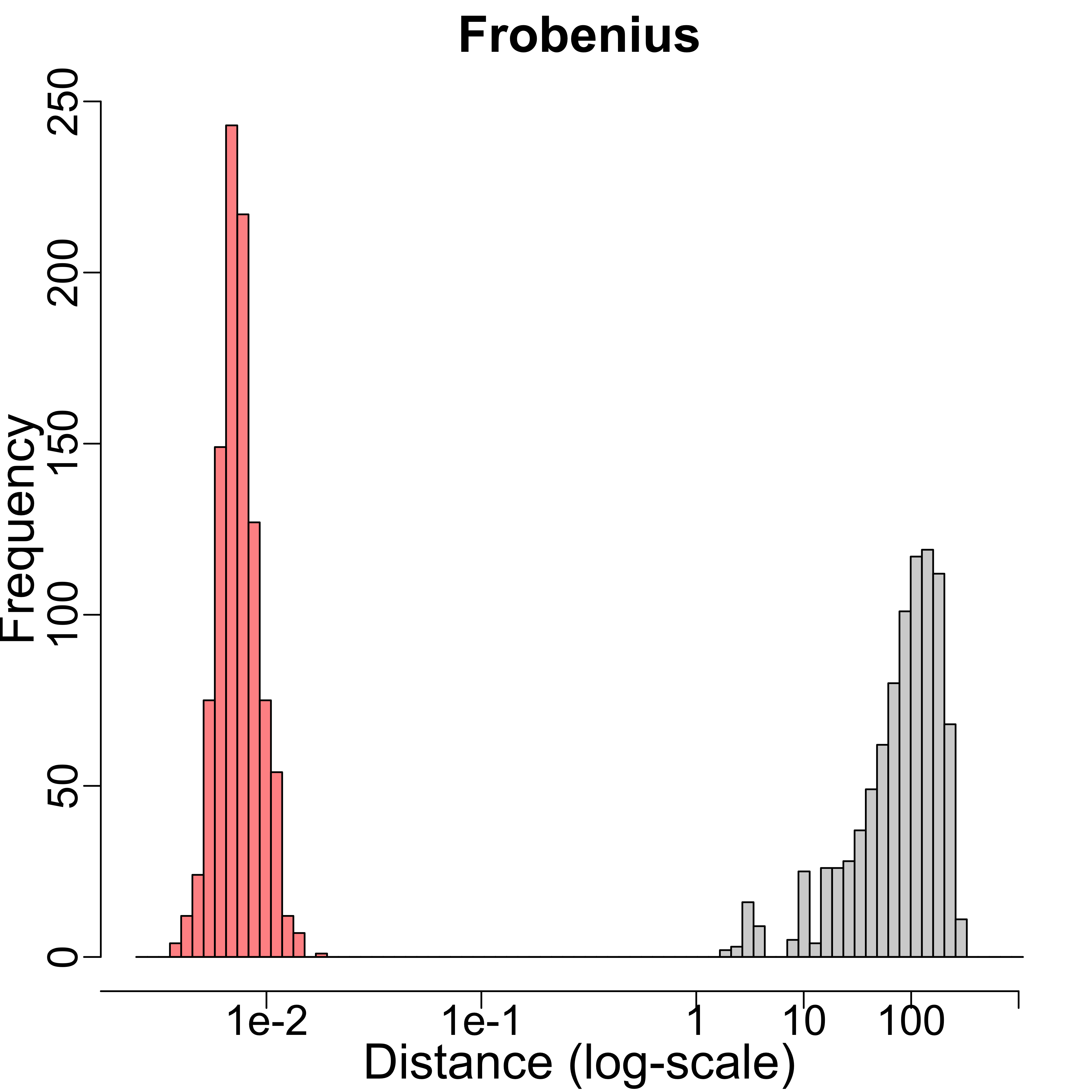}
\includegraphics[width = 0.325\textwidth]{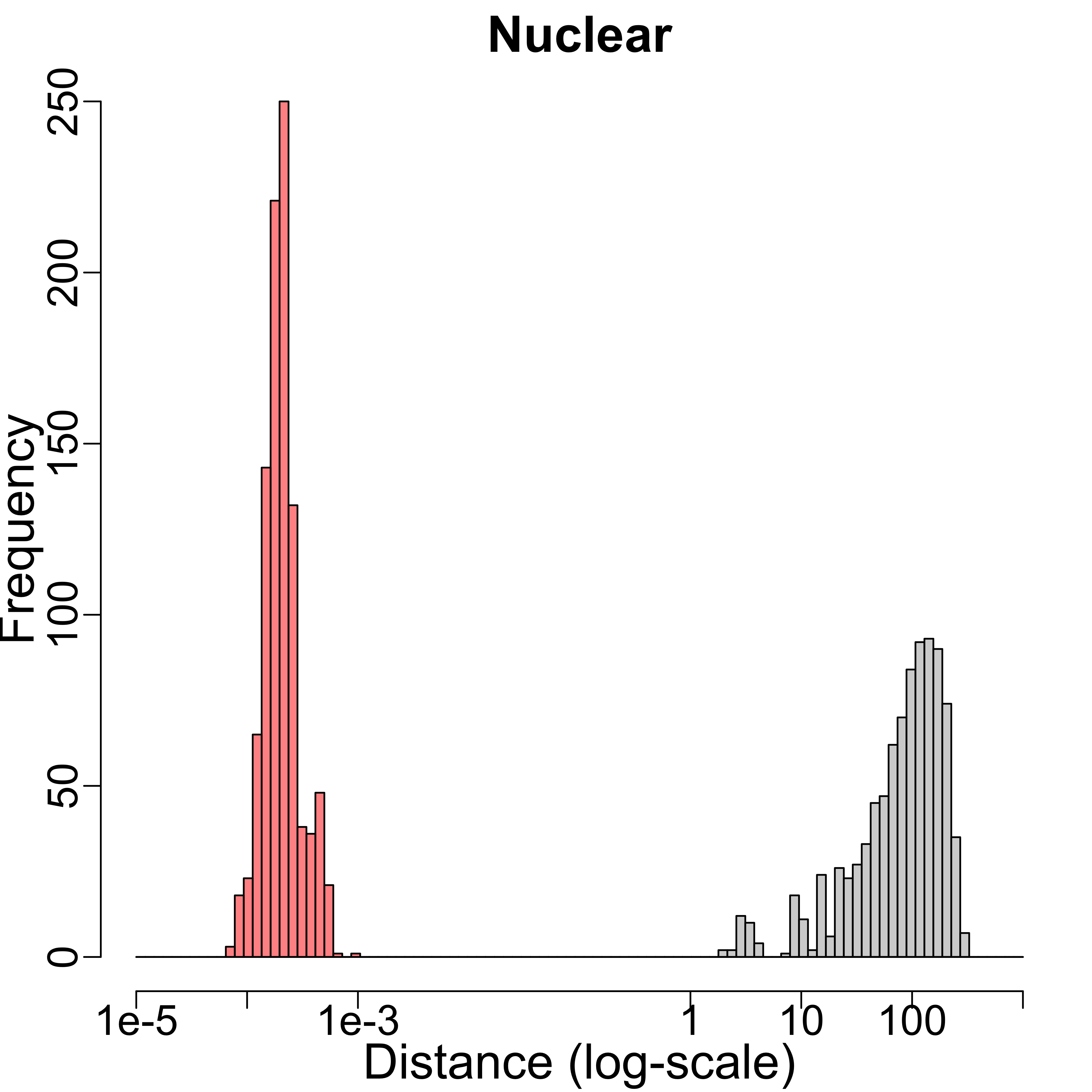}
\includegraphics[width = 0.325\textwidth]{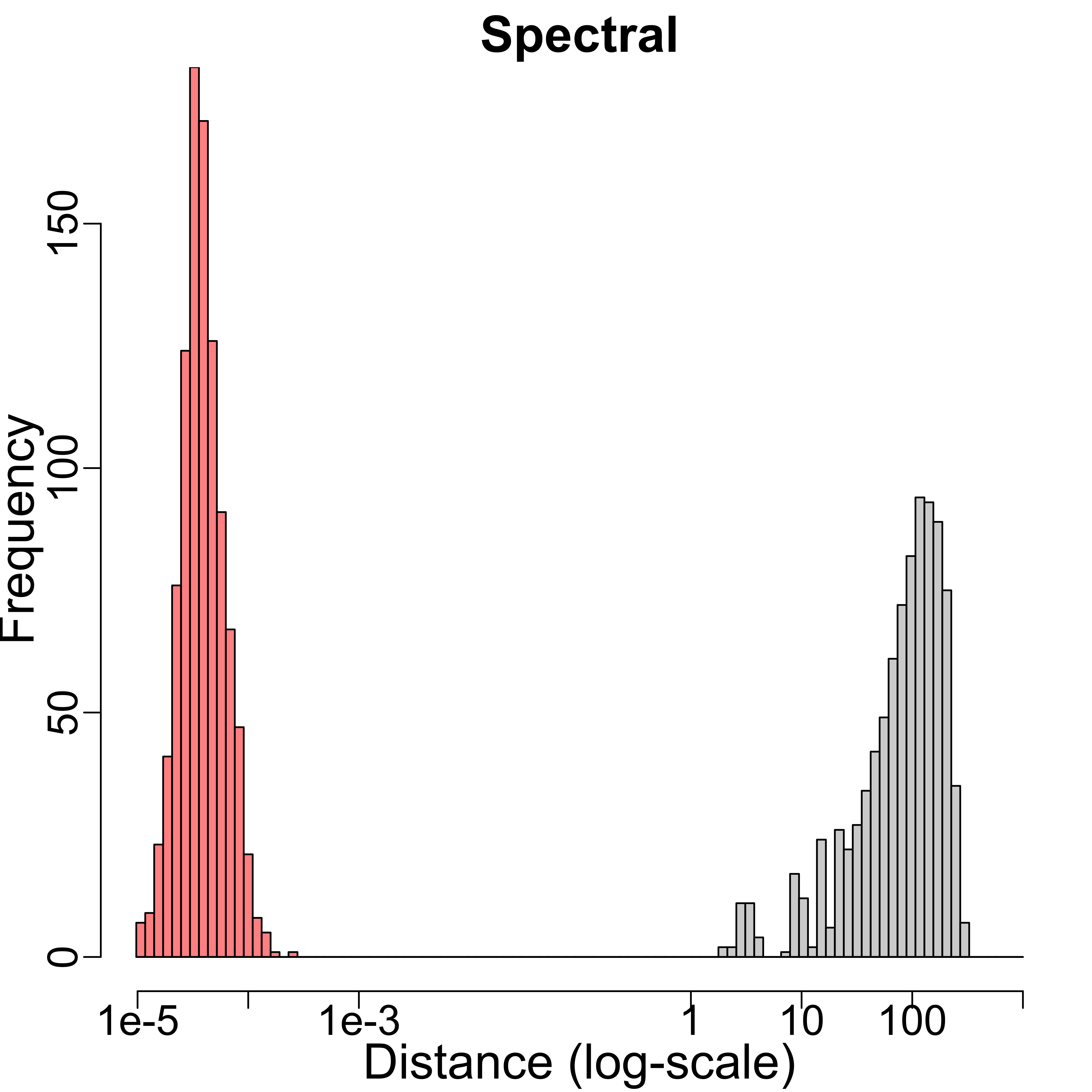}

\includegraphics[width = 0.325\textwidth]{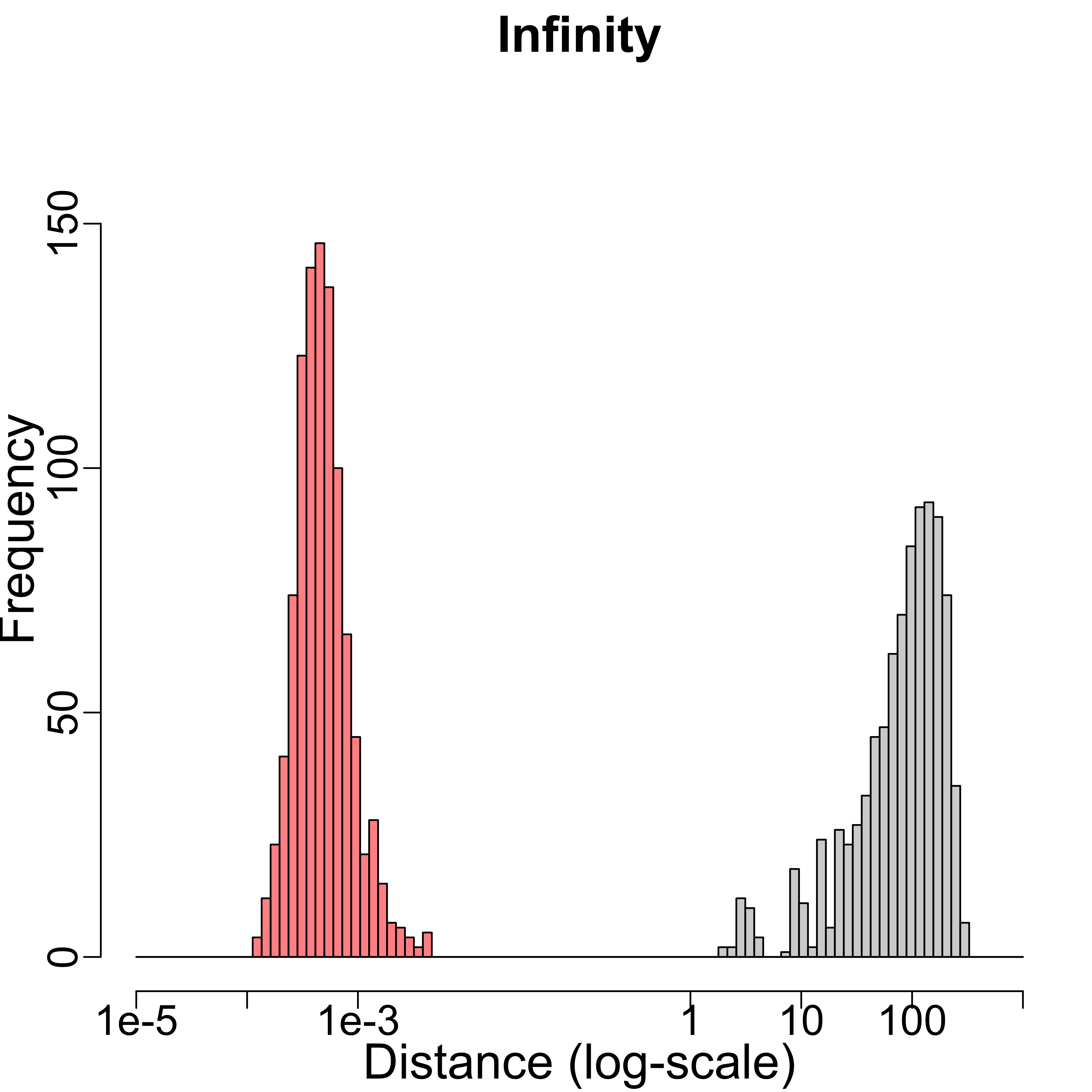}
\includegraphics[width = 0.325\textwidth]{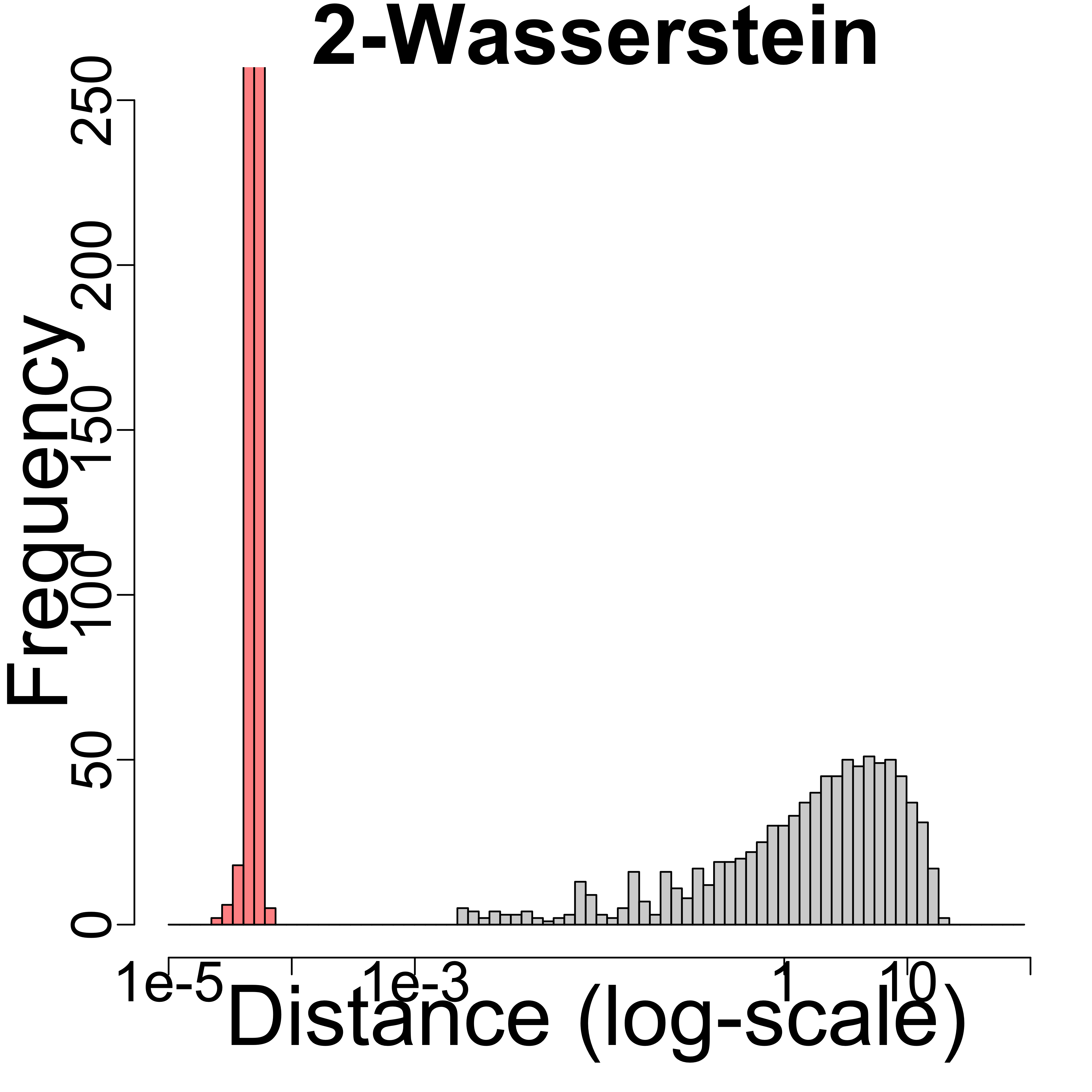}
\includegraphics[width = 0.325\textwidth]{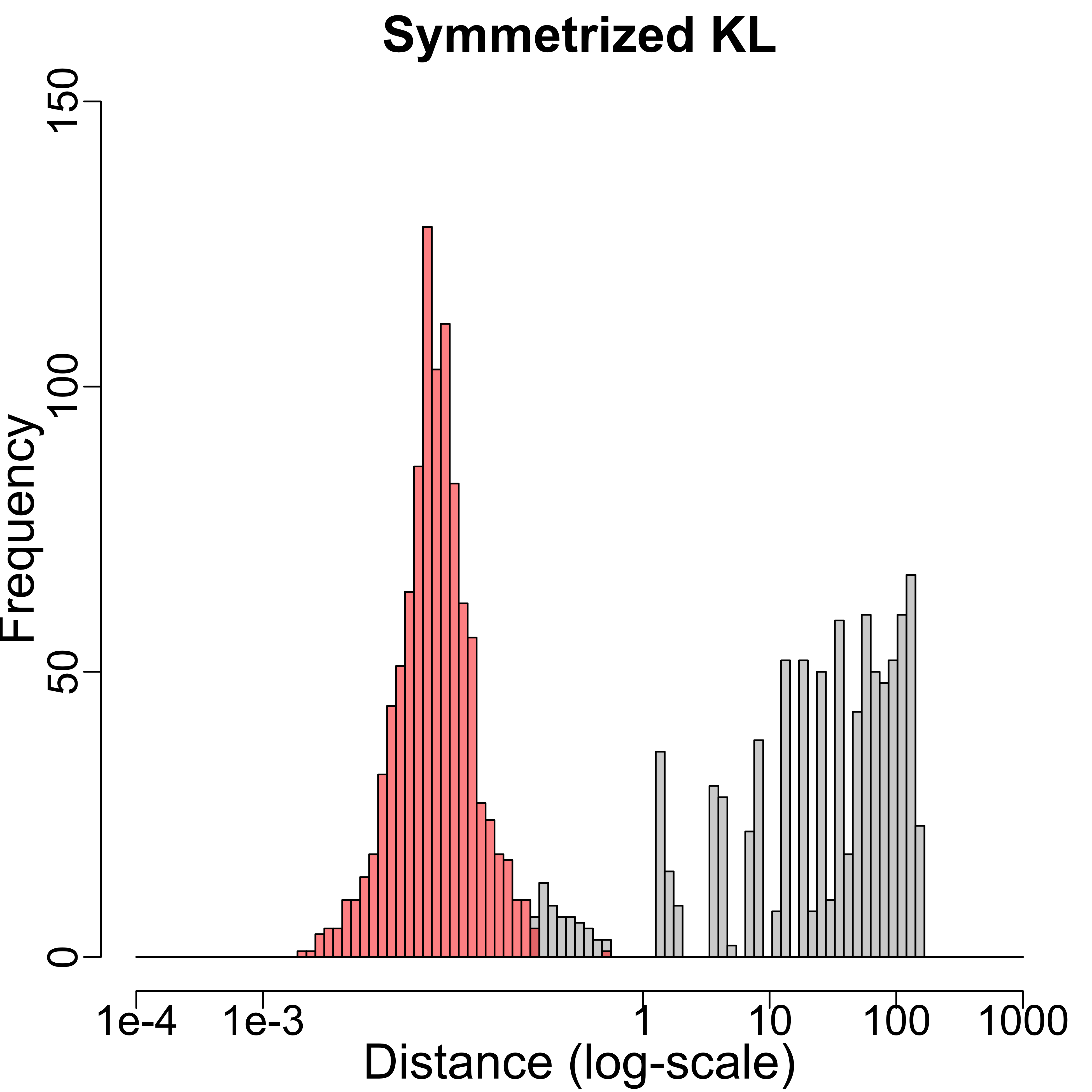}
\caption{Extra hyperparameter uncertainty histograms for our MNIST experiment in \cref{sec:MNIST} in which we find non-robustness. We compare the difference between $k_{0}$ and $\kperturb$ (red) to bootstrapped hyperparameter uncertainty (gray) in several distances. Note, distances are plotted on the log-scale.}
\label{fig:mnist_histograms}
\end{figure*}

Our \co2 modeling example and MNIST example do not use the stationary constraints from \cref{constraintSet:stationary}.
We see in \cref{fig:mauna_loa_histograms,fig:mnist_histograms} that under all alternative choices of $d$ we consider here (including the spectral and infinity norms), we reach the same conclusions about qualitative interchangeability as we did under the 2-Wasserstein distance.

%\input{app_assets_used}

%
%%%%%%%%%%%%%%%%%%%%%%%%%%%%%%%%%%%%%%%%%%%%%%%%%%%%%%%%%%%%%
%
%\appendix
%
%\section{Appendix}
%
%Optionally include extra information (complete proofs, additional experiments and plots) in the appendix.
%This section will often be part of the supplemental material.

\end{document}